\documentclass[sigconf]{acmart}

\usepackage{xspace}
\usepackage{subcaption}
\usepackage{multirow}
\usepackage{algorithm}
\usepackage{algorithmic}
\usepackage{xfrac}

\newcommand{\framework}{RIDGE\xspace}

\AtBeginDocument{%
  }

\copyrightyear{2025}
\acmYear{2025}
\setcopyright{cc}
\setcctype{by}
\acmConference[MM '25]{Proceedings of the 33rd ACM International Conference on Multimedia}{October 27--31, 2025}{Dublin, Ireland}
\acmBooktitle{Proceedings of the 33rd ACM International Conference on Multimedia (MM '25), October 27--31, 2025, Dublin, Ireland}
\acmDOI{10.1145/3746027.3755645}
\acmISBN{979-8-4007-2035-2/2025/10}

\begin{document}

\title[Toward Robust Signed Graph Learning through Joint Input-Target Denoising]{Toward Robust Signed Graph Learning through Joint Input-Target Denoising}

% \title[RINSE: Robust Link Sign Prediction via Mutual Information Compression]{RINSE: Robust Link Sign Prediction via \\ Mutual Information Compression}

% Toward Robust Signed Graph Learning through Joint Input-Target Denoising


\author{Junran Wu}
\orcid{0000-0001-6742-4332}
\affiliation{%
  \institution{National University of Singapore}
  \city{Singapore}
  \country{Singapore}}
\email{junran@nus.edu.sg}

\author{Beng Chin Ooi}
\orcid{0000-0003-4446-1100}
\authornote{The work was done while working for National University of Singapore.}
\affiliation{%
  \institution{Zhejiang University}
  \city{Ningbo}
  \country{China}}
\email{ooibc@zju.edu.cn}

\author{Ke Xu}
\orcid{0000-0002-6241-8352}
\affiliation{%
  \institution{Beihang University}
  \city{Beijing}
  \country{China}}
\email{kexu@buaa.edu.cn}

\begin{abstract}
Signed Graph Neural Networks (SGNNs) are widely adopted to analyze complex patterns in signed graphs with both positive and negative links. 
Given the noisy nature of real-world connections, the robustness of SGNN has also emerged as a pivotal research area. 
Under the supervision of empirical properties, graph structure learning has shown its robustness on signed graph representation learning, however, there remains a paucity of research investigating a robust SGNN with theoretical guidance.
Inspired by the success of graph information bottleneck (GIB) in information extraction, we propose \textbf{\framework}, a novel framework for \textit{R}obust s\textit{I}gned graph learning through joint \textit{D}enoising of \textit{G}raph inputs and supervision targ\textit{E}ts.
Different from the basic GIB, we extend the GIB theory with the capability of target space denoising as the co-existence of noise in both input and target spaces.
In instantiation, \framework effectively cleanses input data and supervision targets via a tractable objective function produced by reparameterization mechanism and variational approximation. 
We extensively validate our method on four prevalent signed graph datasets, and the results show that \framework clearly improves the robustness of popular SGNN models under various levels of noise.
\end{abstract}






\maketitle

\section{Introduction}
Real-world relationships invariably exhibit complexity. Beyond the general unsigned graphs, which only model positive correlations, there exists an alternative class of graphs that incorporate both positive and negative links~\cite{kunegis2013added}. 
These links typically represent diametrically opposed relationships between entities, such as friendship/enmity, trust/distrust, and similarity/dissimilarity. 
Therefore, besides the conventional Graph Neural Networks (GNNs) that are tailored for unsigned networks and learn node representations relying on the principle of homophily~\cite{mcpherson2001birds}, various Signed Graph Neural Networks (SGNNs) have been proposed. These SGNNs are designed to analyze the signed graphs that are composed of links annotated with different polarities~\cite{derr2018signed,li2020learning,li2023signed}.

\begin{figure*}[!ht]
\centering
\includegraphics[width=0.9\textwidth]{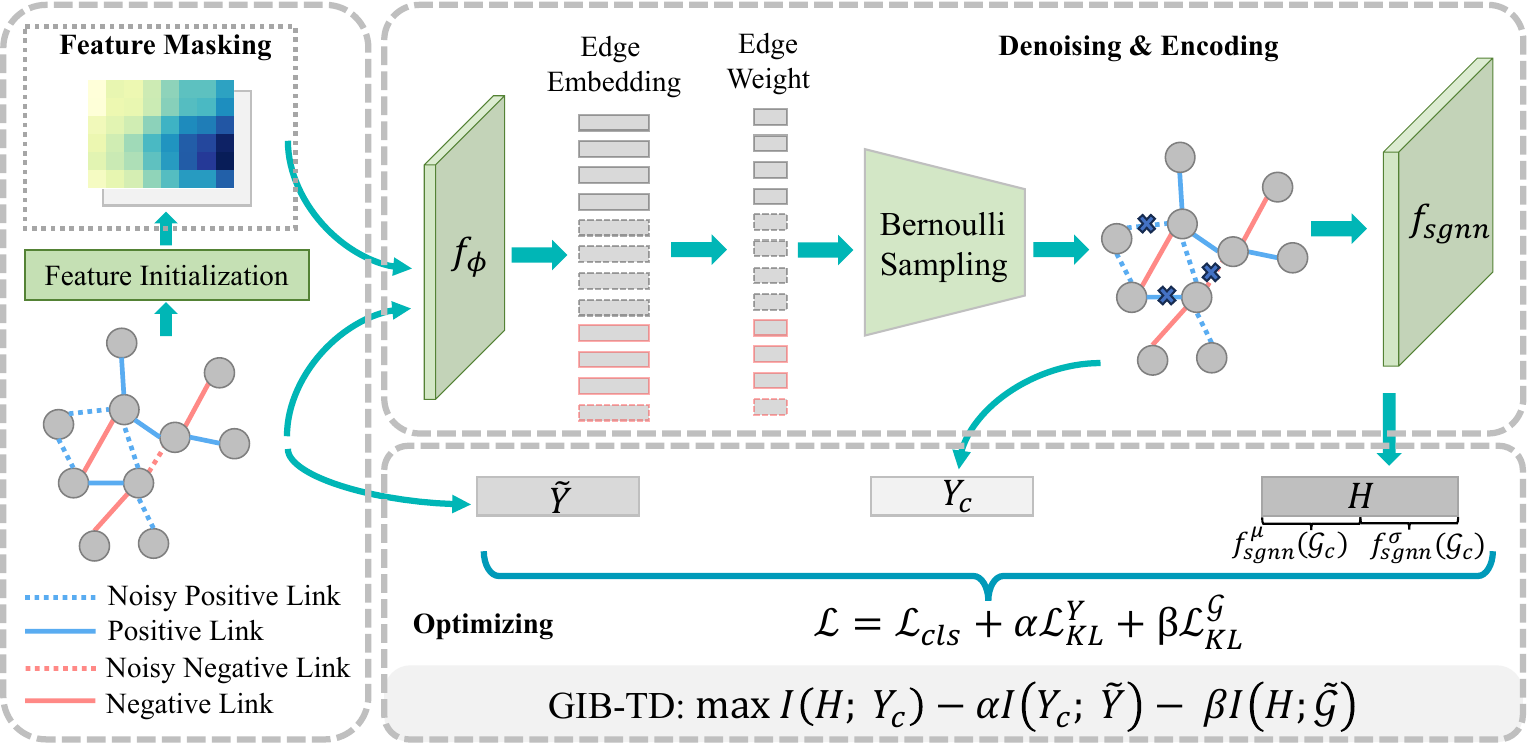}
\caption{\textbf{Framework overview of \framework.} \framework extends the GIB theory with the ability of target space denoising, referred to as \textit{GIB-TD}, in which $Y_c$ denotes the clean subset extracted from the noisy label set $\tilde{Y}$. Given a signed graph with random noise, \framework is capable of deriving robust representations by combating the noise inherent in the input and target spaces using the reparameterization mechanism and variational approximation.}
\label{fig:framework}
\end{figure*}

Despite the progress made, recent research has revealed that SGNNs are highly sensitive to random noise in the dataset~\cite{zhang2023contrastive,zhang2023rsgnn}.
Graph datasets can be affected by noise stemming from the data acquisition process, whether due to deliberate manipulation or unintentional errors.
Consider, for instance, an e-commerce platform that offers incentives for users to leave product reviews; some users may submit arbitrary feedback purely to claim these incentives.
In these cases, the resulting signed edges may fail to faithfully capture the genuine interactions between entities in the graph.
Consequently, the performance of SGNNs would significantly deteriorate under the influence of these subtle perturbations in graphs. 
This lack of robustness can have serious implications for critical applications, particularly those related to safety and privacy. 
Therefore, the development of robust SGNN models capable of resisting noisy connections is of paramount importance.

In general robust GNNs, graph structure learning (GSL)\cite{jin2020graph,li2022chart} has emerged as a key area of interest. The objective of GSL is to refine or derive a graph structure from a given adjacency matrix by aligning the learned structure with certain properties~\cite{franceschi2019learning,wang2023prose}.
Following the routine of robust GNNs, RSGNN~\cite{zhang2023rsgnn} serves as the pioneer for signed graph denoising by aligning the GSL process with three empirical properties, including sparsity, feature smoothness and high balance degree.
However, among these properties, the first two (i.e., sparsity and feature smoothness) are extrapolated from unsigned graphs. 
While the concept of balance degree stems from classical balance theory~\cite{heider1946attitudes,cartwright1956structural}, it fundamentally relies on a simplified premise:
nodes can be divided into two non-overlapping groups, with positive edges only appearing within each group and negative edges occurring exclusively between them~\cite{mercado2016clustering}.
This assumption rarely holds in real-world signed networks~\footnote{A detailed analysis of balance theory is presented in Sec.\ref{sec:notation}.}, which largely limits the efficacy of learned node representations in downstream signed graph tasks.
This underscores the existing research gap in the investigation of a robust SGNN guided by a solid theory.

Recalling the above concerns, the crux of robust SGNNs lies in uncovering the underlying connections that remain invariant to task-irrelevant information. 
The Graph Information Bottleneck (GIB)~\cite{wu2020graph} offers a framework to constrain such task-irrelevant information by striking a balance between prediction and compression, formally expressed as $\max I(H;Y)-\beta I(H; \mathcal{G})$.
Given the robustness of the GIB principle, our approach diverges from the pursuit of a structure learning method aimed at producing a more balanced graph. Instead, we leverage the robust theoretical foundation of GIB to bolster the robustness of SGNNs.

We now present our framework \textbf{\framework}, as depicted in Fig.~\ref{fig:framework}.~\footnote{The implementation of \framework is available at \url{https://github.com/Junranus/RIDGE}.}
Conceptually, \framework diverges from the basic GIB theory by fostering new learning objectives that extract effective information from the noisy inputs (i.e., $\tilde{\mathcal{G}} = \{\tilde{X}, \tilde{A}\}$) and noisy labels (i.e., $\tilde{Y}$).
In particular, to address the limitation of GIB in the target space denoising, we propose a new learning objective (refer to Proposition~\ref{prop:up_hy}) for prediction label denoising.
Technically, \framework utilizes irrelevant feature masking to distill actionable information for downstream tasks.
Furthermore, \framework explicitly cleanses the graph topology and supervision targets using a reparameterization mechanism and variational approximation.
To effectively estimate and balance informative terms in a tractable manner, the overall framework can be efficiently trained using supervised classification loss and relaxed information constraints on both input and label spaces.
We conduct extensive experiments on four real-world signed graph datasets to assess the effectiveness of \framework, showing that it consistently outperforms existing methods across diverse noisy settings.
Our main contributions are outlined below:

\begin{itemize}
\item To the best of our knowledge, we are the first to propose the application of GIB theory onto robust SGNNs.

\item Derived from GIB, we introduce \framework that utilizes the data reparameterization mechanism and variational approximation to construct a tractable objective that effectively cleanses both input and target spaces.

\item \framework is evaluated on four signed graph datasets, where it consistently achieves strong performance and resilience to noise, yielding up to a 5.45\% gain in Binary-F1 score.
\end{itemize}

\section{Related Work}
\label{sec:app_related}
Within the scope of this study, we provide a concise overview of the most relevant research on signed graph representation learning and the graph information bottleneck framework.

\noindent \textbf{Signed graph representation learning} has its origins in early works on signed Laplacian eigendecomposition~\cite{hou2003laplacian} and matrix factorization techniques~\cite{hsieh2012low}.
Later developments have primarily leveraged balance theory and derived extensions to representing signed relations.
For instance, SIDE~\cite{kim2018side} and SIGNET~\cite{islam2018signet} utilize random walk approaches to capture structural balance.
SGCN~\cite{derr2018signed} presents the first signed GCN grounded in balance theory, while models like SiGAT~\cite{huang2019signed} and SNEA~\cite{li2020learning} incorporate attention mechanisms~\cite{velivckovic2018graph} to weigh neighbors by relevance.
GS-GNN~\cite{liu2021signed} generalizes balance theory into k-group theory to jointly learn local and global representations.
To move beyond the limitations of balance-based modeling, SLGNN~\cite{li2023signed} introduces a Laplacian regularization framework tailored for signed graphs.
More recently, SE-SGformer~\cite{li2025se} proposes a transformer-based architecture with self-explainability to improve both performance and interpretability in link sign prediction.
However, the efficacy of these SGNNs is largely contingent upon the high-quality graph topology, an assumption that is excessively idealistic for real-world graphs.

Beyond the aforementioned methods, several other approaches for signed graphs have been designed with robustness to noise in mind.
SGCL~\cite{shu2021sgcl} integrates contrastive learning into SGNNs and introduces controlled graph augmentations, aiming to improve resilience through exposure to synthetic perturbations.
GS-GNN~\cite{liu2021signed} broadens the classical balance theory to a k-group formulation, offering greater adaptability to tolerate random structural noise.
RSGNN~\cite{zhang2023rsgnn} is the first dedicated effort toward robust signed GNNs via graph structure learning, informed by empirical insights.
LRWSB~\cite{minici2024link} adopts a semi-supervised strategy that improves robustness by incorporating multiscale notions of social balance.
Nevertheless, both RSGNN and LRWSB fundamentally depend on balance theory assumptions, which often fall short in complex, noisy real-world graphs.
Moreover, RSGNN focuses exclusively on edge perturbations, overlooking that input node features themselves are derived from potentially corrupted structures.
In contrast, our approach advances the GIB theory to the signed graph setting, offering a general and theoretically grounded solution for denoising through an information-theoretic lens.

\noindent \textbf{Graph information bottleneck} extends the general IB principle for graph data, incorporating regularization of both structure and feature information to yield robust node representations~\cite{wu2020graph}. Subsequently, SIB~\cite{yu2020graph} was proposed to address the subgraph recognition problem, building upon the foundation of GIB. HGIB~\cite{yang2021heterogeneous} was proposed to capture the consensus principle in heterogeneous information networks without relying on supervision signals.
While AD-GCL~\cite{suresh2021adversarial} applies the GIB principle in graph contrastive learning to avoid the capture of redundant information. 
Despite the various applications of the GIB theory~\cite{zhu2024hill,hou2025structural}, the question of how to leverage GIB for robust SGNNs remains an open problem. 
The employment of GIB in GNNs is intended to fortify their robustness in the presence of noise within both the feature and topology spaces. In contrast, our proposed \framework framework diverges by addressing not only the noise in the feature and topology spaces but also undertaking the compression of noise within the target space.
As far as we are aware, \framework represents the first formal work that introduces the theory of GIB for robust signed graph representation learning.

\begin{figure}[!tp]
  \centering
  \begin{subfigure}{.47\linewidth}
    \centering
    \includegraphics[width=\linewidth]{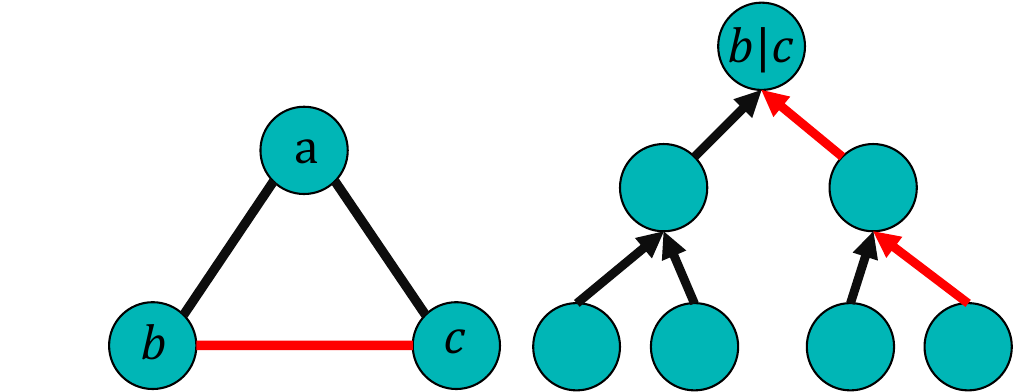}
    \caption{Case 1.}
    \label{fig:unbalance_case1}
  \end{subfigure} 
  \begin{subfigure}{.47\linewidth}
    \centering
    \includegraphics[width=\linewidth]{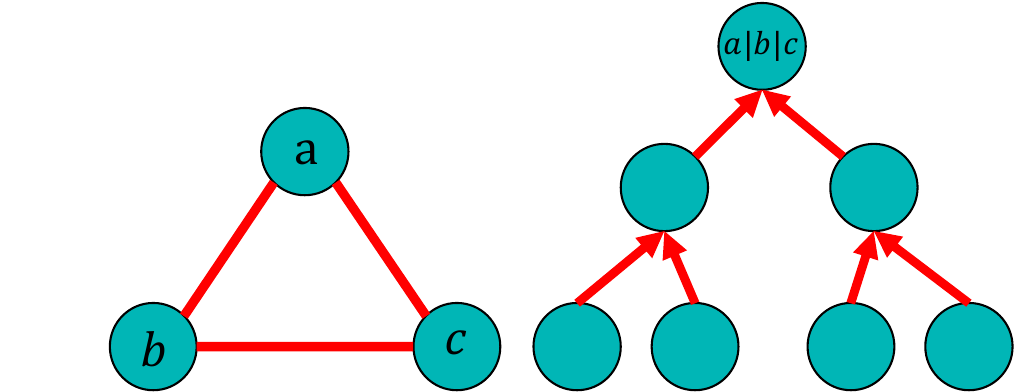}
    \caption{Case 2.}
    \label{fig:unbalance_case2}
  \end{subfigure} 
  \caption{\textbf{Two common unbalanced triangles in signed graphs and their corresponding ego-trees.} Positive edges are shown in black, while negative edges are depicted in red.}
  \label{fig:unbalance_case}
\end{figure}

\section{Notations and Preliminaries}
\label{sec:notation}

\noindent \textbf{Notations.} For the convenience of presentation, we begin by outlining the key notations used throughout this paper.
Let $\mathcal{G}=\{\mathcal{U}, \mathcal{E}^+, \mathcal{E}^-\}$ be a signed graph, where $\mathcal{U} = \{u_1, u_2, \dots, u_n\}$ is the set of nodes and $|\mathcal{U}|=n$, and $\mathcal{E}^+ \cup \mathcal{E}^-$ contains $m$ edges in total.
$\mathcal{E}^+\subset \mathcal{U}\times \mathcal{U}$ and $\mathcal{E}^-\subset \mathcal{U}\times \mathcal{U}$ represent the sets of positive and negative edges, respectively, with the constraint $\mathcal{E}^+ \cap \mathcal{E}^- = \emptyset$ ensuring that no edge is simultaneously positive and negative.
$A \in \mathbb{P}_{\upsilon}^{n \times n}$ is designed to indicate the signed adjacency matrix of the given $\mathcal{G}$, where $A_{ij} = 1$ indicates a positive edge from node $u_i$ to $u_j$, $A_{ij} = -1$ refers to a negative relation, and $A_{ij} = 0$ means no direct link exists between them.
The training label set $Y$, which is composed of the link signs, is derived from the adjacency matrix $A$. 
Correspondingly, the noisy input graph is $\tilde{\mathcal{G}}=\{\mathcal{U}, \tilde{\mathcal{E}^+}, \tilde{\mathcal{E}^-\}}$ with noisy adjacency matrix $\tilde{A}$ and noisy label set $\tilde{Y}$.
Detailed notations frequently used in this paper are shown in Table A.1.

\noindent \textbf{Link sign prediction.}
Let $\mathcal{G} = \{\mathcal{U}, \mathcal{E}\}$ represent the input signed graph used for training, where the edge set $\mathcal{E} = \{\mathcal{E}^+, \mathcal{E}^-\}$ contains observed edges labeled with positive or negative signs.
The goal is to infer the signs of a separate set of unseen edges, denoted as $\mathcal{E}_p = \{\mathcal{E}^+_{p}, \mathcal{E}^-_{p}\}$, whose labels are not available during training.
This task is generally formulated as a binary classification problem, where each edge is assigned a label of $+1$ or $-1$ based on patterns learned from the observed edges. A further comparison with other graph tasks is illustrated in Appendix B.

\begin{figure}[!tp]
  \centering
  \includegraphics[width=0.85\linewidth]{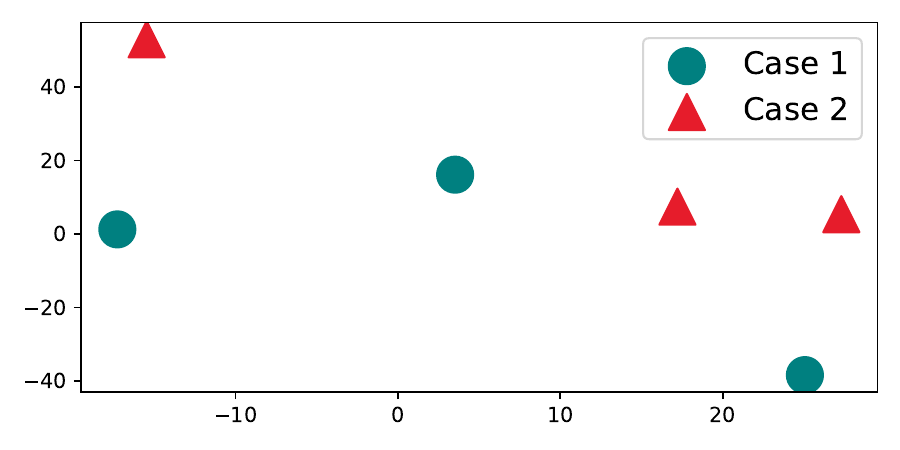}
  \caption{t-SNE visualization of the node features produced by Truncated-SVD of the two unbalanced triangles in Fig~\ref{fig:unbalance_case}.}
  \label{fig:unbalance_features}
\end{figure}

\noindent \textbf{Random noise.} Let $\mathcal{G} = \{\mathcal{U}, \mathcal{E}^+, \mathcal{E}^-\}$ be signed graph for training, we introduce synthetic noise by randomly flipping the signs of a subset of edges. The fraction of flipped edges determines the noise level applied to graph.
Specifically, for a given noise ratio $\gamma$, the noisy adjacency matrix $\tilde{A}$ is generated by multiplying each element in $A$ by -1 with probability $\gamma$, such that $\tilde{A} = (1-\text{NoiseMask}) \odot A - \text{NoiseMask} \odot A$. This ensures that $|\tilde{A}| = |A|$ and $\gamma = |\mathbf{nonzero}(\text{NoiseMask})|/|\mathbf{nonzero}(A)|$. The corresponding noisy label set $\tilde{Y}$ is then generated based on $\tilde{A}$.

In practice, widely used four signed graph benchmarks typically lack explicit node features.
To address this, prior work~\cite{derr2018signed,li2023signed} applies Truncated-SVD~\cite{halko2011finding} on the graph structure to derive initial node embeddings.
Consequently, these features, denoted as $X$, are inherently influenced by the topology and thus also affected by noise, yielding perturbed features $\tilde{X}$.
In summary, the injected noise affects both the input components ($\tilde{A}$, $\tilde{X}$) and the supervision signal ($\tilde{Y}$), simulating a realistic noisy learning scenario.

\noindent \textbf{Why random noise?} In line with the settings in RSGNN~\cite{zhang2023rsgnn}, our main goal is to enhance the resilience of SGNNs to random noise, an artifact stemming from imperfect data acquisition, which is distinct from adversarial perturbations crafted with malicious intent.
Beyond this main focus, we also validate the effectiveness of our method under adversarial noise and alternative types of random noise, with detailed results provided in Appendix E and Appendix F.

\noindent \textbf{Utopia of balance theory.} At present, most existing SGNNs in their framework design~\cite{derr2018signed,li2020learning,zhang2023contrastive} heavily relies on balance theory~\cite{heider1946attitudes,cartwright1956structural}, a classic concept from social psychology. In essence, it reflects principles like ``the foe of my foe is my friend''~\cite{heider2013psychology}.
Despite its intuitive appeal, balance theory is mathematically equivalent to assuming that nodes can be perfectly separated into two clusters, where all inter-group links are negative and all intra-group edges are positive~\cite{mercado2016clustering}.
This assumption, however, rarely holds in complex real-world signed networks, making it an unrealistic foundation for robust modeling.

As shown in Fig.~\ref{fig:unbalance_case}, the two unbalanced triangles are usually adopted by previous research to illustrate the inherent limitation of general GNNs in proper node representation learning.
The primary reason is that nodes $b$ and $c$ in \textit{Case 1} possess identical ego trees for message passing, leading prior researchers to infer that nodes $b$ and $c$ in this case will yield the same representation~\footnote{Analogously, all nodes ($a$, $b$, and $c$) in \textit{Case 2} would have the same representation.}. 
However, such a conclusion hinges on a tough assumption that all nodes possess identical initial labels or input features, a premise that diverges from real-world application scenarios.
In practice, previous SGNNs~\cite{derr2018signed,li2023signed} utilize Truncated-SVD~\cite{halko2011finding} to generate initial node features by encoding the input signed graph structure. Taking the two cases in Fig.~\ref{fig:unbalance_case} as input, the resulting initial node embeddings are illustrated in Fig.~\ref{fig:unbalance_features}, which suggests that all nodes have distinct initial features and challenges the assumption for improper node representation.
Hence, instead of pursuing a GSL algorithm that produces a more balanced graph, we resort to the GIB theory to bolster the robustness of SGNNs.

\section{Methodology}
In this following, we elaborate on the proposed \framework for robust link sign prediction.
We first present the theory of GIB and extend it to encompass target space denoising. 
Following this, we give the instantiation of \framework, which is predicated on feature masking and substructure sampling, facilitated by the tractable upper bound via reparameterization mechanism and variational approximation.
At last, we compare \framework with two of its most closely related works, underscoring our unique advantages and contributions.

\subsection{Theoretical Formulation of \framework}
\label{sec:gib-td}
Given the degraded edge representation $H$, produced by a signed graph encoder $f_{sgnn}$ as $H = f_{sgnn}(\tilde{\mathcal{G}})$, where $\tilde{\mathcal{G}} = (\tilde{X}, \tilde{A})$, we aim to robustify $H$ using an information constraint based on GIB. Formally,
\begin{equation}
\text{GIB}: \max I(H; \tilde{Y})-\beta I(H; \tilde{\mathcal{G}}),
\label{eq:gib}
\end{equation}
where $I(\cdot)$ measure the mutual information and $\beta > 0$ that constrains the mutual information $I(H;\tilde{\mathcal{G}})$ to avoid $H$ from capturing excess task-irrelevant information from $\tilde{\mathcal{G}}$.

\noindent \textbf{Inability of GIB for sign graph with random noise.} In contrast to the general robust research in GNNs, the scenario of robust link sign prediction presents a unique challenge. Here, noise permeates not only the input data but also persists within the label sets, denoted as $\tilde{Y}$. However, merely constraining $I(H; \tilde{Y})$ to regularize the noise would be suboptimal, particularly given that it conflicts with the purpose of Eq.~\ref{eq:gib}.
Furthermore, if we strictly adhere to the directive of Eq.~\ref{eq:gib}, that is, $\max I(H; \tilde{Y})$, the basic GIB becomes inherently susceptible to label noise. This is because it preserves the label supervision in its entirety and maximizes the noisy supervision. 
The ablation study in Sec.~\ref{sec:exp} demonstrates that optimization becomes ineffective when the prediction target is contaminated by noise.
Therefore, it becomes imperative to extract effective information from the noisy label set. Considering the second term of GIB, which pertains to noise from input data, we propose an upper bound for the first term, specifically aimed at compressing label noise.

\begin{proposition}
(\textbf{Upper bound of $-I(H; \tilde{Y})$}). Given the clean subset $Y_c$ of the noisy label set $\tilde{Y}$, and the representation $H$ of all query edges learned from $\tilde{\mathcal{G}}$, we have 
\begin{equation}
-I(H; \tilde{Y}) \leq -I(H; Y_c) + I(Y_c; \tilde{Y}).
\label{eq:ub_hy}
\end{equation}
\label{prop:up_hy}
\end{proposition}

\begin{proof}
By the data processing inequality~\cite{thomas2006elements}, we first have:
\begin{equation}
I(H; \tilde{Y})  \geq I(H; Y_c) \geq I(H; Y_c) - I(Y_c; \tilde{Y}).
\end{equation}
Thus, we obtain $-I(H; \tilde{Y}) \leq -I(H; Y_c) + I(Y_c; \tilde{Y})$.
\end{proof}

The clean subset $Y_c$ is extracted from $\tilde{Y}$, and its complementary part $Y'$ is considered as noise, satisfying $\tilde{Y} = Y_c + Y'$. When $Y_c\equiv \tilde{Y}$, the upper bound can be degenerated to the basic half part of GIB. Here, the $I(H; Y_c)$ measures the supervised signals with selected samples $Y_c$, while constraining $I(Y_c; \tilde{Y})$ aims to select the cleanest and most task-relevant information from $\tilde{Y}$.

\noindent \textbf{GIB with target space denoising (i.e., GIB-TD).} Drawing from the above analysis of GIB, we unveil its ineffectiveness when learning with noisy supervision. To address this issue, we introduce Prop.~\ref{prop:up_hy}, which augments the naive GIB with the capability of target space denoising. In this study, we formally present the overarching theoretical principle that guides robust link sign prediction:
\begin{equation}
\text{GIB-TD}: \max I(H; Y_c)-\alpha I(Y_c; \tilde{Y}) - \beta I(H; \tilde{\mathcal{G}}),
\label{eq:gib-td}
\end{equation}
where $\alpha$ and $\beta$ are employed to balance the information extracted from different aspects. Specifically, the $I(H; Y_c)$ and $I(H; \tilde{\mathcal{G}})$ are rooted in the basic GIB, effectively defending the input perturbation~\cite{wu2020graph}. The $I(Y_c; \tilde{Y})$ is derived from Prop.~\ref{prop:up_hy} and serves to extract effective supervision signals. 
Next, we will instantiate these information terms specifically for robust link sign prediction.

\subsection{Instantiation}
Building upon the theory formulated in Eq.~\ref{eq:gib-td}, we introduce our method, \framework, which strategically implements subsequent steps to extract effective information from the noisy data. 

\noindent \textbf{Feature masking.} In link sign prediction, noise would propagate from the structure to the initial node input (i.e., $\tilde{X}$) due to feature initialization. Thus, we first use a feature masking scheme to discretely drop features that are irrelevant to the downstream task:
\begin{equation}
X_c = \{X_i\odot M, \,\,i=1, 2,\cdots, |\mathcal{U}|\},
\end{equation}
where $M\in\mathbb{P}_{\upsilon}^d$ is a learnable binary feature mask and $\odot$ denotes the element-wise product. Intuitively, if a particular feature is irrelevant to task, the corresponding weight in $M$ takes value close to zero. We can reparameterize $X_c$ using the reparameterization trick~\cite{kingma2014auto} to backpropagate through a $d$-dimensional random variable:
\begin{equation}
X_c = X_r + (X - X_r)\odot M, 
\end{equation}
where $X_r$ is sampled from the empirical distribution of $X$.

\noindent \textbf{Substructure sampling.} Analogously, we adopt the same reparameterization mechanism to cleanse the topology and target space. For obtaining a clean substructure regarding $A_c$ and $Y_c$, a parameterized sampler $f_\phi(\cdot)$ sharing the same architecture and weights as $f_{sgnn}(\cdot)$ is adopted here. $f_\phi(\cdot)$ generates the
probabilistic distribution of edges that include both $\tilde{A}$ and $\tilde{Y}$ by $P = \sigma (H_\phi, H_\phi^T) \in (0, 1)^{|\mathcal{U}|\times |\mathcal{U}|}$, where representation $H_\phi =f_\phi(\tilde{A}, X_c)$ and $\sigma$ is the activation function (i.e., Sigmoid). Bernoulli sampling is then used to obtain high-confidence edges; put differently, $A_c=\text{SAMP}(P|\tilde{A})$ and $Y_c = \text{SAMP}(P|\tilde{Y})$, where $|A_c|\leq|\tilde{A}|$ and $|Y_c|\leq|\tilde{Y}|$.
In this work, the clean subset $Y_c$ is sampled from the original label set, ensuring that $Y_c$ inherently retains supervision signals from $\tilde{Y}$, thereby eliminating the risk of no supervision.
As shown in Sec.~\ref{sec:results}, substructure sampling would not lead to high variance and unstable training.

\noindent \textbf{GIB-TD instantiation.} Given the denoised input graph, denoted as $\mathcal{G}_c=(X_c, A_c)$, and its corresponding supervision signal $Y_c$, the mutual information terms in Eq.~\ref{eq:gib-td} are subsequently implemented by the following tractable objective functions.

\begin{proposition} 
\label{prop:cls}
The supervision term $I(H; Y_c)$ in Eq.~\ref{eq:gib-td} can be empirically reduced to the classification loss:
\begin{align} 
I(H; Y_c) & \geq \mathbb{E}_{Y_c, \mathcal{G}_c}[\log\mathbb{P}_{sgnn}(Y_c|\mathcal{G}_c)] \nonumber \\ 
          & \approx -\mathcal{L}_{cls}(f_{sgnn}(\mathcal{G}_c), Y_c), 
\end{align}
where $\mathcal{L}_{cls}$ is the standard cross-entropy loss.
\end{proposition}

\begin{proof}
The lower bound of $I(H; Y_c)$ can be obtained as:
\begin{align}
\label{eq:lb_cls}
I(H; Y_c) & = I(f_{sgnn}(\mathcal{G}_c); Y_c) \nonumber \\
& \geq \mathbb{E}_{Y_c, \mathcal{G}_c}[\log\frac{\mathbb{P}_{sgnn}(Y_c|\mathcal{G}_c)}{\mathbb{P}(Y_c)}] \nonumber \\
& = \mathbb{E}_{Y_c, \mathcal{G}_c}[\log\mathbb{P}_{sgnn}(Y_c|\mathcal{G}_c) - \log \mathbb{P}(Y_c)] \nonumber \\
& = \mathbb{E}_{Y_c, \mathcal{G}_c}[\log\mathbb{P}_{sgnn}(Y_c|\mathcal{G}_c) + H(Y_c)] \nonumber \\
& \geq \mathbb{E}_{Y_c, \mathcal{G}_c}[\log\mathbb{P}_{sgnn}(Y_c|\mathcal{G}_c)] \nonumber \\
& \approx -\frac{1}{|Y_c|}\sum_{e_{ij}\in Y_c}\mathcal{L}_{cls}(f_{sgnn}(\mathcal{G}_c), Y_c).
\end{align}

Thus, the mutual information $I(H; Y_c)$ can be lower bounded by the standard classification loss, and $\min -I(H; Y_c)$ in Eq.\ref{eq:gib-td} is upper bounded by $\min -1/|Y_c|\sum_{e_{ij}\in Y_c}\mathcal{L}_{cls}(f_{sgnn}(\mathcal{G}_c), Y_c)$ as Eq.\ref{eq:lb_cls}.
\end{proof}

\begin{algorithm}[!tp]
\caption{Learning process of \framework}
\label{code:design} 
\begin{flushleft}
\textbf{Input:} a noisy signed graph $\tilde{\mathcal{G}}=(\tilde{X}, \tilde{A})$, $\tilde{X}=\text{Truncated-SVD}(\tilde{A})$ and corresponding noisy label set $\tilde{Y}$\\
\textbf{Output:} a trained SGNN $f_{sgnn}$
\end{flushleft}

\begin{algorithmic}[1]
\STATE Parameter initialization;
\FOR{each epoch} {
\STATE \textit{// Feature Masking}
\STATE $X_c \leftarrow \{X_i\odot M, i\in\mathcal{U}\}$;
\STATE $H_\phi = f_\phi(\tilde{A}, X_c)$;
\STATE \textit{// Substructure Sampling}
\STATE $P = \sigma (H_\phi, H_\phi^T) \in (0, 1)^{|\mathcal{U}|\times |\mathcal{U}|}$;
\STATE $A_c = SAMP(P|\tilde{A})$;
\STATE $Y_c = SAMP(P|\tilde{Y})$;
\STATE \textit{// GIB-TD}
\STATE $H = f_{sgnn}(X_c, A_c)$;
\STATE Encode $(f_{\phi}^\mu(X_c, A_c), f_{\phi}^\sigma(X_c, A_c))$ by $f_{sgnn}$;
\STATE $\mathcal{L} = \mathcal{L}_{cls} + \alpha\mathcal{L}_{KL}^{Y} + \beta\mathcal{L}_{KL}^{\mathcal{G}}$;
\STATE \textit{// Optimization}
\STATE Minimize $\mathcal{L}$ via model updating;
}
\ENDFOR

\STATE return $f_{sgnn}$;
\end{algorithmic} 
\end{algorithm}

\begin{proposition} 
\label{prop:up_ycy}
(\textbf{Upper bound of $I(Y_c, \tilde{Y})$}).
Given noisy label set $|\tilde{Y}|=m$, the marginal distribution of $Y_c$ is $\mathbb{Q}(Y_c)=\mathbb{P}(m)\prod^m_{e_{ij}}P_{ij}$, and $Y_c$ satisfies 
\begin{align}
I(Y_c; \tilde{Y}) & \leq \mathbb{E}[\text{KL}(\mathbb{P}_\phi(Y_c|Y)||\mathbb{Q}(Y_c))] \nonumber \\
                  & =\sum_{e_{ij}\in\tilde{A}}P_{ij}\log\frac{P_{ij}}{\tau}+(1-P_{ij})\log\frac{1-P_{ij}}{1-\tau}, 
\end{align}
where $\tau$ is a constant and KL denotes the Kullback-Leiber (KL) divergence~\cite{hershey2007approximating}.
\end{proposition}

\begin{proof}
Given $\mathbb{Q}(Y_c)=\mathbb{P}(m)\prod^m_{e_{ij}}P_{ij}$, we obtain the upper bound mutual information $I(Y_c; \tilde{Y})$ as:
\begin{align}
I(Y_c; \tilde{Y}) & = \mathbb{E}_{Y_c, \tilde{Y}}[\log \frac{\mathbb{P}(Y_c|\tilde{Y})}{\mathbb{P}(Y_c)}] \nonumber \\
& = \mathbb{E}_{Y_c, \tilde{Y}}[\log \frac{\mathbb{P}(Y_c|\tilde{Y})}{\mathbb{Q}(Y_c)}] - \text{KL}(\mathbb{P}(Y_c)||\mathbb{Q}(Y_c)) \nonumber \\
& \leq \mathbb{E}[\text{KL}(\mathbb{P}_\phi(Y_c|Y)||\mathbb{Q}(Y_c))] \nonumber \\
& = \sum_{e_{ij}} (1-P_{ij})\log\frac{1-P_{ij}}{1-\tau} + P_{ij}\log\frac{P_{ij}}{\tau}.
\label{eq:ub_iycy}
\end{align}
The KL divergence on two distribution $\mathbb{P}(x)$ and $\mathbb{Q}(x)$ is defined as $\text{KL}(\mathbb{P}(x)||\mathbb{Q}(x)) = \sum_x\mathbb{P}(x)\log\mathbb{P}(x)/\mathbb{Q}(x)$. Based on this definition, the bound of $I(Y_c; \tilde{Y})$ can be obtained via Eq.~\ref{eq:ub_iycy}.
\end{proof}

\begin{table*}[!ht]
\centering
\caption{\textbf{Average Binary-F1 (\%)$\pm$Std over different 5 runs under various noise settings.} AUC is omitted as it exhibits analogous trends. Best performance over all methods is marked with \textbf{Bold} font. \underline{Underline} highlights the best results among robust SGNNs.}
\label{tab:main_result}
\resizebox{\textwidth}{!}{%
\begin{tabular}{l|c|cccccc|ccc|ccc}
\hline \hline
\multirow{2}{*}{Dataset} & \multirow{2}{*}{Noise (\%)} & \multicolumn{6}{c|}{basic SGNN} & \multicolumn{3}{c|}{noise-tolerant SGNN} & \multicolumn{3}{c}{robust SGNN} \\ \cline{3-14} 
 &  & SiNE & SGCN & SNEA & MSGNN & SDGNN & SE-SGformer & SGCL & GS-GNN & SBGCL & RSGNN & LRWSB & \textbf{\framework} \\ \hline \hline
\multirow{4}{*}{Bitcoin\_OTC} & 0 & 88.36{\tiny$\pm$4.27} & 94.17{\tiny$\pm$2.01} & 91.42{\tiny$\pm$1.07} & 95.37{\tiny$\pm$3.41} & 94.77{\tiny$\pm$3.16} & 95.49{\tiny$\pm$1.53} & 95.18{\tiny$\pm$0.89} & 95.23{\tiny$\pm$1.50} & \textbf{95.84{\tiny$\pm$1.18}} & 93.69{\tiny$\pm$0.82} & 94.39{\tiny$\pm$1.94} & \underline{94.75\tiny$\pm$0.65} \\
 & 10 & 81.01{\tiny$\pm$3.51} & 81.95{\tiny$\pm$2.10} & 86.95{\tiny$\pm$2.56} & 85.81{\tiny$\pm$4.73} & 87.52{\tiny$\pm$2.32} & 84.34{\tiny$\pm$1.24} & 88.21{\tiny$\pm$1.16} & 87.55{\tiny$\pm$2.75} & 89.01{\tiny$\pm$4.22} & 89.96{\tiny$\pm$0.52} & 90.03{\tiny$\pm$2.44} & \textbf{\underline{91.71{\tiny$\pm$0.81}}} \\
 & 20 & 75.31{\tiny$\pm$3.77} & 78.56{\tiny$\pm$2.38} & 75.81{\tiny$\pm$3.41} & 77.26{\tiny$\pm$2.45} & 76.15{\tiny$\pm$2.57} & 79.78{\tiny$\pm$1.33} & 80.77{\tiny$\pm$2.09} & 79.88{\tiny$\pm$1.34} & 83.94{\tiny$\pm$2.39} & 84.96{\tiny$\pm$0.76} & 85.61{\tiny$\pm$1.17} & \textbf{\underline{87.97{\tiny$\pm$1.34}}} \\
 & 25 & 73.39{\tiny$\pm$5.11} & 73.48{\tiny$\pm$2.31} & 71.31{\tiny$\pm$2.27} & 74.16{\tiny$\pm$3.30} & 72.33{\tiny$\pm$1.14} & 75.46{\tiny$\pm$1.18} & 75.33{\tiny$\pm$1.09} & 74.32{\tiny$\pm$1.91} & 78.67{\tiny$\pm$2.47} & 80.60{\tiny$\pm$0.71} & 82.42{\tiny$\pm$1.69} & \textbf{\underline{86.05{\tiny$\pm$2.46}}} \\ \hline
\multirow{4}{*}{Bitcoin\_Alpha} & 0 & 93.07{\tiny$\pm$3.78} & 94.43{\tiny$\pm$3.09} & 92.45{\tiny$\pm$2.16} & 95.81{\tiny$\pm$1.76} & 95.95{\tiny$\pm$2.84} & 95.75{\tiny$\pm$1.09} & 96.22{\tiny$\pm$2.30} & 95.70{\tiny$\pm$1.97} & \textbf{96.38{\tiny$\pm$1.65}} & 94.57{\tiny$\pm$1.02} & 94.62{\tiny$\pm$2.48} & \underline{95.18\tiny$\pm$0.44} \\
 & 10 & 82.01{\tiny$\pm$3.15} & 82.58{\tiny$\pm$2.78} & 81.33{\tiny$\pm$2.51} & 83.18{\tiny$\pm$2.33} & 82.96{\tiny$\pm$3.43} & 84.38{\tiny$\pm$0.82} & 87.23{\tiny$\pm$2.09} & 84.09{\tiny$\pm$2.52} & 87.64{\tiny$\pm$1.68} & 88.11{\tiny$\pm$1.16} & 88.57{\tiny$\pm$2.35} & \textbf{\underline{90.01{\tiny$\pm$2.18}}} \\
 & 20 & 75.42{\tiny$\pm$3.26} & 76.21{\tiny$\pm$2.31} & 74.22{\tiny$\pm$2.20} & 76.27{\tiny$\pm$3.31} & 77.18{\tiny$\pm$2.59} & 78.13{\tiny$\pm$0.99} & 79.21{\tiny$\pm$2.59} & 78.81{\tiny$\pm$1.89} & 84.09{\tiny$\pm$2.27} & 85.25{\tiny$\pm$0.59} & 85.33{\tiny$\pm$3.13} & \textbf{\underline{87.01{\tiny$\pm$1.44}}} \\
 & 25 & 71.33{\tiny$\pm$2.71} & 73.34{\tiny$\pm$2.20} & 71.25{\tiny$\pm$2.78} & 71.62{\tiny$\pm$2.76} & 73.85{\tiny$\pm$2.58} & 75.47{\tiny$\pm$1.03} & 76.22{\tiny$\pm$1.85} & 76.23{\tiny$\pm$1.92} & 83.48{\tiny$\pm$1.97} & 83.36{\tiny$\pm$0.74} & 83.18{\tiny$\pm$2.77} & \textbf{\underline{85.41{\tiny$\pm$1.84}}} \\ \hline
\multirow{4}{*}{Epinions} & 0 & 91.27{\tiny$\pm$3.55} & 92.43{\tiny$\pm$3.20} & 92.27{\tiny$\pm$2.29} & 93.58{\tiny$\pm$4.29} & 94.07{\tiny$\pm$3.25} & 90.19{\tiny$\pm$2.25} & 93.22{\tiny$\pm$2.11} & \textbf{95.44{\tiny$\pm$1.99}} & 94.26{\tiny$\pm$4.27} & 93.51{\tiny$\pm$0.85} & 93.43{\tiny$\pm$1.44} & \underline{94.57\tiny$\pm$0.16} \\
 & 10 & 82.72{\tiny$\pm$3.56} & 84.19{\tiny$\pm$4.19} & 83.22{\tiny$\pm$2.40} & 83.57{\tiny$\pm$2.01} & 84.74{\tiny$\pm$1.50} & 81.44{\tiny$\pm$1.87} & 87.49{\tiny$\pm$2.51} & 85.37{\tiny$\pm$1.09} & 87.53{\tiny$\pm$3.32} & 88.31{\tiny$\pm$0.65} & 87.86{\tiny$\pm$1.63} & \textbf{\underline{90.35{\tiny$\pm$0.54}}} \\
 & 20 & 75.44{\tiny$\pm$3.34} & 77.54{\tiny$\pm$2.47} & 78.32{\tiny$\pm$2.01} & 77.16{\tiny$\pm$3.31} & 79.87{\tiny$\pm$2.20} & 72.57{\tiny$\pm$1.94} & 80.10{\tiny$\pm$1.32} & 78.42{\tiny$\pm$1.02} & 82.82{\tiny$\pm$3.51} & 83.65{\tiny$\pm$1.23} & 81.55{\tiny$\pm$0.96} & \textbf{\underline{85.48{\tiny$\pm$0.51}}} \\
 & 25 & 72.19{\tiny$\pm$2.85} & 72.88{\tiny$\pm$2.43} & 73.11{\tiny$\pm$2.98} & 71.07{\tiny$\pm$2.52} & 73.31{\tiny$\pm$3.26} & 70.35{\tiny$\pm$1.58} & 76.33{\tiny$\pm$2.01} & 75.22{\tiny$\pm$2.03} & 80.28{\tiny$\pm$3.10} & 81.49{\tiny$\pm$1.04} & 80.35{\tiny$\pm$1.31} & \textbf{\underline{83.73{\tiny$\pm$0.82}}} \\ \hline
\multirow{4}{*}{Slashdot} & 0 & 85.23{\tiny$\pm$3.16} & 88.21{\tiny$\pm$3.05} & 86.46{\tiny$\pm$3.47} & 90.78{\tiny$\pm$3.81} & 89.40{\tiny$\pm$2.89} & 85.74{\tiny$\pm$1.24} & 89.51{\tiny$\pm$2.01} & \textbf{90.82{\tiny$\pm$2.07}} & 89.10{\tiny$\pm$2.47} & 89.32{\tiny$\pm$1.13} & 88.64{\tiny$\pm$2.29} & \underline{89.71\tiny$\pm$0.84} \\
 & 10 & 81.25{\tiny$\pm$3.74} & 80.17{\tiny$\pm$2.68} & 81.22{\tiny$\pm$2.19} & 80.29{\tiny$\pm$3.07} & 80.65{\tiny$\pm$2.28} & 80.38{\tiny$\pm$1.35} & 81.14{\tiny$\pm$2.09} & 82.89{\tiny$\pm$2.09} & 84.95{\tiny$\pm$1.73} & 85.37{\tiny$\pm$1.59} & 83.49{\tiny$\pm$1.74} & \textbf{\underline{86.35{\tiny$\pm$0.44}}} \\
 & 20 & 72.82{\tiny$\pm$3.79} & 73.22{\tiny$\pm$3.21} & 74.21{\tiny$\pm$3.53} & 73.57{\tiny$\pm$2.83} & 74.47{\tiny$\pm$2.35} & 72.37{\tiny$\pm$1.13} & 75.82{\tiny$\pm$2.09} & 74.33{\tiny$\pm$2.50} & 81.21{\tiny$\pm$2.09} & 80.87{\tiny$\pm$1.62} & 78.57{\tiny$\pm$1.48} & \textbf{\underline{83.52{\tiny$\pm$0.39}}} \\
 & 25 & 70.12{\tiny$\pm$2.16} & 71.54{\tiny$\pm$2.61} & 72.10{\tiny$\pm$2.50} & 72.07{\tiny$\pm$1.85} & 72.48{\tiny$\pm$3.48} & 69.01{\tiny$\pm$1.26} & 73.61{\tiny$\pm$1.94} & 72.31{\tiny$\pm$1.63} & 78.54{\tiny$\pm$2.11} & 79.81{\tiny$\pm$0.91} & 76.68{\tiny$\pm$2.15} & \textbf{\underline{81.04{\tiny$\pm$0.77}}} \\ \hline \hline
\end{tabular}%
}
\end{table*}

\noindent \textbf{Remark.} Training with this upper bound encourages the predicted labels to become independent of the noisy labels. However, it may also lead $\mathcal{P}_{\phi}$ to place reduced emphasis on the true label distribution, potentially causing the outputs to approximate a Bernoulli distribution and impairing the model to recover accurate labels.
To address this, the unified optimization guided by Eq.~\ref{eq:gib-td} maintains an effective balance between noise filtering and information retention. This integrated framework prevents the model from collapsing into trivial solutions, such as conforming to a Bernoulli distribution, and ensures that noisy labels are suppressed while meaningful patterns are preserved.

\begin{table}[!tp]
\centering
\caption{Statistics of the signed graph datasets.}
\label{tab:data-des}
\resizebox{\linewidth}{!}{%
\begin{tabular}{lcccc}
\hline 
Dataset       & \#Node & \#Edges & \#Positive Edges & \#Negative Edges \\\hline 
Bitcoin\_Alpha & 3,784  & 14,145  & 12,729           & 1,416            \\
Bitcoin\_OTC   & 5,901  & 21,522  & 18,390           & 3,132            \\
Epinions      & 16,992 & 327,227 & 276,309          & 50,918           \\
Slashdot      & 33,586 & 396,003 & 295,201          & 100,802    \\ \hline 
\end{tabular}%
}
\end{table}

\begin{proposition}
\label{prop:ub_hg}
(\textbf{Upper bound of $I(H; \tilde{\mathcal{G}})$}). For noisy graph $\tilde{\mathcal{G}}$ and the edge representation $H$, we have
\begin{equation}
I(H; \tilde{\mathcal{G}}) \leq \iint \mathbb{P}(H, \tilde{\mathcal{G}})\log\frac{\mathbb{P}(H|\tilde{\mathcal{G}})}{\mathbb{P}_{\upsilon}(H)}dHd\tilde{\mathcal{G}},
\label{eq:ub_hg}
\end{equation}
where $\mathbb{P}_{\upsilon}(H)$ is variational approximation to prior distribution $\mathbb{P}(H)$.
\end{proposition}

\begin{proof}
Formally, $I(H; \tilde{\mathcal{G}})$ can be expanded via mutual information definition:
\begin{equation}
I(H; \tilde{\mathcal{G}}) =  \iint \mathbb{P}(H, \tilde{\mathcal{G}})\log\frac{\mathbb{P}(H|\tilde{\mathcal{G}})}{\mathbb{P}(H)}dHd\tilde{\mathcal{G}}.
\label{eq:ihg}
\end{equation}

In general, computing the distribution $\mathbb{P}(H) = \int \mathbb{P}(H|\tilde{\mathcal{G}})\mathbb{P}(\tilde{\mathcal{G}})d\tilde{\mathcal{G}}$ is very difficult, so we use $\mathbb{P}_{\upsilon}(H)$ as the variational approximation to $\mathbb{P}(H)$. Since the KL divergence $\mathcal{D}_{KL}(\mathbb{P}(S)||\mathbb{P}_{\upsilon}(S))\geq 0$,
\begin{align}
\mathcal{D}_{KL}(\mathbb{P}(S)||\mathbb{P}_{\upsilon}(S)) & \geq 0 \Rightarrow \nonumber \\ 
 \int \mathbb{P}(s)\log \mathbb{P}(s)ds & \geq \int \mathbb{P}(s)\log \mathbb{P}_{\upsilon}(s)ds.
\label{eq:dkl}
\end{align}

Taking Eq.\ref{eq:ihg} and Eq.\ref{eq:dkl} together:
\begin{equation}
I(H; \tilde{\mathcal{G}}) \leq \iint \mathbb{P}(H, \tilde{\mathcal{G}})\log\frac{\mathbb{P}(H|\tilde{\mathcal{G}})}{\mathbb{P}_{\upsilon}(H)}dHd\tilde{\mathcal{G}}.
\end{equation}

Then, we have the upper bound for $I(H; \tilde{\mathcal{G}})$.
\end{proof}

For the information term $I(H; \tilde{\mathcal{G}})$ in Eq.\ref{eq:ub_hg}, we consider a parametric Gaussian distribution as prior $\mathbb{P}_{\upsilon}(H)$ and $\mathbb{P}(H|\tilde{\mathcal{G}})$ to allow an analytic computation of KL divergence:
\begin{equation}
\mathbb{P}_{\upsilon}(H) = \mathcal{N}(\mu_0, \sigma_0), \quad
\mathbb{P}(H|\tilde{\mathcal{G}}) = \mathcal{N}(f_{sgnn}^\mu(\mathcal{G}_c), f_{sgnn}^\sigma(\mathcal{G}_c)),
\end{equation}
where $\mu\in\mathbb{P}_{\upsilon}^h$ is the mean vector and $\sigma\in\mathbb{P}_{\upsilon}^h$ denotes the diagonal co-variance matrix of $H$ encoded by $f_{sgnn}(\mathcal{G}_c)$. 
The first half outputs of $H$ encode $\mu$ and the remaining half outputs encode $\sigma$. We adopt a softplus transform for $f_{sgnn}^\sigma(\mathcal{G}_c)$ to ensure the non-negativity. At last, we treat $\mathbb{P}_{\upsilon}(H)$ as a fixed $d$-dimensional spherical Gaussian $\mathbb{P}_{\upsilon}(H) = \mathcal{N}(H|0, \text{I})$ like~\cite{alemi2016deep}.

\noindent \textbf{Optimization.} With the derived bounds in Props.~\ref{prop:cls}, \ref{prop:ub_hg} and \ref{prop:up_ycy}, the relaxation is then conducted on the three mutual information terms in Eq.~\ref{eq:gib-td}. Correspondingly, the final training loss for minimization is formulated as:
\begin{equation}
\mathcal{L} = \mathcal{L}_{cls} + \alpha\mathcal{L}_{KL}^{Y} + \beta\mathcal{L}_{KL}^{\mathcal{G}},
\label{eq:final_loss}
\end{equation}
where $\alpha$ and $\beta$ are two hyperparameters that control the loss balance, $\mathcal{L}_{cls}$ is the standard cross-entropy loss, $\mathcal{L}_{KL}^{Y} = \text{KL}(\mathbb{P}_\phi(Y_c|Y)$ $||\mathbb{Q}(Y_c))$, and $\mathcal{L}_{KL}^{G} = \text{KL}(\mathbb{P}(H|\tilde{\mathcal{G}})||\mathbb{P}_{\upsilon}(H))$. $\text{KL}(\cdot||\cdot)$ is the KL divergence.
The overall training process is outlined in Algorithm ~\ref{code:design}.~\footnote{Analyses regarding runtime complexity and efficiency are presented in Appendix D.}

\begin{table*}[!tp]
\centering
\caption{\textbf{Variants of \framework with different backbones.} \textbf{Bold} means the best performance within the same backbone. Std. is omitted for simplicity. F1 is the abbreviation of Binary-F1.}
\label{tab:backbones}
\resizebox{\textwidth}{!}{%
\begin{tabular}{l|c|cccccccc|cccccccc}
\hline \hline
\multirow{2}{*}{Dataset} & \multirow{2}{*}{Noise (\%)} & \multicolumn{2}{c}{SGCN} & \multicolumn{2}{c}{RSGNN {\tiny(SGCN)}} & \multicolumn{2}{c}{LRWSE {\tiny(SGCN)}} & \multicolumn{2}{c|}{\textbf{\framework} {\tiny(SGCN)}} & \multicolumn{2}{c}{SNEA} & \multicolumn{2}{c}{RSGNN {\tiny(SNEA)}} & \multicolumn{2}{c}{LRWSE {\tiny(SNEA)}} & \multicolumn{2}{c}{\textbf{\framework} {\tiny(SNEA)}} \\ \cline{3-18} 
 &  & AUC & F1 & AUC & F1 & AUC & F1 & AUC & F1 & AUC & F1 & AUC & F1 & AUC & F1 & AUC & F1 \\ \hline \hline
\multirow{4}{*}{Bitcoin\_OTC} & 0 & 84.56 & 94.17 & 85.01 & 93.69 & 86.11 & 94.39 & \textbf{86.51} & \textbf{94.75} & 86.10 & 91.42 & \textbf{87.32} & 92.35 & 86.97 & 93.13 & 87.19 & \textbf{94.23} \\
 & 10 & 78.43 & 81.95 & 82.00 & 89.96 & 81.98 & 90.03 & \textbf{82.48} & \textbf{91.71} & 77.32 & 86.95 & 80.32 & 89.21 & 79.78 & 89.41 & \textbf{81.96} & \textbf{90.53} \\
 & 20 & 74.33 & 78.56 & 77.77 & 84.96 & 77.92 & 85.61 & \textbf{78.39} & \textbf{87.97} & 71.72 & 75.81 & 75.69 & 83.26 & 75.86 & 83.85 & \textbf{77.31} & \textbf{86.84} \\
 & 25 & 69.65 & 73.48 & 74.66 & 80.60 & 75.04 & 82.42 & \textbf{76.57} & \textbf{86.05} & 68.32 & 71.31 & 73.26 & 80.20 & 74.03 & 81.33 & \textbf{75.68} & \textbf{84.20} \\ \hline
\multirow{4}{*}{Bitcoin\_Alpha} & 0 & 84.30 & 94.43 & 83.47 & 94.57 & 84.74 & 94.62 & \textbf{85.98} & \textbf{95.18} & 85.10 & 92.45 & 83.47 & 93.34 & 84.67 & \textbf{93.73} & \textbf{85.42} & 93.56 \\
 & 10 & 75.49 & 82.58 & 77.68 & 88.11 & 77.39 & 88.57 & \textbf{79.57} & \textbf{90.01} & 74.92 & 81.33 & 78.27 & 86.23 & 77.92 & 86.55 & \textbf{79.66} & \textbf{88.21} \\
 & 20 & 71.57 & 76.21 & 72.15 & 85.25 & 72.48 & 85.33 & \textbf{75.90} & \textbf{87.01} & 70.37 & 74.22 & 72.69 & 83.24 & 73.27 & 83.63 & \textbf{75.23} & \textbf{87.20} \\
 & 25 & 69.57 & 73.34 & 70.82 & 83.36 & 70.60 & 83.18 & \textbf{73.36} & \textbf{85.41} & 67.32 & 71.25 & 70.33 & 80.32 & 69.90 & 80.06 & \textbf{72.20} & \textbf{85.29} \\ \hline
\multirow{4}{*}{Epinions} & 0 & 79.81 & 92.43 & 80.35 & 93.51 & 80.53 & 93.43 & \textbf{80.82} & \textbf{94.57} & 85.47 & 92.27 & 85.33 & 92.78 & 85.57 & 93.24 & \textbf{88.12} & \textbf{94.24} \\
 & 10 & 75.70 & 84.19 & 77.56 & 88.31 & 77.15 & 87.86 & \textbf{78.94} & \textbf{90.35} & 80.66 & 83.22 & 82.39 & 88.32 & 81.85 & 87.68 & \textbf{84.86} & \textbf{90.18} \\
 & 20 & 72.12 & 77.54 & 74.02 & 83.65 & 73.32 & 81.55 & \textbf{76.18} & \textbf{85.48} & 72.17 & 78.32 & 78.05 & 84.62 & 77.64 & 82.73 & \textbf{81.00} & \textbf{86.73} \\
 & 25 & 69.53 & 72.88 & 71.39 & 81.49 & 70.84 & 80.35 & \textbf{73.89} & \textbf{83.73} & 68.91 & 73.11 & 74.22 & 80.49 & 73.71 & 80.09 & \textbf{76.34} & \textbf{84.65} \\ \hline
\multirow{4}{*}{Slashdot} & 0 & 85.83 & 88.21 & 85.27 & 89.32 & 85.02 & 88.64 & \textbf{86.82} & \textbf{89.71} & 80.12 & 86.46 & 80.03 & \textbf{87.21} & 80.49 & 86.89 & \textbf{82.73} & 87.13 \\
 & 10 & 77.32 & 80.17 & 82.09 & 85.37 & 81.39 & 83.49 & \textbf{84.18} & \textbf{86.35} & 75.33 & 81.22 & 77.48 & 83.21 & 76.86 & 82.12 & \textbf{79.18} & \textbf{84.33} \\
 & 20 & 73.20 & 73.22 & 75.31 & 80.87 & 74.76 & 78.57 & \textbf{79.19} & \textbf{83.52} & 71.09 & 74.21 & 73.28 & 80.77 & 73.63 & 79.31 & \textbf{73.84} & \textbf{81.32} \\
 & 25 & 70.16 & 71.54 & 71.54 & 79.81 & 70.51 & 76.68 & \textbf{74.99} & \textbf{81.04} & 68.51 & 72.10 & 71.99 & 78.42 & 70.91 & 76.64 & \textbf{72.78} & \textbf{80.12} \\ \hline \hline
\end{tabular}%
}
\end{table*}

\section{Experiments}
\label{sec:exp}
Comprehensive experiments are conducted below to assess the robustness of \framework in mitigating varying levels of noise with the aim of predicting link sign.

\subsection{Experiment Setup}
\noindent \textbf{Datasets.} We evaluate our method on four widely-used real-world signed graph datasets: Bitcoin\_OTC, Bitcoin\_Alpha, Epinions, and Slashdot. Table~\ref{tab:data-des} summarizes the key statistics of these datasets. More details are in Appendix C.1.

\noindent \textbf{Random noise.} Following RSGNN~\cite{zhang2023rsgnn}, we simulate noisy environments via randomly flipping a specified percentage of edge signs. The noise ratio is varied between 10\% and 25\% to simulate different levels of perturbation. Experiments regarding adversarial noise and alternative types of random noise, such as random deletion and addition, can be found in Appendix E and Appendix F.

\noindent \textbf{Task and evaluation metrics.} Consistent with prior work in signed graph representation learning, we evaluate methods with link sign prediction, aiming to infer the signs of unseen links. Edges are split randomly into training and testing parts by 8:2.
Two standard metrics, Area Under the Curve (AUC) and Binary-F1 score, are adopted as performance metrics. Higher scores indicate better prediction performance.

\noindent \textbf{Baselines.} Three kinds of methods are adopted for comparison: \textit{a.} basic SGNN for signed graph encoding: SiNE~\cite{wang2017signed}, SGCN~\cite{derr2018signed}, SNEA~\cite{li2020learning}, SDGNN~\cite{huang2021sdgnn}, MSGNN~\cite{he2022msgnn} and SE-SGformer~\cite{li2025se}; \textit{b.} signed graph learning methods with noise-aware representation: SGCL~\cite{shu2021sgcl}, GS-GNN~\cite{liu2021signed} and SBGCL~\cite{zhang2023contrastive}; \textit{c.} robust signed graph neural networks: RSGNN\cite{zhang2023rsgnn} and LRWSB~\cite{minici2024link}.
More details are presented in Appendix C.2.

\noindent \textbf{Hyper-parameters.} 
We follow the hyper-parameter configuration of previous SGNNs.
Hidden dimension is set to 64. 
SGCN with 4 message aggregation layers is adopted as the signed graph encoder, the collaboration of \framework with other SGNNs are shown in Sec.~\ref{sec:results}.
The constant $\tau$ in $\mathcal{L}_{KL}^Y$ is set to $0.8$.
The values of $\alpha$ and $\beta$ that control the loss balance is selected from $10^{\text{-}3}$ to $10^{3}$.
Fixing the learning rate to $0.01$, optimization is performed with Adam optimizer~\cite{kingma2015adam} for 1,000 training epochs.
Since these benchmark datasets do not come with built-in node features, we follow prior studies~\cite{derr2018signed,zhang2023rsgnn} and use Truncated-SVD~\cite{halko2011finding} on the graph structure to generate the initial node embeddings.

\begin{table}[!tp]
\centering
\caption{\textbf{Ablation study.}}
\label{tab:ablation}
\resizebox{1\linewidth}{!}{%
\begin{tabular}{l|l|c|cccc}
\hline \hline
Dataset & Metric & Noise (\%) & \framework & w/o $\mathcal{L}_{KL}^Y$ & w/o $\mathcal{L}_{KL}^\mathcal{G}$ & w/o $\mathcal{L}_{KL}^Y$ \& $\mathcal{L}_{KL}^\mathcal{G}$ \\ \hline \hline
\multirow{6}{*}{Bitcoin\_OTC} & \multirow{3}{*}{AUC} & 10 & 82.48$\pm$0.44 & 81.46$\pm$0.52 & 81.38$\pm$0.54 & 81.20$\pm$0.63 \\
 &  & 20 & 78.39$\pm$0.44 & 76.75$\pm$0.35 & 77.30$\pm$0.27 & 76.41$\pm$0.43 \\
 &  & 25 & 76.57$\pm$0.56 & 73.90$\pm$0.66 & 74.34$\pm$0.63 & 73.13$\pm$1.03 \\ \cline{2-7} 
 & \multirow{3}{*}{Binary-F1} & 10 & 91.71$\pm$0.81 & 90.93$\pm$0.88 & 90.46$\pm$1.44 & 89.64$\pm$1.07 \\
 &  & 20 & 87.97$\pm$1.34 & 86.55$\pm$1.38 & 86.97$\pm$.141 & 84.34$\pm$1.96 \\
 &  & 25 & 86.05$\pm$2.46 & 83.99$\pm$0.90 & 84.37$\pm$1.50 & 81.53$\pm$2.33 \\ \hline
\multirow{6}{*}{Bitcoin\_Alpha} & \multirow{3}{*}{AUC} & 10 & 79.57$\pm$0.56 & 79.08$\pm$0.41 & 78.99$\pm$0.32 & 78.23$\pm$0.56 \\
 &  & 20 & 75.90$\pm$0.96 & 73.32$\pm$0.73 & 74.72$\pm$0.62 & 72.36$\pm$0.37 \\
 &  & 25 & 73.36$\pm$0.92 & 70.56$\pm$1.46 & 72.06$\pm$0.58 & 70.50$\pm$0.93 \\ \cline{2-7} 
 & \multirow{3}{*}{Binary-F1} & 10 & 90.01$\pm$2.18 & 89.21$\pm$1.55 & 88.17$\pm$2.19 & 87.48$\pm$2.49 \\
 &  & 20 & 87.01$\pm$1.44 & 82.65$\pm$2.97 & 84.33$\pm$1.93 & 83.63$\pm$2.12 \\
 &  & 25 & 85.41$\pm$1.84 & 82.55$\pm$3.62 & 83.35$\pm$1.51 & 81.85$\pm$1.29 \\ \hline
\multirow{6}{*}{Epinions} & \multirow{3}{*}{AUC} & 10 & 78.94$\pm$0.27 & 76.86$\pm$0.19 & 77.41$\pm$0.31 & 76.75$\pm$0.16 \\
 &  & 20 & 76.18$\pm$0.20 & 74.76$\pm$0.19 & 74.48$\pm$0.25 & 74.06$\pm$0.37 \\
 &  & 25 & 73.89$\pm$0.50 & 71.67$\pm$0.28 & 72.68$\pm$0.11 & 70.74$\pm$0.20 \\ \cline{2-7} 
 & \multirow{3}{*}{Binary-F1} & 10 & 90.35$\pm$0.54 & 89.77$\pm$0.31 & 88.23$\pm$0.57 & 89.30$\pm$0.53 \\
 &  & 20 & 85.48$\pm$0.51 & 83.18$\pm$0.59 & 84.66$\pm$0.39 & 82.79$\pm$0.47 \\
 &  & 25 & 83.73$\pm$0.82 & 81.50$\pm$0.77 & 81.74$\pm$1.03 & 80.30$\pm$0.72 \\ \hline
\multirow{6}{*}{Slashdot} & \multirow{3}{*}{AUC} & 10 & 84.18$\pm$0.17 & 81.78$\pm$0.08 & 83.11$\pm$0.16 & 82.13$\pm$0.20 \\
 &  & 20 & 79.19$\pm$0.23 & 77.73$\pm$0.29 & 77.00$\pm$0.13 & 76.67$\pm$0.18 \\
 &  & 25 & 74.99$\pm$0.14 & 73.47$\pm$0.37 & 74.53$\pm$0.24 & 72.30$\pm$0.19 \\ \cline{2-7} 
 & \multirow{3}{*}{Binary-F1} & 10 & 86.35$\pm$0.44 & 84.06$\pm$0.28 & 85.44$\pm$0.42 & 84.22$\pm$0.29 \\
 &  & 20 & 83.52$\pm$0.39 & 82.55$\pm$0.55 & 81.67$\pm$0.31 & 80.59$\pm$0.86 \\
 &  & 25 & 81.04$\pm$0.77 & 79.61$\pm$1.16 & 79.52$\pm$0.79 & 78.22$\pm$0.65 \\ \hline \hline
\end{tabular}%
}
\end{table}

\subsection{Results}
\label{sec:results}
The results of \framework along with three kinds of baselines are given by Table~\ref{tab:main_result}. Binary-F1 is exclusively reported as the chosen metric, as other metrics (i.e., AUC) exhibit analogous trends. The key insights emerge from the results: \textbf{a.} When evaluated on the original graph, \framework, while not attaining peak performance, demonstrates a noteworthy enhancement compared to its signed graph encoder, SGCN. Furthermore, in contrast to robust SGNNs (i.e., RSGNN and LRWSE), which employ an identical signed graph encoder, \framework consistently exhibits substantial performance improvements. \textbf{b.} After introducing noise, \framework, in general, exhibits superior performance contrast to all other methods. Specifically, \framework effectively mitigates the adverse effects of noise, yielding a performance gain ranging from 6.16\% to 12.57\% in comparison to SGCN.
Furthermore, in a direct comparison with robust SGNN, \framework achieves up to a 5.45\% improvement in Binary-F1 over RSGNN and up to a 4.95\% improvement over LRWSB.
These results collectively underscore the robustness of our information theory-based method, consistently yielding enhanced performance in various noisy scenarios.

Although our method shows a minor trade-off in performance under noise-free conditions, this compromise leads to significant gains when noise is introduced. Notably, our approach is orthogonal to other noise-tolerant SGNNs, and an interesting avenue for future research lies in combining our method with these existing techniques, which could further enhance performance, irrespective of the presence of noise.

\noindent \textbf{Variants of \framework.} 
As discussed in experiment setup, besides SGCN, our proposed \framework exhibits compatibility with various signed graph encoders. This aspect is further investigated to assess the collaborative capabilities of \framework with other SGNNs. Table~\ref{tab:backbones} reports the link sign prediction performance of our method compared with other robust SGNN models, wherein \framework collaborates with SGCN~\cite{derr2018signed} and SNEA~\cite{li2020learning}. The collaboration outcomes of \framework demonstrate a consistent suppression of the respective backbones, observed on both original and noise graphs. This not only affirms the robustness of our method but also underscores the efficacy of \framework in the realm of signed graph representation learning. Compared to the balance theory-based robust SGNNs (i.e., RSGNN and LRWSE), our method consistently outperforms, reinforcing the superiority of GIB in robust signed graph representation learning.

\begin{figure}[!tp]
  \centering
  \begin{subfigure}{\linewidth}
    \centering
    \includegraphics[width=\linewidth]{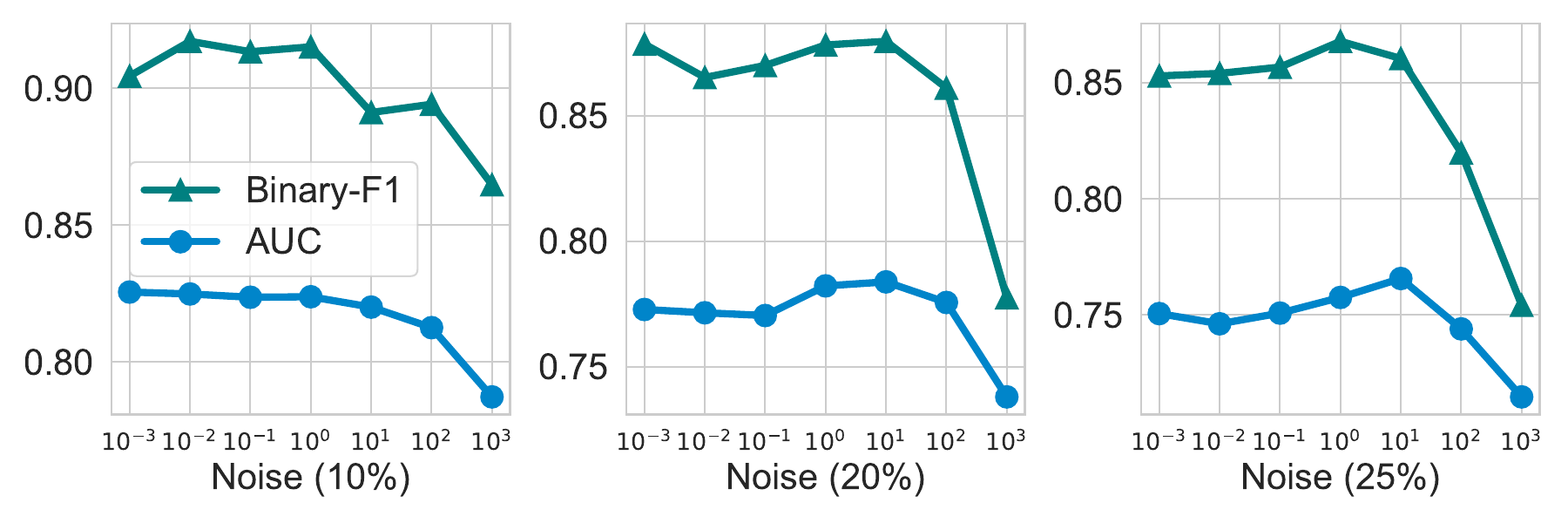}
    \vskip -5pt
    \caption{Parameter $\alpha$.}
    \label{fig:hyper-otc-alpha}
  \end{subfigure} \\
  \begin{subfigure}{\linewidth}
    \centering
    \includegraphics[width=\linewidth]{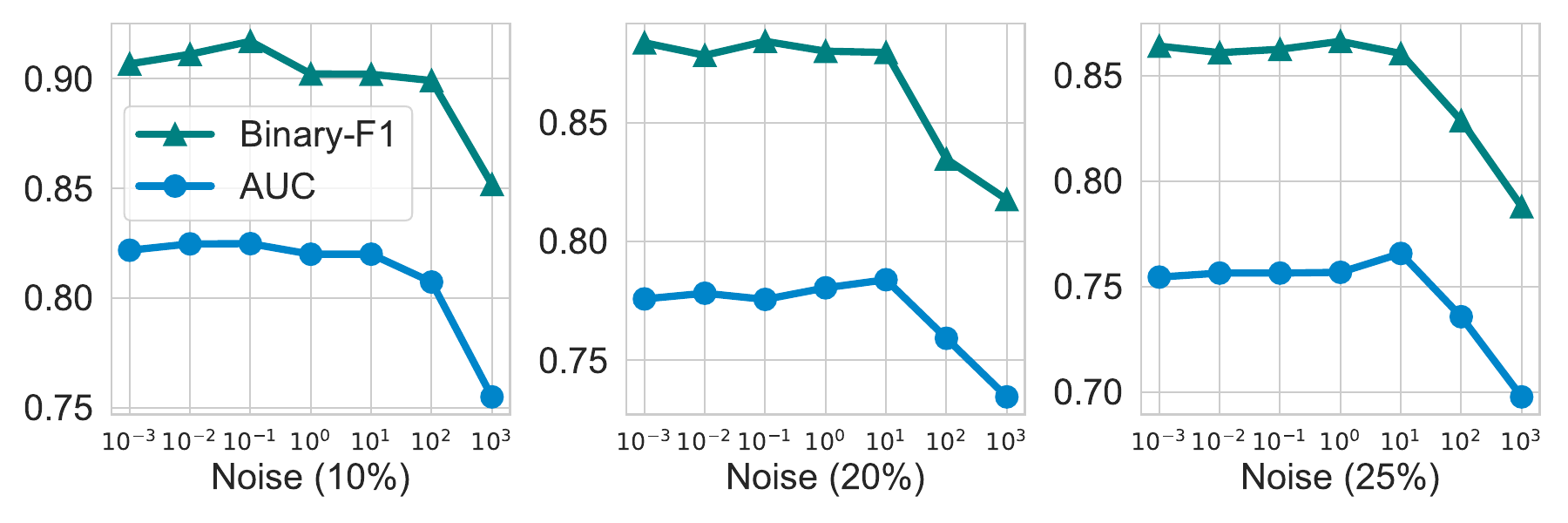}
    \vskip -5pt
    \caption{Parameter $\beta$.}
    \label{fig:hyper-otc-beta}
  \end{subfigure} 
  % \vskip -5pt
  \caption{\textbf{Hyper-parameter sensitivity on Bitcoin\_OTC.}}
  \label{fig:hyper-otc}
\end{figure}

\noindent \textbf{Ablation study.} In contrast to general GSL, our method achieves robust link sign prediction by extrapolating from the GIB theory. To analyze the impact of mutual information terms introduced by GIB-TD in noise combating, specifically $\mathcal{L}_{KL}^Y$ and $\mathcal{L}_{KL}^\mathcal{G}$, we perform ablation studies across the benchmark datasets. 
Same dataset split in prior experiments is adopted, using 80\% of the edges for training and 20\% for testing, with random noise applied only to the training portion.
Evaluation is conducted using AUC and Binary-F1 scores, and the results are reported in Table~\ref{tab:ablation}.
We observe that removing different components results in varied levels of performance degradation. The decline is most significant when both loss terms are removed, highlighting the critical role of these information terms in enhancing robustness.

\noindent \textbf{Hyper-parameter sensitivity.} 
Here, we assess the model performance under varying hyper-parameters $\alpha$ and $\beta$, which balance information extraction in Eq.~\ref{eq:final_loss}. During the experiments, we maintain all other parameters constant, aligning them with the parameters that yield the primary results in Table~\ref{tab:main_result}, while adjusting $\alpha$ or $\beta$.
Fig.~\ref{fig:hyper-otc} illustrates the trend of two metrics on the Bitcoin\_OTC datasets under various random noise ratios.
From the reported outcomes, we identify several key observations:
\textit{a.} Generally, $\alpha$ and $\beta$ exhibit stable performance when their values are not excessively large. This observation suggests that an increase in $\alpha$ and $\beta$ leads to over-compression of information from the input and target, thereby negatively impacting the model's performance.
\textit{b.} As the noise ratio increases, the optimal values of $\alpha$ and $\beta$ for peak performance of \framework shift towards the left. This trend indicates that higher levels of noise necessitate more aggressive information compression to extract useful information.

\begin{figure}[!tp]
  \centering
  \begin{subfigure}{.48\linewidth}
    \centering
    \includegraphics[width=\linewidth]{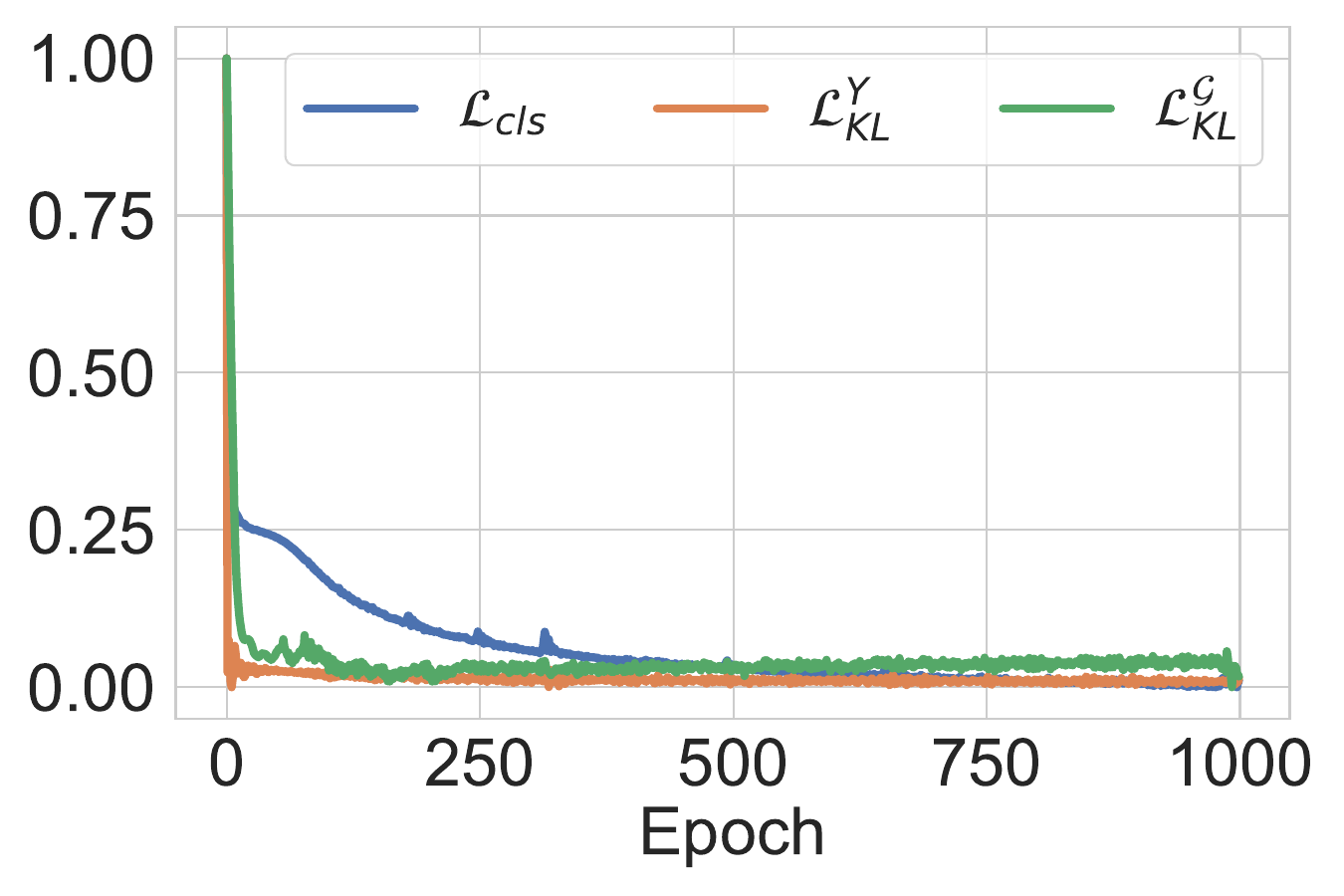}
    \caption{Bitcoin\_OTC.}
    \label{fig:loss-bit-otc}
  \end{subfigure} 
  \begin{subfigure}{.48\linewidth}
    \centering
    \includegraphics[width=\linewidth]{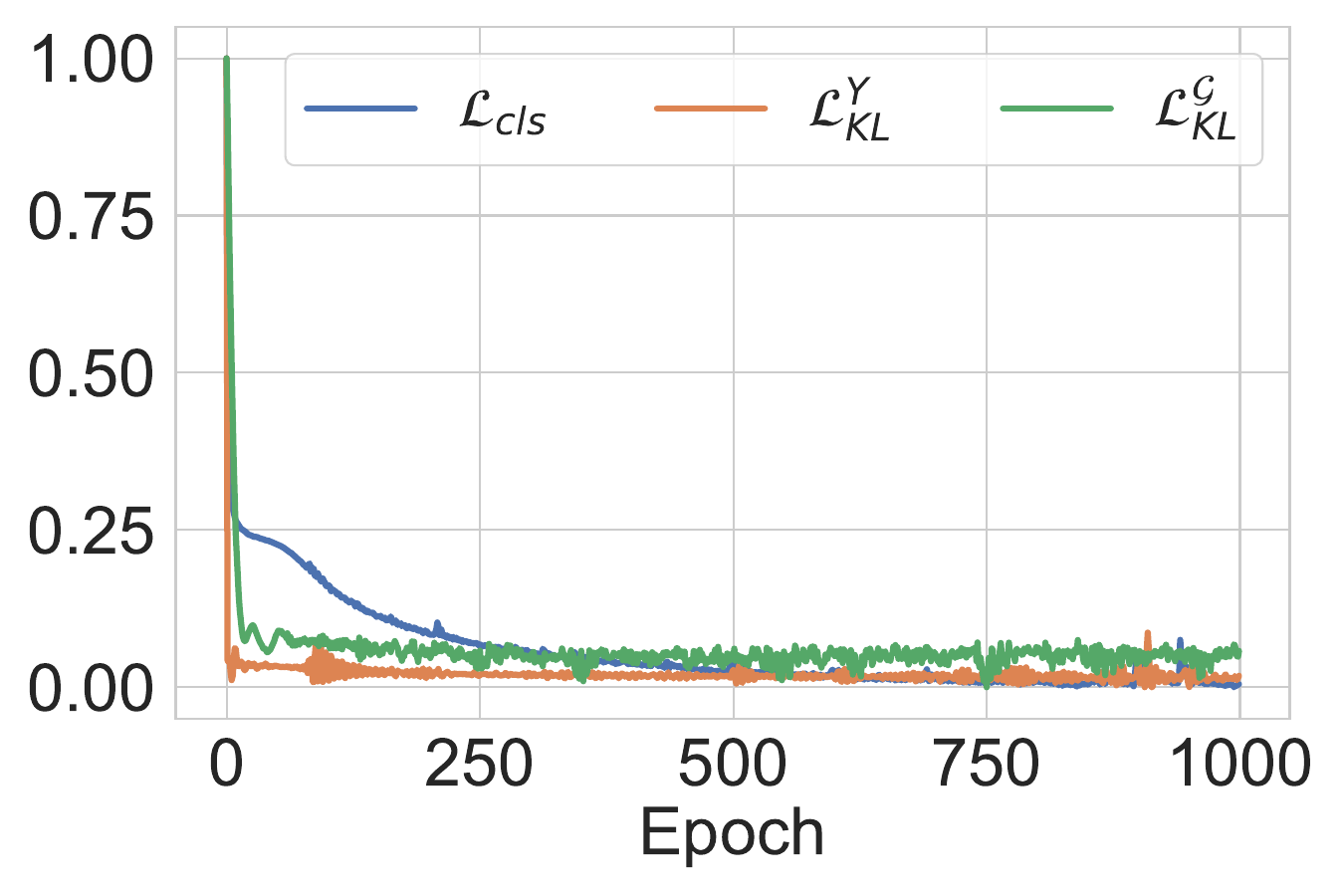}
    \caption{Bitcoin\_Alpha.}
    \label{fig:loss-bit-alpha}
  \end{subfigure} 
  \vskip -5pt
  \caption{\textbf{Loss curves of \framework with 25\% random noise.} Due to the scale difference, we normalize the curves to 0-1. }
  \label{fig:loss_curve}
\end{figure}
  
\begin{figure}[!tp]
  \centering
  \begin{subfigure}{.48\linewidth}
    \centering
    \includegraphics[width=\linewidth]{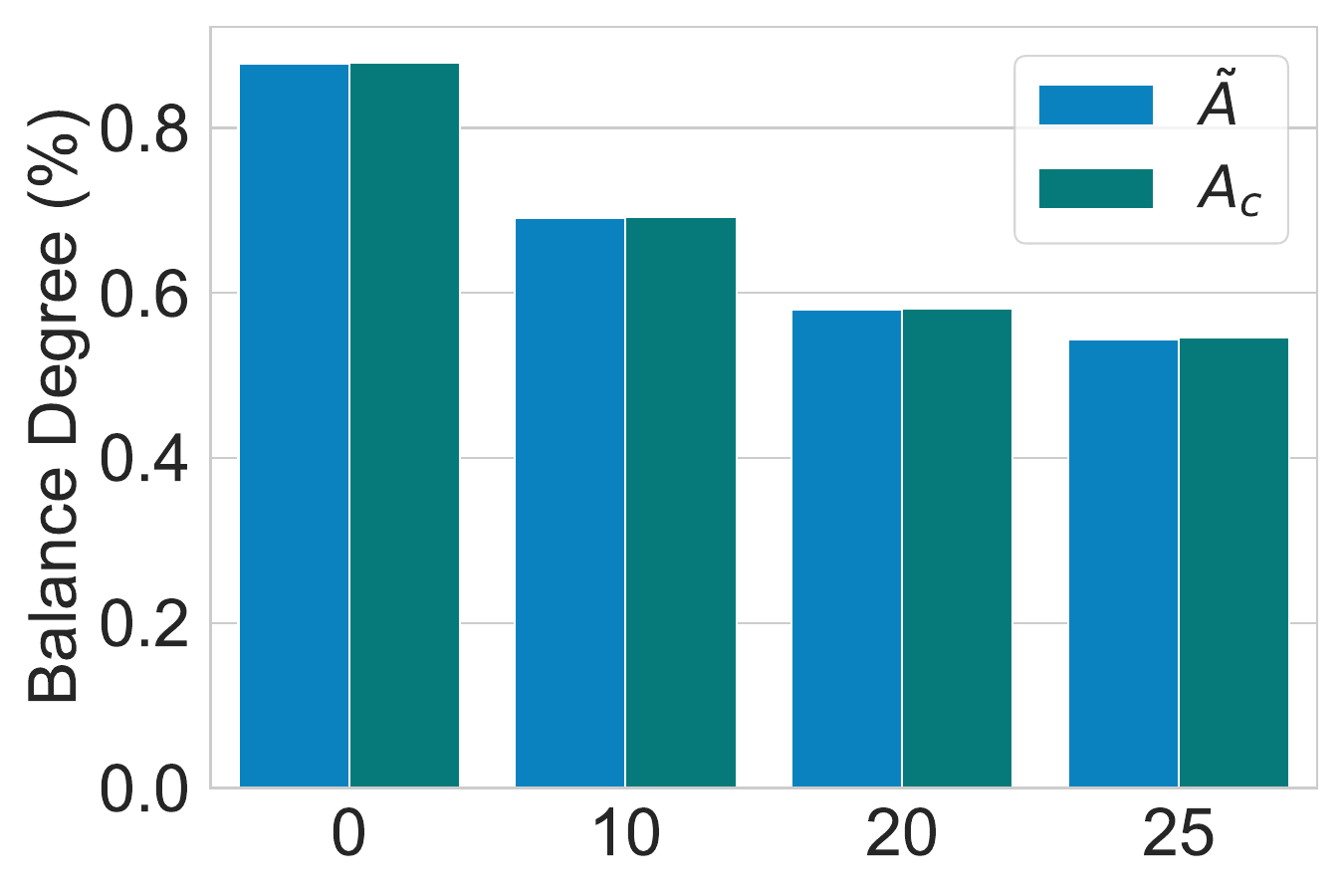}
    \caption{Bitcoin\_OTC.}
    \label{fig:bd-bit-otc}
  \end{subfigure} 
  \begin{subfigure}{.48\linewidth}
    \centering
    \includegraphics[width=\linewidth]{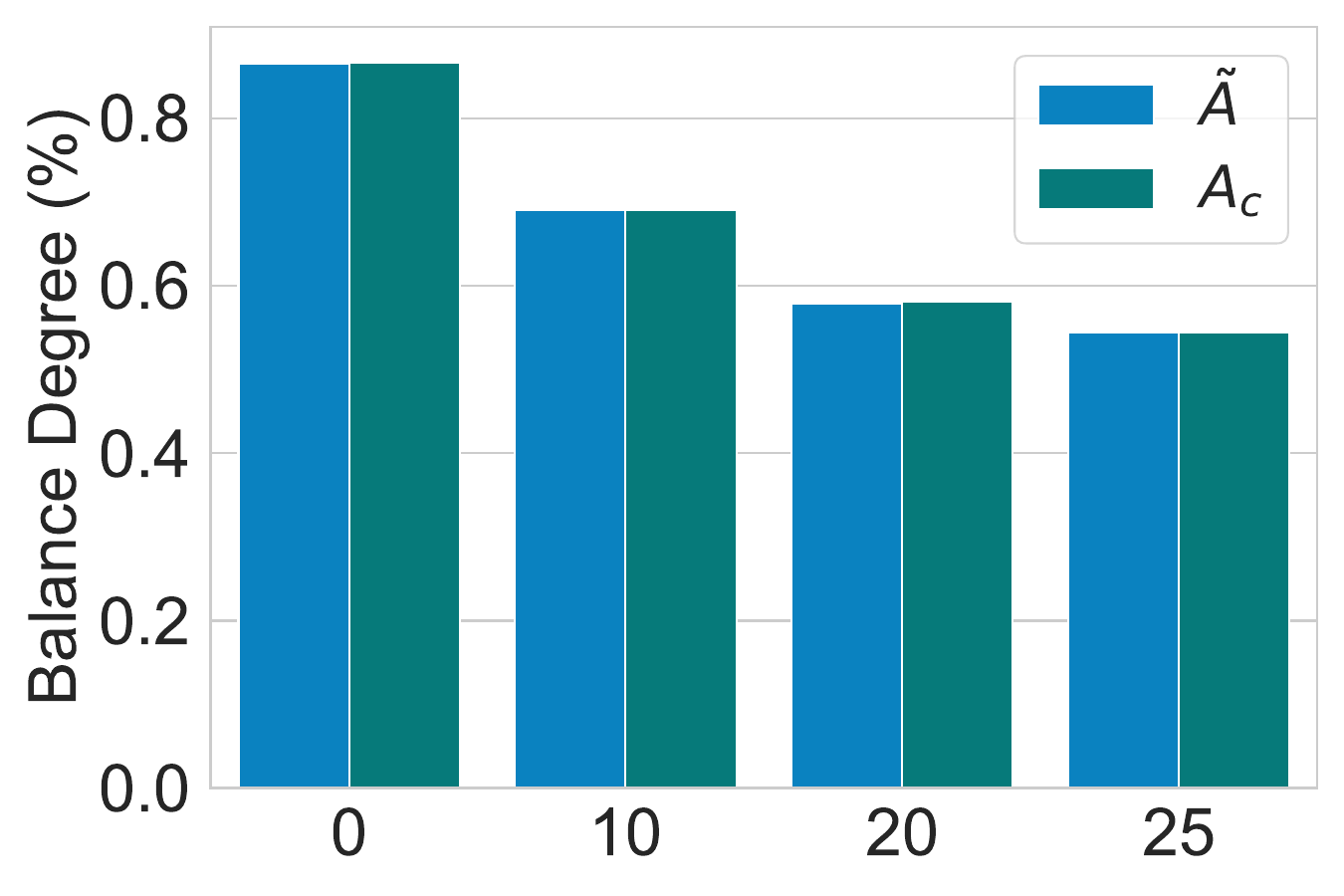}
    \caption{Bitcoin\_Alpha.}
    \label{fig:bd-bit-alpha}
  \end{subfigure} 
  \vskip -5pt
  \caption{\textbf{The balance degrees under random noise.} $\tilde{A}$ indicates the input graph structure. $A_c$ means the graph topology after \framework learning.}
  \label{fig:balance_degree}
\end{figure}

\noindent \textbf{Training stability.} 
Here, we analyze the convergence of \framework on Bitcoin\_OTC and Bitcoin\_Alpha with 1,000 training epochs. 
Due to the scale difference, we normalize the training losses to the range of 0-1.
Fig.~\ref{fig:loss_curve} depicts the learning curves of \framework (i.e., $\mathcal{L}_{cls}$, $\mathcal{L}_{KL}^{Y}$ and $\mathcal{L}_{KL}^{\mathcal{G}}$), where the learning processes are generally stable. 

\noindent \textbf{Do we need a more balanced graph for proper signed graph representation learning?} TThe structural balance of a signed graph can be quantified using the balance degree~\cite{aref2018measuring}, $D_3(\mathcal{G}) = \mathcal{O}^+_3 / \mathcal{O}_3 = (\text{Tr}(A^3) + \text{Tr}(|A|^3)) / 2\text{Tr}(|A|^3)$, where $\mathcal{O}^+_3$ and $\mathcal{O}_3$ counts the balanced and total triangles, respectively. 
In RSGNN, the introduction of noise edges negatively impacts the balance degree and subsequently the model performance. The authors attribute this degradation to the disruption of triangle structures, concluding that \textit{``random noise can increase the rate of unbalanced triangles, and then worsen the performance of SGNNs''}. To counteract this effect, they encourage the model to infer an alternative adjacency matrix with enhanced structural balance. 
In this context, we compute the balance degrees of the graph structures both before and after the \framework learning. As shown in Fig.~\ref{fig:balance_degree}, the balance degrees of $\tilde{A}$ and $A_c$ exhibit nearly identical values. Nevertheless, the results of \framework presented in Table~\ref{tab:backbones} reveal a noteworthy enhancement after training from $\tilde{A}$ and producing $A_c$, which suggests that achieving proper signed graph representation learning may not hinge on utilizing a more balanced graph.

In particular, as summarized in Table~\ref{tab:unbalanced_tri}, the four original datasets contain a substantial number of unbalanced triangles. This observation helps explain the suboptimal performance of robustness methods grounded in balance theory. In other words, forcing the learned graph to be overly balanced may distort the data and shift it from the true distribution.

\section{Conclusion}
This study addresses the challenge of robust learning on signed graphs by proposing \textit{\framework}, a novel approach that simultaneously denoises both the graph structure and label space. 
Instead of relying on the prevalent balance theory for signed graphs, which is often overly idealistic for real-world data,
we build our framework upon the GIB theory to achieve robustness under various levels of noise.
Extending the basic GIB formulation, \framework is capable of jointly denoising both the input structure and the supervision signals from noisy target spaces.
For training, \framework employs a data reparameterization mechanism and variational approximation to derive a tractable objective, facilitating the learning of effective representations from contaminated signed graphs.
We assess our approach on four benchmark datasets, where it consistently exhibits strong robustness under varying noise intensities. 
While our focus is on leveraging GIB for this task, future work may benefit from investigating alternative graph structure learning strategies or lightweight denoising heuristics.

\bibliographystyle{ACM-Reference-Format}
\balance
\bibliography{my_ref}

\clearpage
\appendix

\setcounter{table}{0}
\setcounter{figure}{0}
\renewcommand{\thetable}{A.\arabic{table}}
\renewcommand{\thefigure}{A.\arabic{figure}}

\section{Notation Table}
As an expansion of the notations in Sec.3, we summarize the frequently used notations in Table~\ref{tab:notations}.

\begin{table}[!hp]
\centering
\caption{The most frequently used notations in this paper.}
\vspace{-10pt}
\label{tab:notations}
\resizebox{\linewidth}{!}{
\begin{tabular}{c|c}
\hline \hline
Notations                                                      & Descriptions                                           \\ \hline \hline
$\mathcal{G}=\{\mathcal{U}, \mathcal{E}^+, \mathcal{E}^-\}$ & The signed graph                                      \\ \hline
$\mathcal{U} = \{u_i\}_{i=1}^n$                             & The set of nodes                                      \\ \hline
$\mathcal{E}^+$                                             & The set of positive edges                             \\ \hline
$\mathcal{E}^-$                                             & The set of negative edges                             \\ \hline
$A\in \{-1, 0, 1\}^{n\times n}$                             & The adjacency matrix of signed graph                  \\ \hline
$Y\in\{-1, 1\}$                                             & The label set of edge sign                            \\ \hline
$\tilde{\mathcal{G}}$                                       & The noisy signed graph                                \\ \hline
$\tilde{A}$                                                 & The noisy adjacency matrix                            \\ \hline
$\tilde{X}$                                                 & The noisy input node features                         \\ \hline
$\tilde{Y}$                                                 & The noisy label set of edge sign                      \\ \hline
$H$                                                         & The hidden representations of all input edges         \\ \hline
$I(\cdot)$                                                  & The mutual information                                \\ \hline
$Y_c$                                                       & The clean subset of $\tilde{Y}$                     \\ \hline
$A_c$                                                       & The clean subset of $\tilde{A}$                     \\ \hline
$X_c$                                                       & The node hidden representations after feature masking \\ \hline
$M$                                                         & The learnable feature mask                            \\ \hline
$u_i$                                                       & The node $i$                                            \\ \hline
$e_{ij}$                                                    & The edge from node $u_i$ to node $u_j$           \\ \hline
$\mathbb{P}(\cdot)$                                        & The probability distribution \\ \hline
$\mathbb{Q}(\cdot)$                                         & The marginal distribution \\ \hline
$\mathbb{P}_{\phi}(\cdot)$            & The probability distribution of embedding space after $f_{\phi}$ \\ \hline
$\mathbb{P}_{sgnn}(\cdot)$            & The probability distribution of embedding space after $f_{sgnn}$ \\ \hline
$\mathbb{P}_{\upsilon}(\cdot)$            & The variational approximation to the prior distribution $\mathbb{P}(\cdot)$ \\ \hline \hline
\end{tabular}%
}
\vspace{-15pt}
\end{table}

\section{Task Comparison}
\label{sec:task_com}

\begin{figure}[!hp]
  \centering
  \begin{subfigure}{0.48\linewidth}
    \centering
    \includegraphics[width=0.8\linewidth]{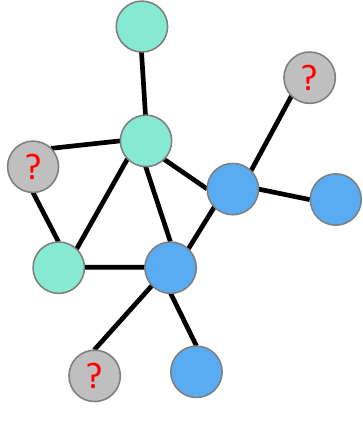}
    \caption{Node Classification.}
    \label{fig:node_cls}
  \end{subfigure}
  \begin{subfigure}{0.48\linewidth}
    \centering
    \includegraphics[width=0.8\linewidth]{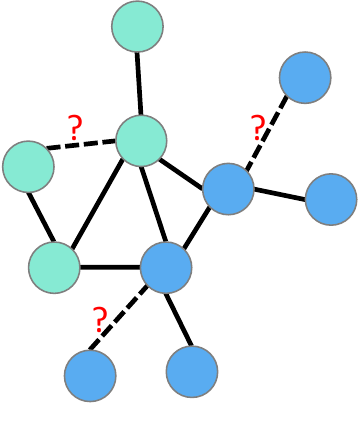}
    \caption{Link Prediction.}
    \label{fig:link_pre}
  \end{subfigure} \\
  \begin{subfigure}{0.48\linewidth}
    \centering
    \includegraphics[width=0.8\linewidth]{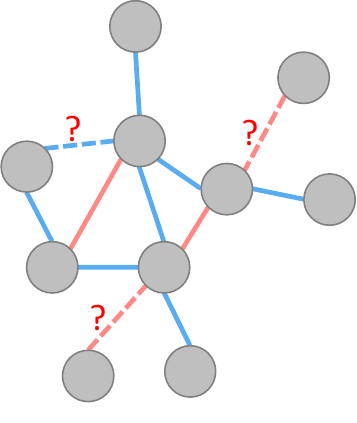}
    \caption{Link Sign Prediction.}
    \label{fig:link_sign_pre}
  \end{subfigure}
  \begin{subfigure}{0.48\linewidth}
    \centering
    \includegraphics[width=0.88\linewidth]{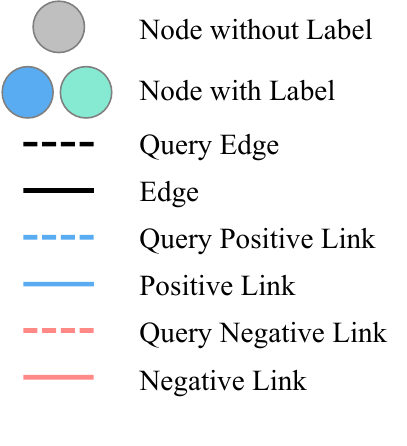}
    \caption{Legend.}
  \end{subfigure} 
  \vspace{-10pt}
  \caption{\textbf{Graph task comparison.}}
  \vspace{-10pt}
  \label{fig:task_com}
\end{figure}

\subsection{Basic Task Comparison}
As shown in Fig.~\ref{fig:task_com}, node classification~\cite{hou2024nc2d,peng2025rumor}, link prediction~\cite{liu2024ipm,chen2025molecular} and link sign prediction~\cite{derr2018signed,kim2018side,li2020learning} are three distinct tasks in the field of graph learning. 

For the task objective, node classification aims to predict the labels of nodes, while link prediction focuses on predicting the existence of a link between two nodes. On the other hand, link sign prediction goes a step further by predicting the sign (positive or negative) of a link between two nodes, which can represent different types of relationships, such as friendship or antagonism in social networks.

In terms of task concerns, node classification is primarily concerned with the attributes of individual nodes and their immediate neighbors. Link prediction, however, requires an understanding of the broader structure of the graph and the intricate interplay between nodes. Link sign prediction adds another layer of complexity by requiring the model to understand the nature of the relationship between nodes.

For empirical validation, the benchmarks for node classification and link prediction generally include the graph structure and attributes of nodes, such as Cora, Citeseer, and Pubmed. However, the datasets for link sign prediction preserve only the graph topology, such as Bitcoin\_OTC, Bitcoin\_Alpha, Epinions, and Slashdot.

\subsection{Robustness Comparison Between Signed and Unsigned Graphs}
\subsubsection{Robustness on Unsigned Graphs}
Unsigned graphs commonly represent systems like social networks, citation networks, and recommendation systems, where all relationships are positive. Noise in these graphs can arise from missing connections, spurious edges, or irrelevant nodes, typically confined to the structural space. Robustness aims to mitigate this noise while preserving the overall integrity of the graph's global structure.

\subsubsection{Robustness on Signed Graphs}
Signed graphs model both cooperation and conflict, as seen in contexts like online reviews, trust/distrust networks, or political polarization. Noise in signed graphs often manifests as incorrect edge signs (e.g., mislabeling positive relationships as negative and vice versa), and this noise propagates beyond the edge structure, affecting both the feature space and the target space while learning. Ensuring robustness in this setting requires handling such sign-specific noise carefully, without oversimplifying the graph's intricate relationships.

\noindent \textbf{DSGC.}
Recent advances in robust SGNNs have primarily been grounded in balance theory. However, as discussed in Sec.3, real-world signed graphs often contain a substantial proportion of unbalanced triangles that deviate from the assumptions of classical balance theory. This observation has motivated a shift in recent literature from strict balance theory toward the more flexible framework of weak balance theory.
A notable example is DSGC\cite{zhao2025robust}, a recent robust SGNN designed for the clustering task. Below, we provide a detailed comparison between \framework and DSGC:
\begin{itemize}
\item Theoretical Foundation: DSGC is grounded in weak balance theory, while \framework is based on GIB theory, which has broader practical applicability.
\item Methodology: \framework enhances robustness by leveraging information compression, whereas DSGC focuses on sign refinement to mitigate noise.
\item Task: \framework is designed for link sign prediction, in contrast to DSGC, which is tailored for node clustering.
\end{itemize}

\section{Detailed Experiment Setup}
\begin{figure*}[!tp]
  \begin{center}
  \begin{subfigure}{0.45\linewidth}
    \centering
    \includegraphics[width=\linewidth]{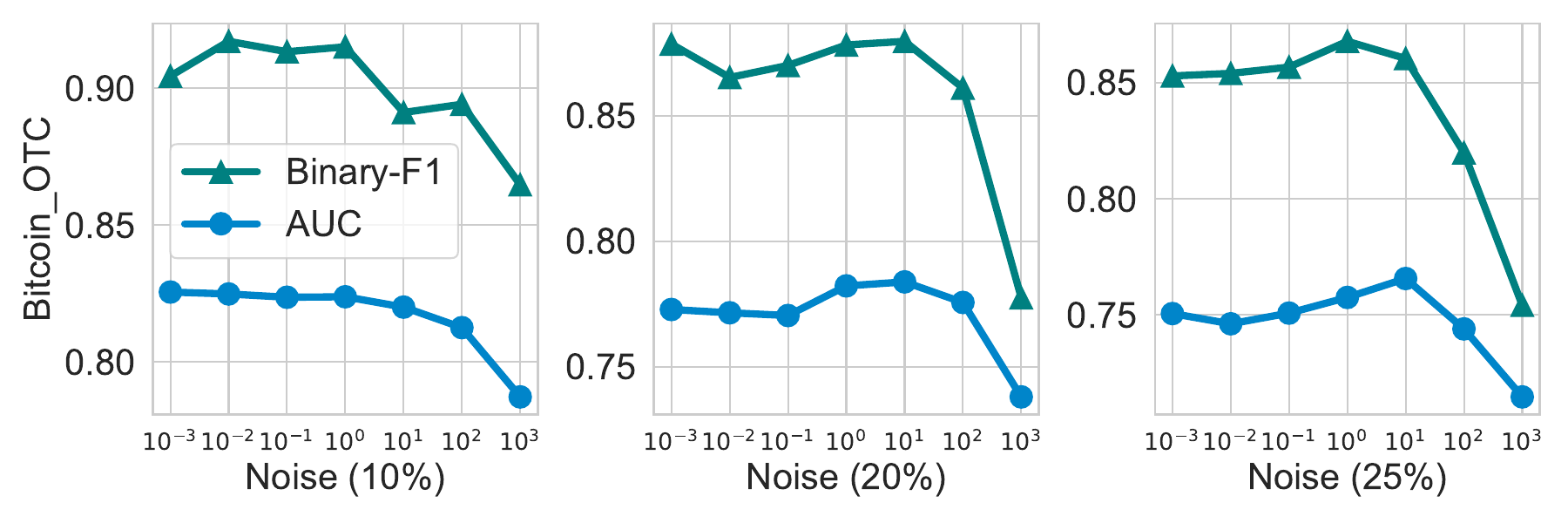}
    \vspace{-20pt}
    \caption{Parameter $\alpha$.}
  \end{subfigure}
  \begin{subfigure}{0.45\linewidth}
    \centering
    \includegraphics[width=\linewidth]{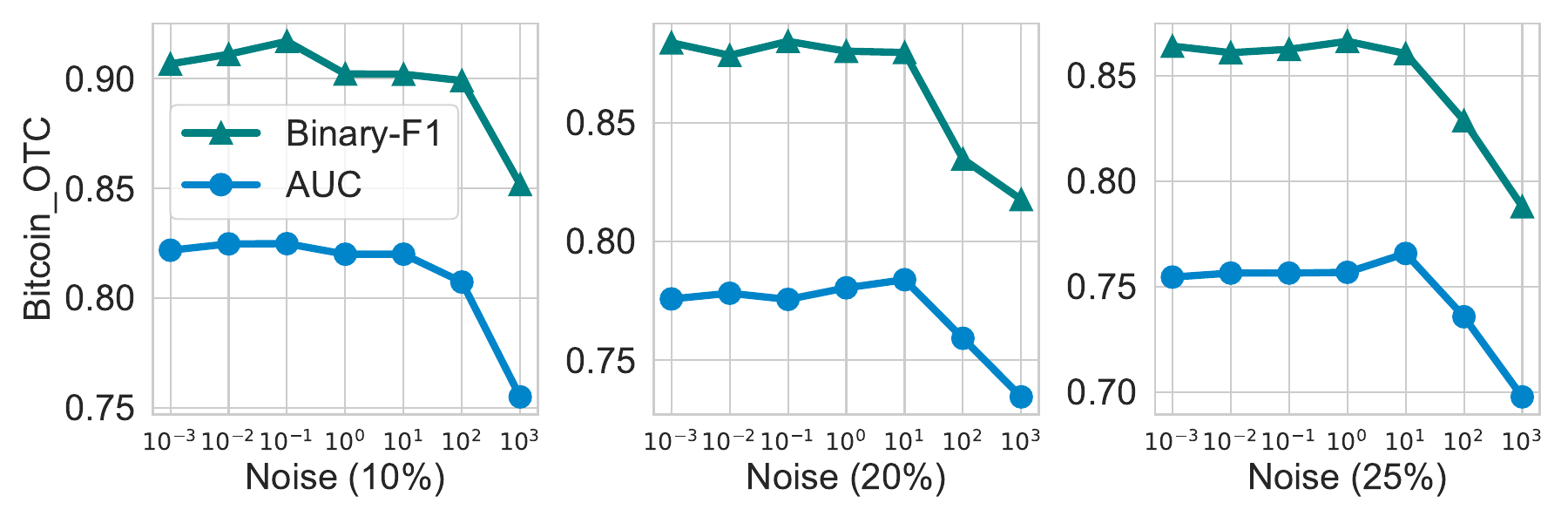}
    \vspace{-20pt}
    \caption{Parameter $\beta$.}
  \end{subfigure} \\
  \begin{subfigure}{0.45\linewidth}
    \centering
    \includegraphics[width=\linewidth]{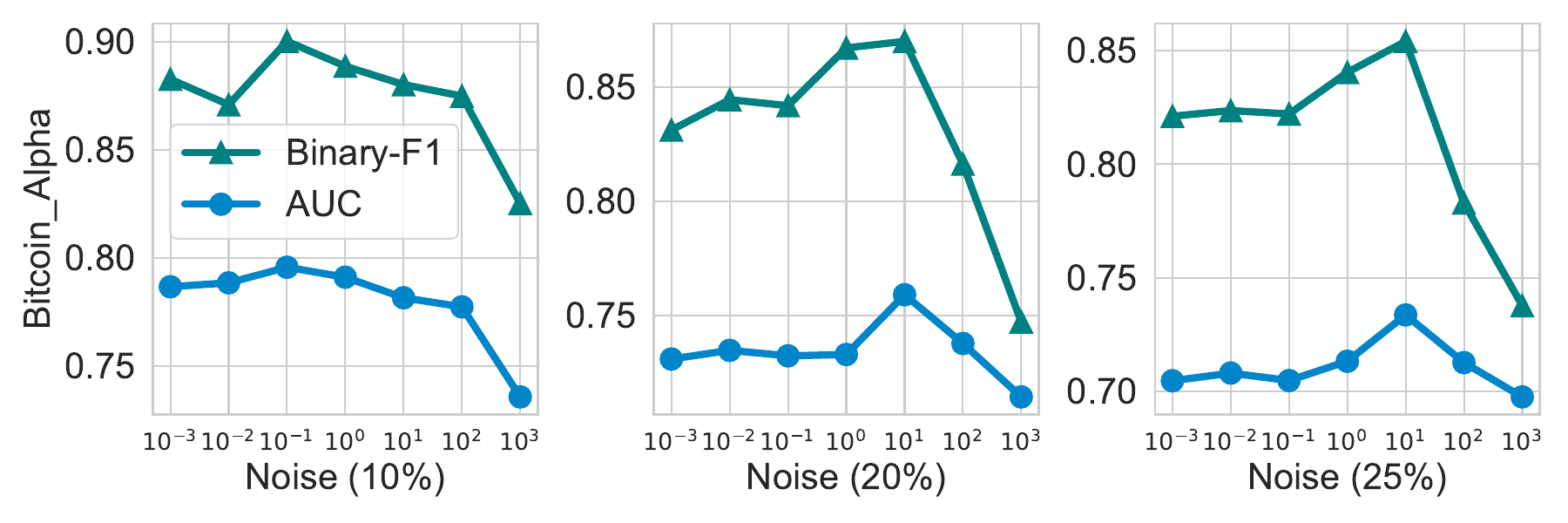}
    \vspace{-20pt}
    \caption{Parameter $\alpha$.}
  \end{subfigure}
  \begin{subfigure}{0.45\linewidth}
    \centering
    \includegraphics[width=\linewidth]{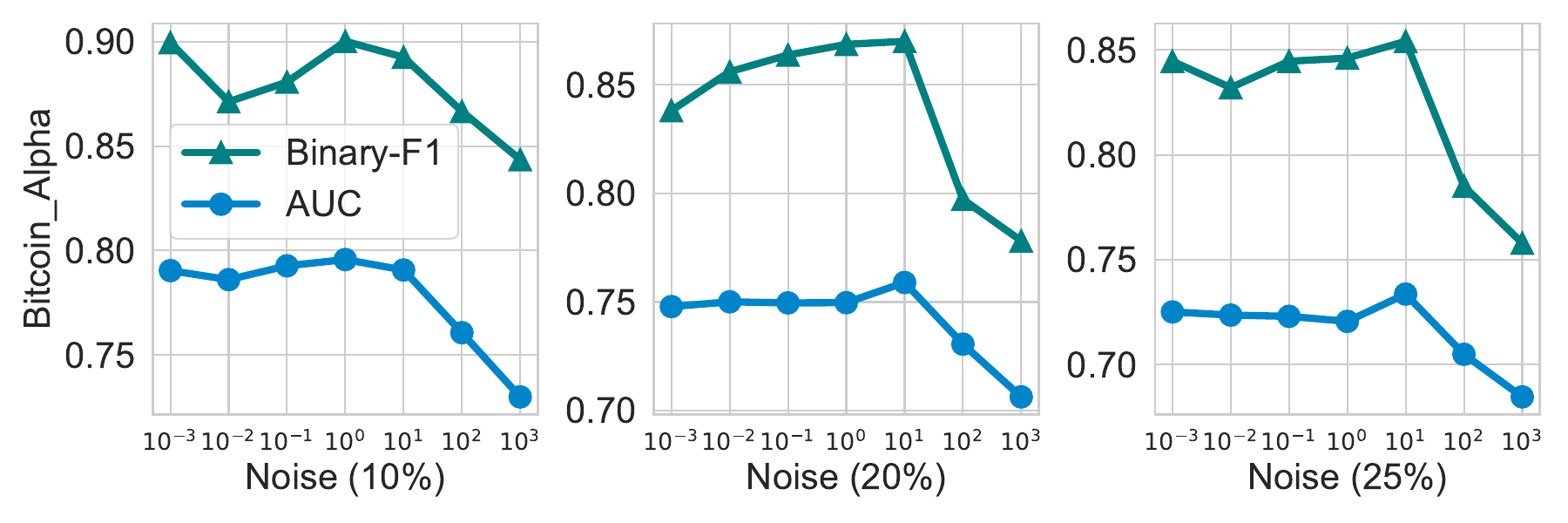}
    \vspace{-20pt}
    \caption{Parameter $\beta$.}
  \end{subfigure} \\
  \begin{subfigure}{0.45\linewidth}
    \centering
    \includegraphics[width=\linewidth]{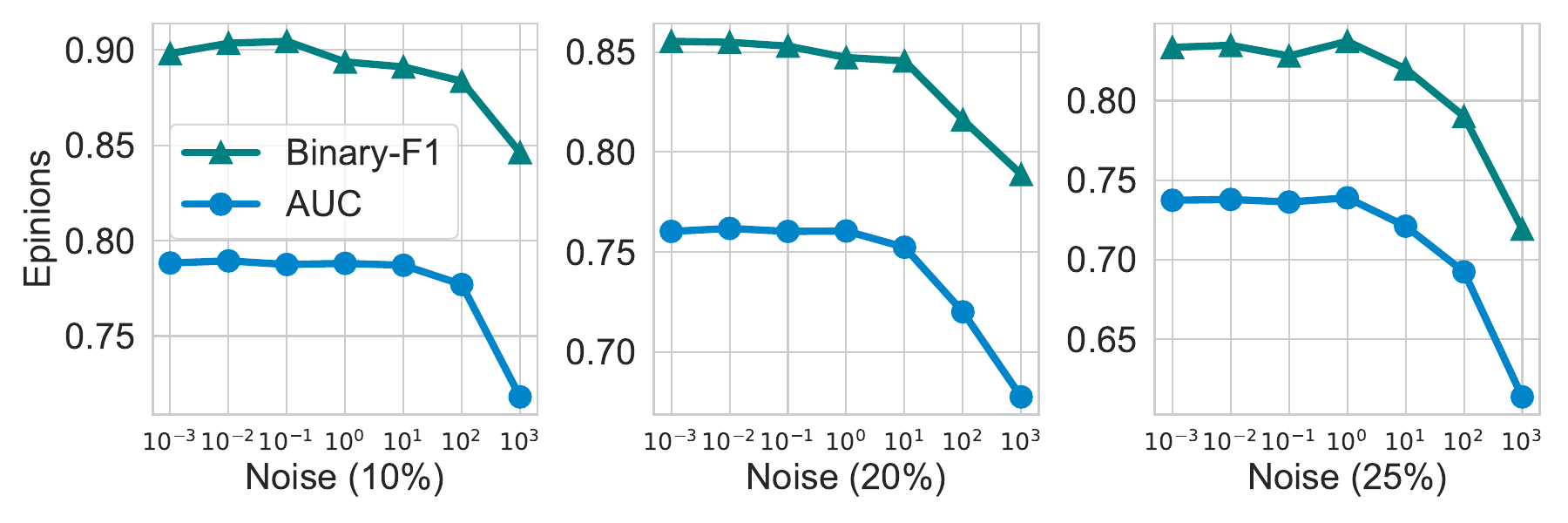}
    \vspace{-20pt}
    \caption{Parameter $\alpha$.}
  \end{subfigure}
  \begin{subfigure}{0.45\linewidth}
    \centering
    \includegraphics[width=\linewidth]{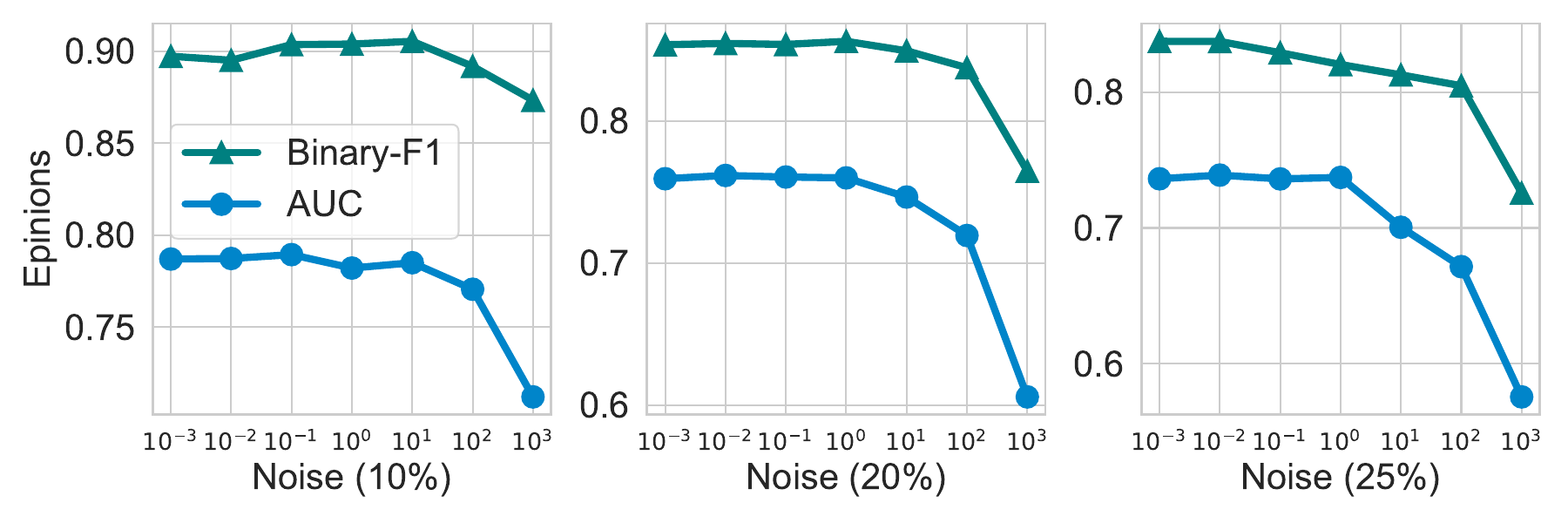}
    \vspace{-20pt}
    \caption{Parameter $\beta$.}
  \end{subfigure} \\
  \begin{subfigure}{0.45\linewidth}
    \centering
    \includegraphics[width=\linewidth]{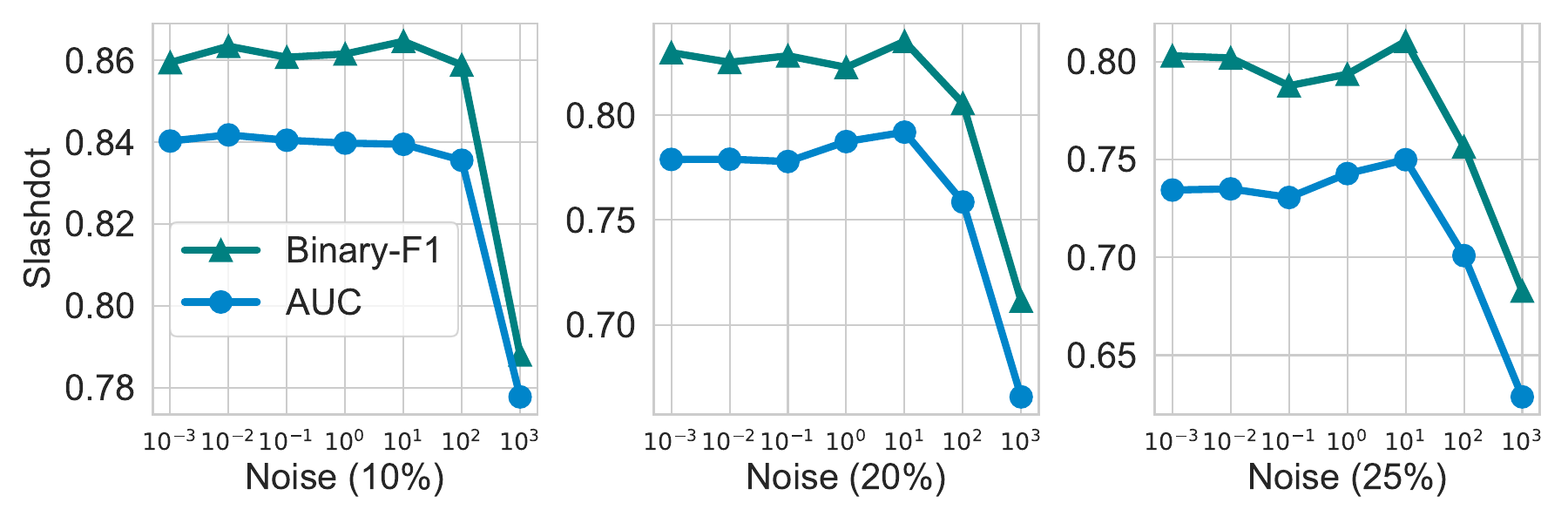}
    \vspace{-20pt}
    \caption{Parameter $\alpha$.}
  \end{subfigure}
  \begin{subfigure}{0.45\linewidth}
    \centering
    \includegraphics[width=\linewidth]{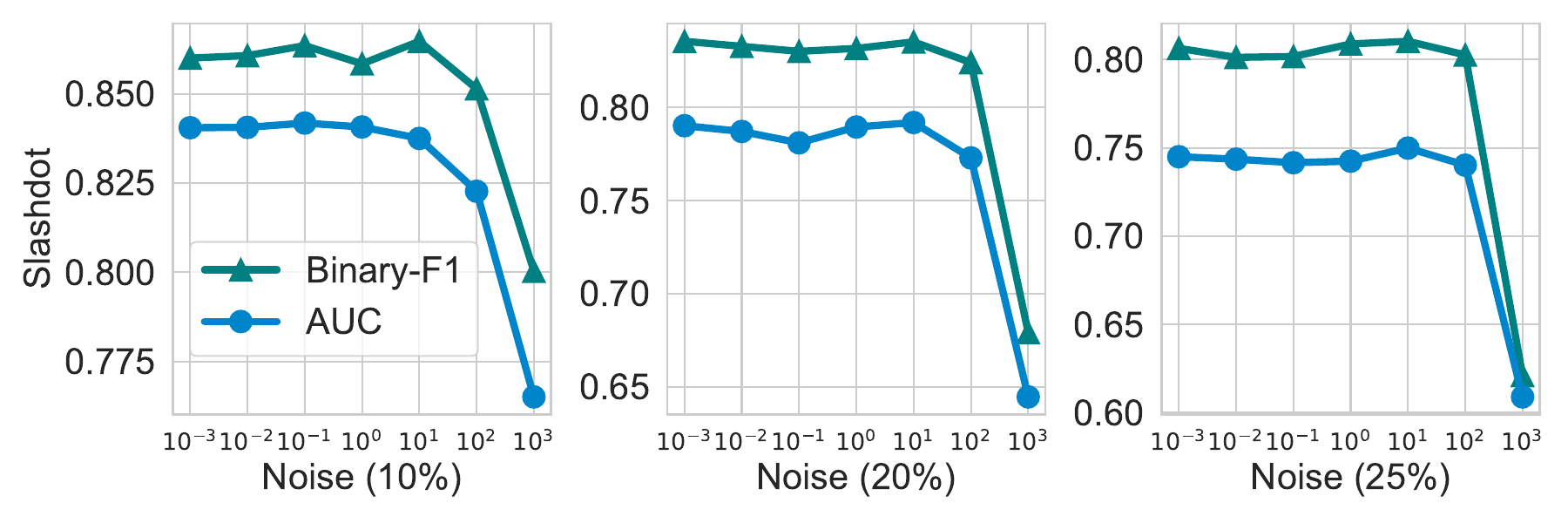}
    \vspace{-20pt}
    \caption{Parameter $\beta$.}
  \end{subfigure} 
  \vspace{-5pt}
  \caption{\textbf{Hyper-parameter sensitivity a.k.a. $\alpha$ and $\beta$.}}
  \label{fig:hyper}
  \end{center}
\end{figure*}

\subsection{Summary of Datasets}
\label{sec:app_data}

\begin{table}[!ht]
\centering
\caption{Triangle statistics (balanced vs. unbalanced) on the four adopted datasets.}
\label{tab:unbalanced_tri}
\resizebox{\columnwidth}{!}{%
\begin{tabular}{l|ccc}
\hline \hline
Datasets      &\#Triangles &\#Balanced-Tri &\#Unbalanced-Tri \\ \hline \hline
Bitcoin\_OTC   & 200,958    & 175,381       & 25,577          \\
Bitcoin\_Alpha & 132,918    & 113,566       & 19,3542         \\
Epinions      & 7,170,984  & 6,322,617     & 818,367         \\
Slashdot      & 2,364,708  & 2,034,077     & 330,631         \\ \hline \hline
\end{tabular}%
}
\end{table}

\begin{itemize}
    \item \textbf{Bitcoin\_Alpha}~\footnote{\url{http://www.btc-alpha.com}} and \textbf{Bitcoin\_OTC}~\footnote{\url{http://www.bitcoin-otc.com}} are who-trusts-who networks of people who using Bitcoin on the platforms of Bitcoin Alpha and Bitcoin OTC. Since Bitcoin users are anonymous, people give trust or not-trust tags to others in order to enhance security.
    \item \textbf{Epinions}~\footnote{\url{http://www.epinions.com}} is a who-trust-whom online social network of a general consumer review site Epinions.com. Members of the site can decide whether to trust each other.
    \item \textbf{Slashdot}~\footnote{\url{http://www.slashdot.com}} is a technology-related news website known for its specific user community which allows users to tag each other as friends or foes.
\end{itemize}

In these datasets, users rate others from -10 (completely distrust) to 10 (completely trust). We treat the marks bigger than 0 as positive edges and others as negative edges. In particular, as shown in Table~\ref{tab:unbalanced_tri}, the four original datasets contain a substantial number of unbalanced triangles. This observation helps explain the suboptimal performance of robustness methods grounded in balance theory. In other words, forcing the learned graph to be overly balanced may distort the data and shift it from the true distribution.

\subsection{Detailed Information for Baselines}
The adopted baselines in this work are listed below:
\label{sec:app_baseline}
\begin{itemize}
  \item SiNE~\cite{wang2017signed}, is a representative signed graph embedding method based on deep neural networks. The loss function is based on extended structural balance theory which drives nodes connected with positive edges closer than those connected with negative edges.
  \item SGCN~\cite{derr2018signed}, generalizes GCN to signed graphs by designing a new information aggregator which is based on balance theory. It is also the encoding part of our \framework model.
  \item SNEA~\cite{li2020learning}, generalizes GAT to signed graphs which adopts attention-based aggregators in message passing mechanism and is also based on the balance theory.
  \item SDGNN~\cite{huang2021sdgnn}, first reviews two fundamental sociological theories (i.e., status theory and balance theory) to analyze the social mechanism in signed directed networks. 
  \item MSGNN~\cite{he2022msgnn}, introduce a signed directed Laplacian matrix, referred to as the magnetic signed Laplacian, as a natural generalization of both the signed Laplacian on signed graphs and the magnetic Laplacian on directed graphs.
  \item SE-SGformer~\cite{li2025se}, proposes a self-explainable transformer-based framework for signed graph neural networks that enhances both link sign prediction accuracy and explainability by leveraging signed random walk positional encoding and a novel K-nearest neighbor-based decision process.
  \item SGCL~\cite{shu2021sgcl}, generalizes graph contrastive learning to signed graphs, which employs graph augmentations to reduce the harm of interaction noise.
  \item GS-GNN~\cite{liu2021signed}, beyond the balance theory assumption and adopt a dual GNN architecture to encoder both global and local information which claims to be noise-tolerant.
  \item SBGCL~\cite{zhang2023contrastive}, is the first to use contrastive learning to improve the robustness of graph representation learning for signed bipartite graphs.
  \item RSGNN~\cite{zhang2023rsgnn}, is the first work designed for robust SGNNs under random noise. Leaded by some empirical properties, RSGNN directs the process of structure learning to denoise the perturbed signed graph.
  \item LRWSB~\cite{minici2024link}, proposes a semi-supervised framework that enhances link polarity prediction in signed graphs with sparse and noisy data by incorporating multiscale social balance theory.
\end{itemize}

In particular, the results for SDGNN, MSGNN, SE-SGformer, SBGCL and LRWSB are reproduced using their publicly available code, following the same experimental settings as RSGNN. The results for the other baselines are derived from the RSGNN paper~\cite{zhang2023rsgnn}.

\subsection{Experiment Environment}
The software includes Python 3.8.12, Pytorch 1.12, CUDA 11.0, and Pytorch-Geometric 2.5.2.
The hardware includes Intel(R) Xeon(R) Silver 4214R CPU, 128GB RAM, and NVIDIA RTX 3090 GPU.

\section{More Experiments}
\subsection{Hyper-parameter Sensitivity}
\label{sec:app-hyper}
Fig.~\ref{fig:hyper} shows trends of AUC and Binary-F1 on the all four datasets under various random noise ratios (i.e., 10\%, 20\%, and 25\%).

\subsection{Runtime Complexity}
\label{sec:runtime}
The overall runtime complexity of \framework is $\mathcal{O}(L \cdot (m \cdot d + n \cdot d^2) + n^2 \cdot d + m\cdot (c+d))$. Specifically, 
\begin{itemize}
\item $\mathcal{O}(L \cdot (m \cdot d + n \cdot d^2))$ is the runtime complexity of SGCN (use SGCN as the basic signed graph encoder, i.e., $f_{sgnn}$), $L$ is the number of convolutional layers, $m$ is the number of edges, $n$ is the number of nodes, and $d$ is the hidden dimension.
\item The runtime complexity of feature masking is $\mathcal{O}(n \cdot d)$.
\item The runtime complexity of substructure sampling is $\mathcal{O}(n^2 \cdot d + m)$.
\item The runtime complexity of losses is $\mathcal{O}(m\cdot (c+d))$, $c$ is the class number.
\end{itemize}

\noindent \textbf{Runtime Complexity Comparison.}
\begin{itemize}
    \item \textbf{RIDGE}: As $c$ and $d$ are always fixed, for simplicity, the complexity can be $\mathcal{O}_{SGNN}+ \mathcal{O}(n^2+m)$.

    \item \textbf{RSGNN}: The runtime complexity of RSGNN can be simplified to $\mathcal{O}(n^3)+\mathcal{O}_{SGNN}$, as the $O(n^3)$ is derived from the training objective of high balance degree.

    \item \textbf{LRWSB}: The runtime complexity of LRWSB can be simplified to $\mathcal{O}(n^3+m)+\mathcal{O}_{SGNN}$, the main complexity comes from enumerating all transitive triads, which is also the main time consumed by LRWSB in experiments.
\end{itemize}

As can be seen, the runtime complexity of our method is much lower than the other two robust SGNNs and consistent with the computation resource consumption in scalability and training efficiency analysis.

\begin{figure*}[!ht]
  \centering
  \begin{subfigure}{0.24\linewidth}
    \centering
    \includegraphics[width=\linewidth]{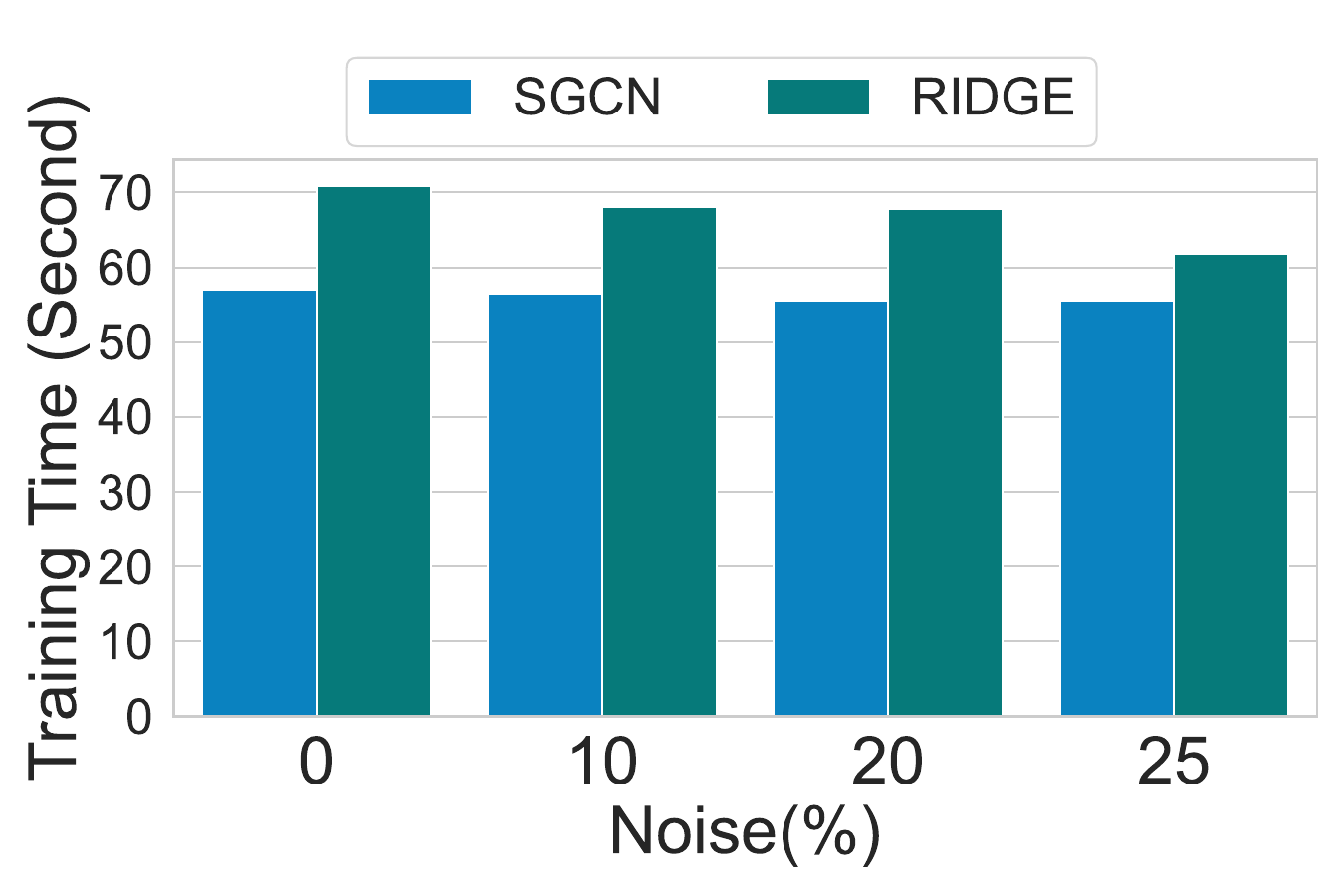}
    \caption{Bitcoin\_OTC.}
    % \label{fig:node_cls}
  \end{subfigure}
  \begin{subfigure}{0.24\linewidth}
    \centering
    \includegraphics[width=\linewidth]{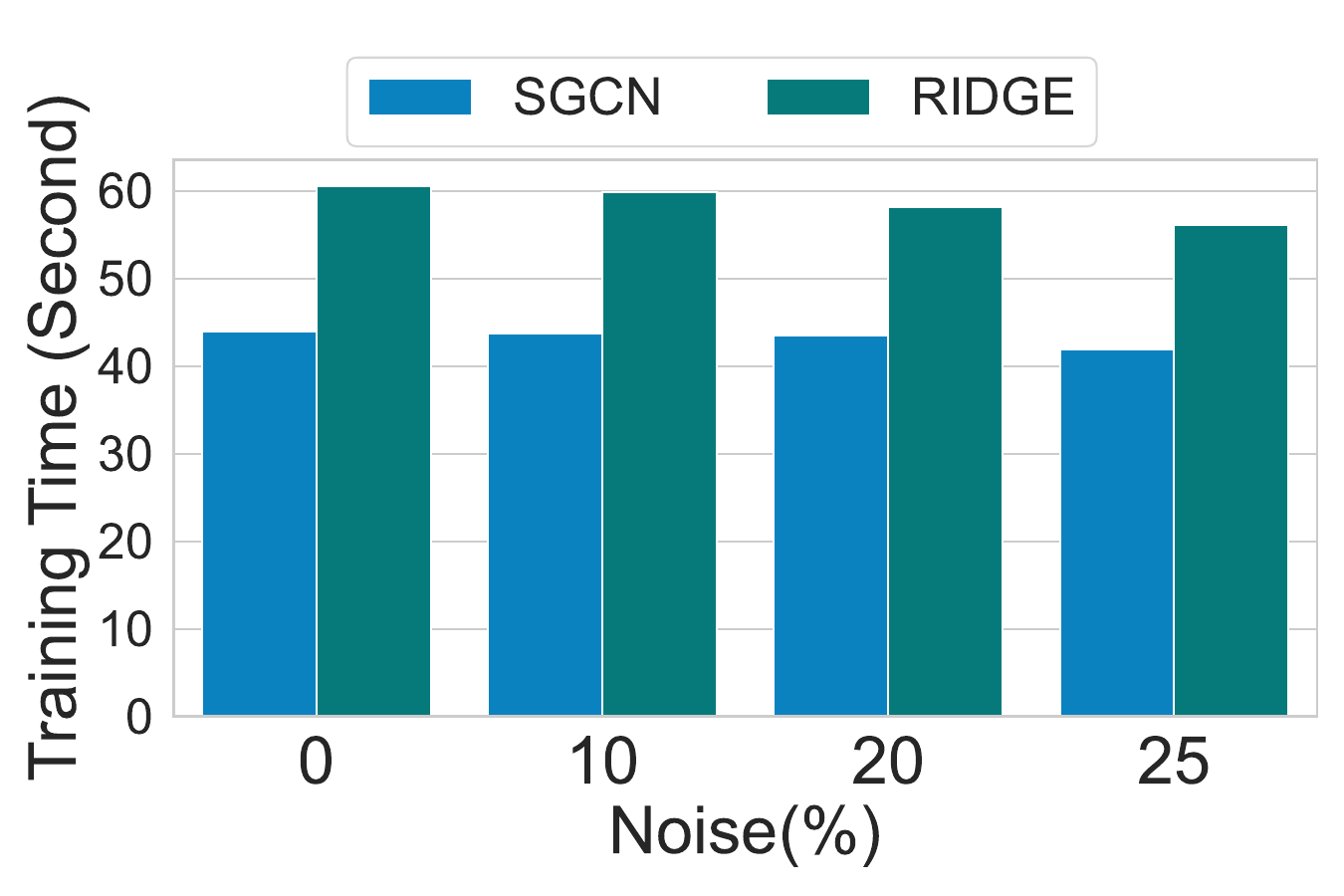}
    \caption{Bitcoin\_Alpha.}
    % \label{fig:link_pre}
  \end{subfigure} 
  \begin{subfigure}{0.24\linewidth}
    \centering
    \includegraphics[width=\linewidth]{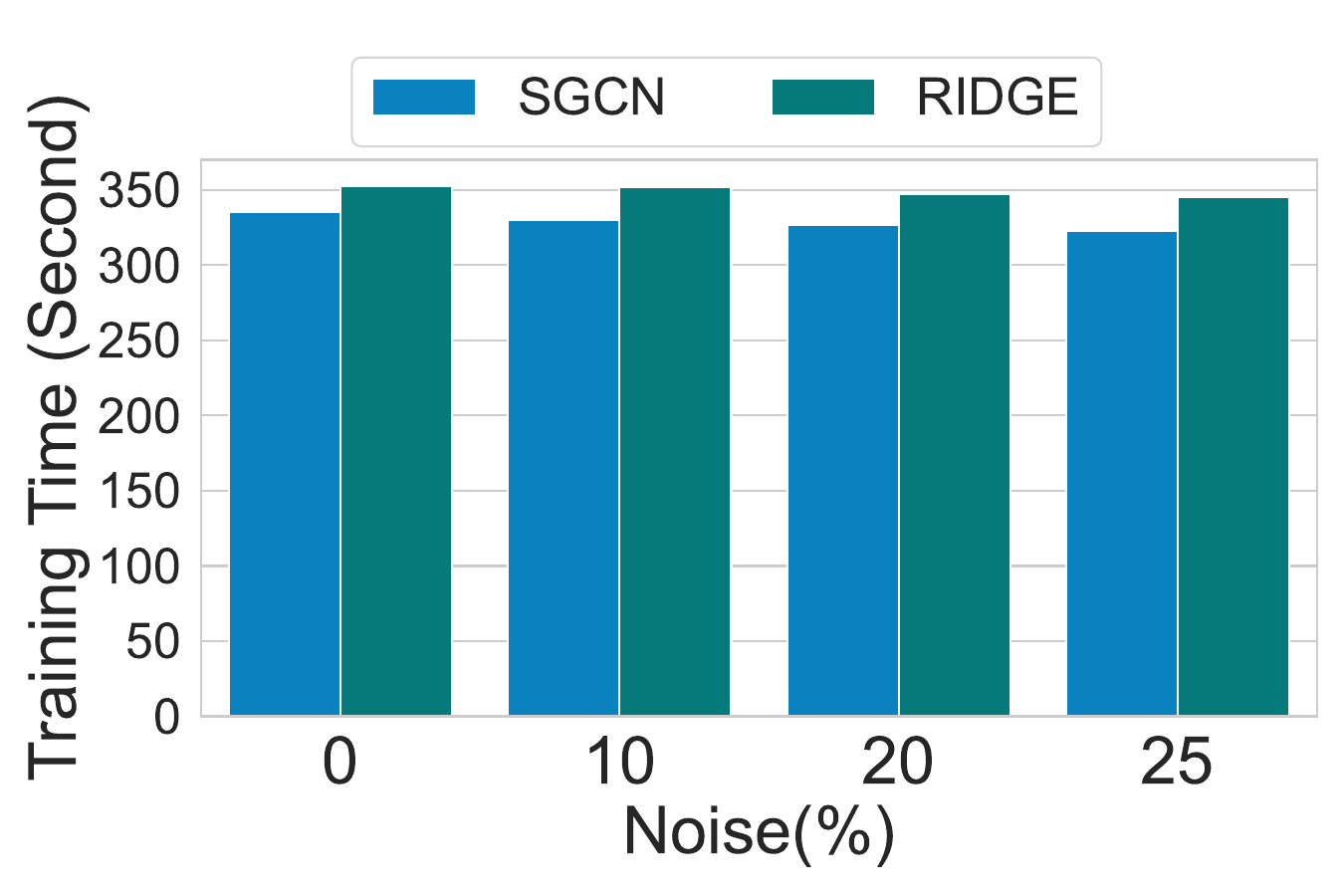}
    \caption{Epinions.}
    % \label{fig:link_sign_pre}
  \end{subfigure}
  \begin{subfigure}{0.24\linewidth}
    \centering
    \includegraphics[width=\linewidth]{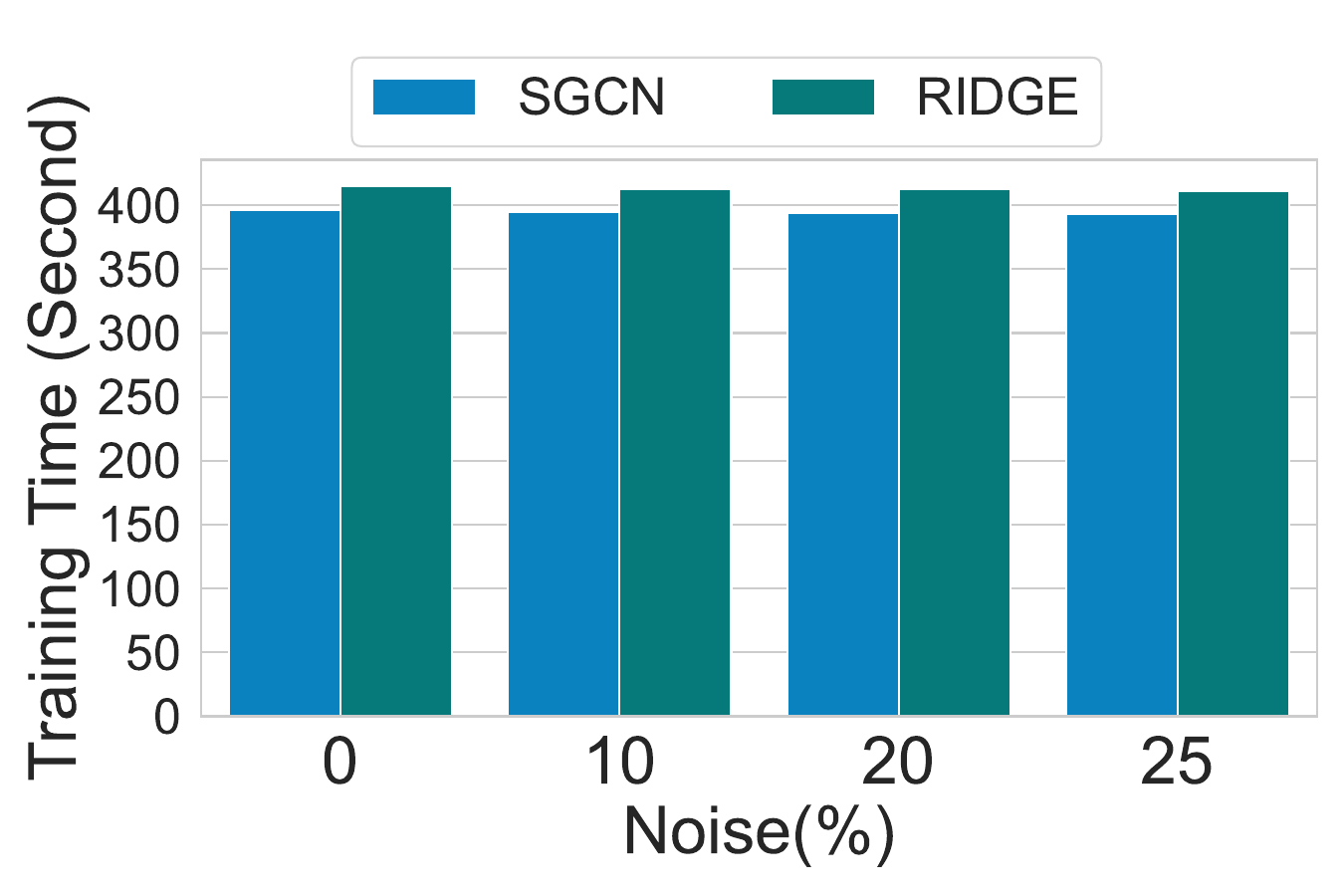}
    \caption{Slashdot.}
  \end{subfigure} 
  \vskip -5pt
  \caption{\textbf{Training time comparison between \framework and SGCN.}}
  \label{fig:training_time}
\end{figure*}

\begin{figure*}[!ht]
  \centering
  \begin{subfigure}{0.24\linewidth}
    \centering
    \includegraphics[width=\linewidth]{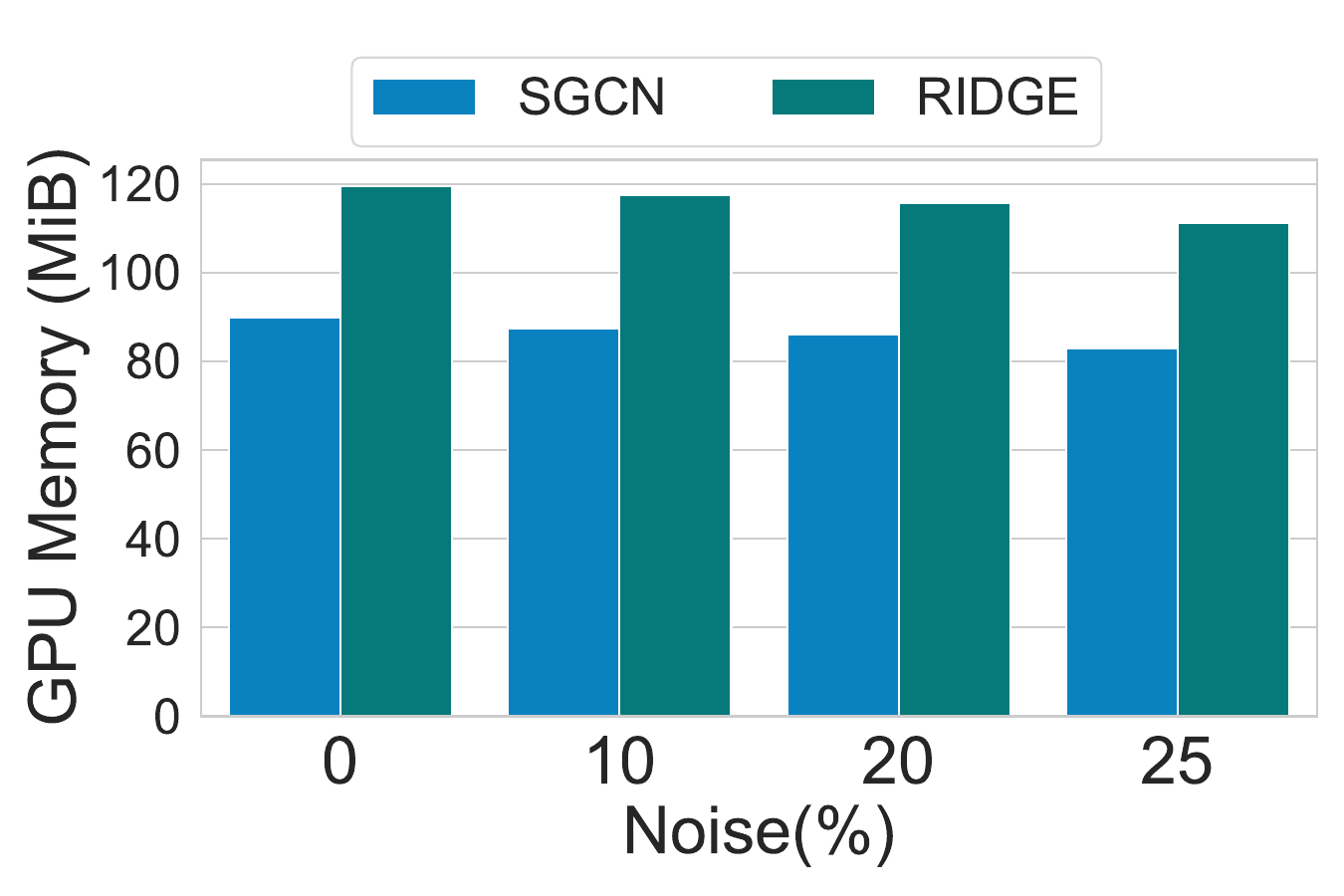}
    \caption{Bitcoin\_OTC.}
    % \label{fig:node_cls}
  \end{subfigure}
  \begin{subfigure}{0.24\linewidth}
    \centering
    \includegraphics[width=\linewidth]{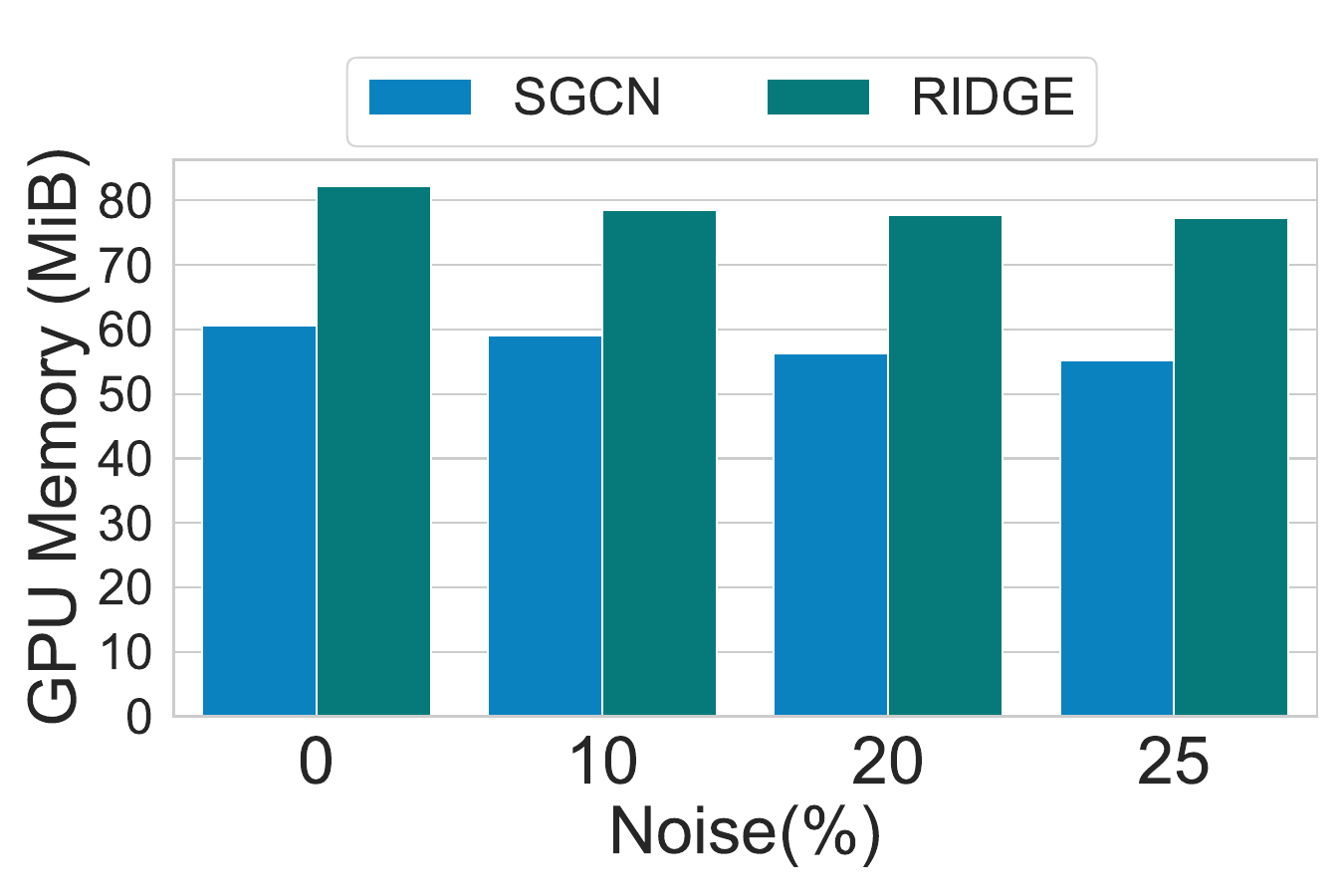}
    \caption{Bitcoin\_Alpha.}
    % \label{fig:link_pre}
  \end{subfigure} 
  \begin{subfigure}{0.24\linewidth}
    \centering
    \includegraphics[width=\linewidth]{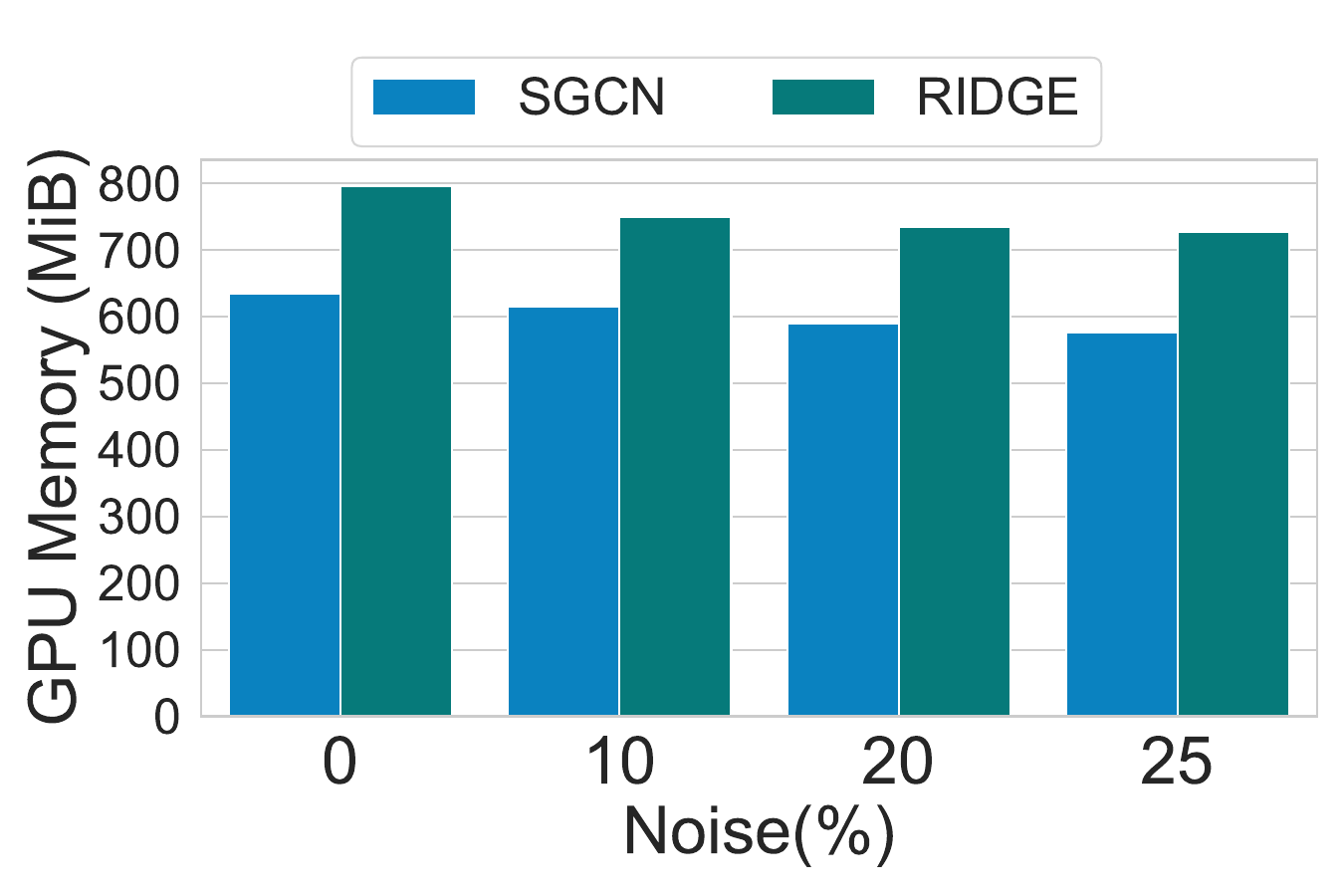}
    \caption{Epinions.}
    % \label{fig:link_sign_pre}
  \end{subfigure}
  \begin{subfigure}{0.24\linewidth}
    \centering
    \includegraphics[width=\linewidth]{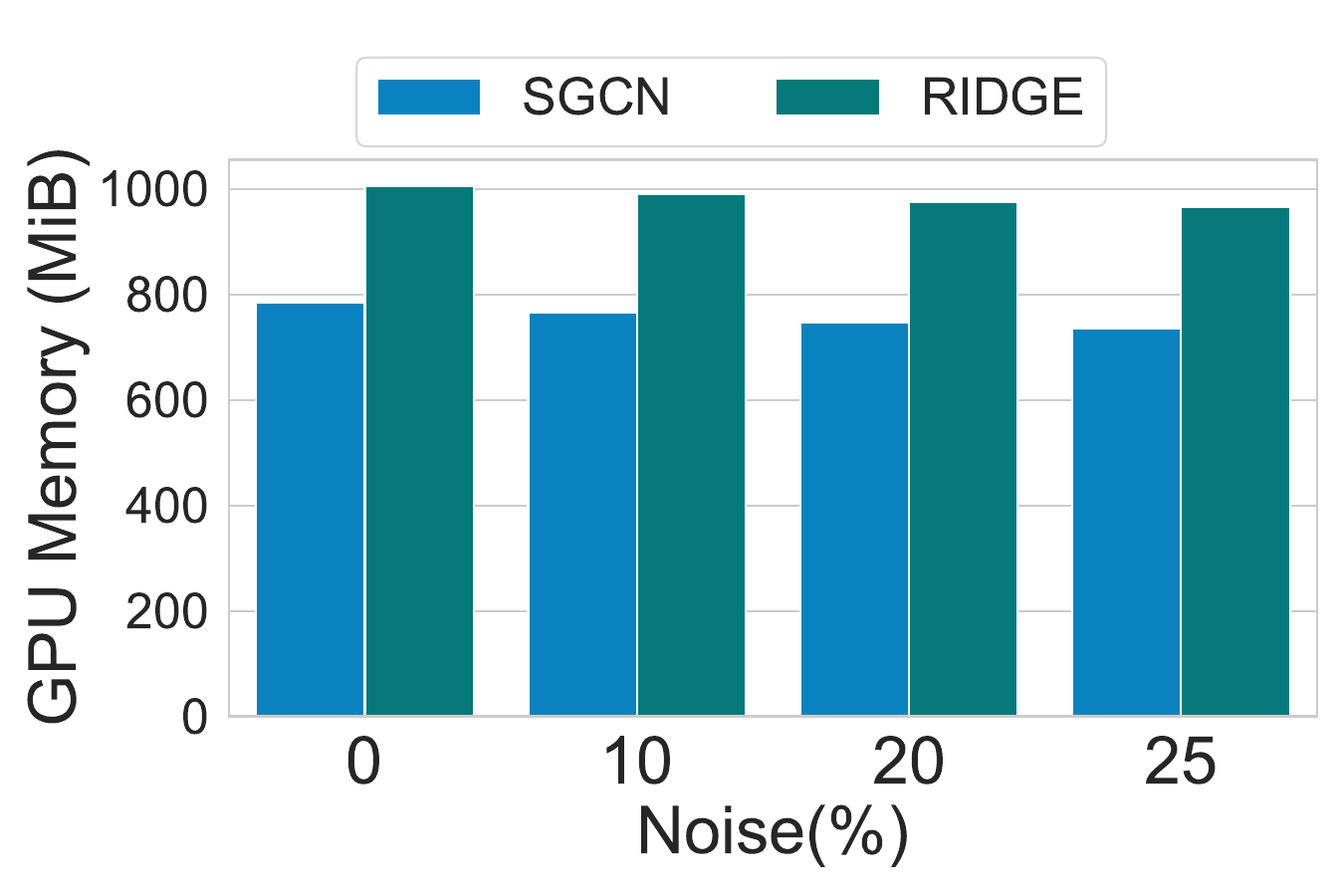}
    \caption{Slashdot.}
  \end{subfigure} 
  \vskip -5pt
  \caption{\textbf{GPU memory usage comparison between \framework and SGCN.}}
  \label{fig:gpu_mem}
\end{figure*}

\begin{figure*}[!htbp]
  \centering
    \includegraphics[width=\linewidth]{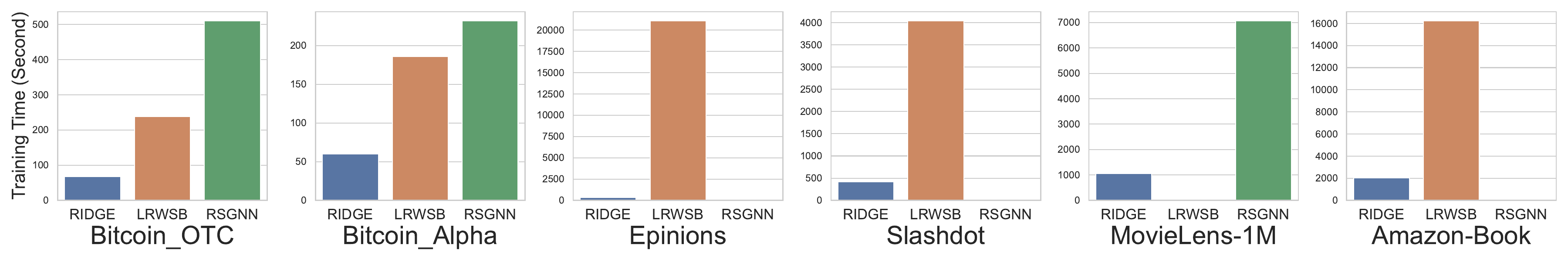}
    \vspace{-20pt}
    \caption{Training time comparison. RSGNN is out of memory on Epinions, Slashdot and Amazon-Book. LRWSb is out of time ($\geq$ 48 Hours) on MovieLens-1M.}
    \label{fig:training_time_comparison}
\end{figure*}

\begin{figure*}[!htbp]
    \centering
    \includegraphics[width=\linewidth]{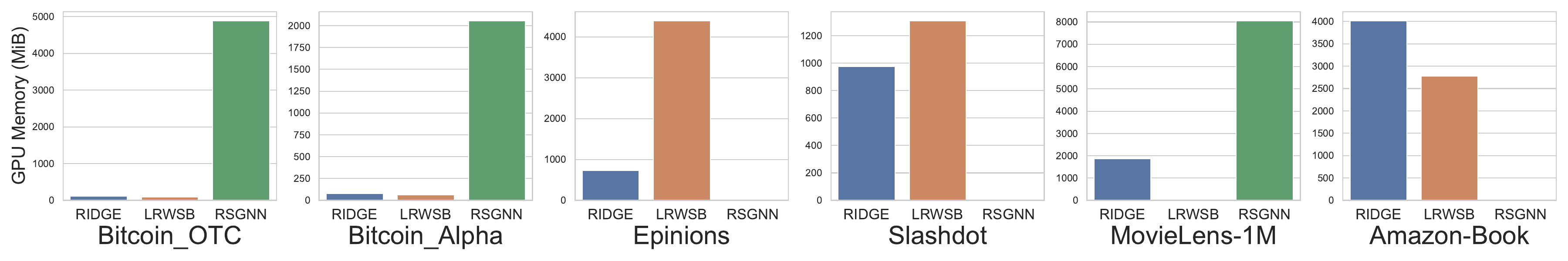}
    \vspace{-20pt}
    \caption{Training GPU memory comparison. RSGNN is out of memory on Epinions, Slashdot and Amazon-Book. LRWSb is out of time ($\geq$ 48 Hours) on MovieLens-1M.}
    \label{fig:training_mem_comparison}
\end{figure*} 

\subsection{Efficiency}
To further reveal the efficiency of our method, we have conducted an experimental analysis to evaluate the computational overhead of our approach. Specifically, we compared the training time and GPU memory consumption of our framework against the corresponding baseline encoder (i.e., SGCN). As shown in Fig.~\ref{fig:training_time} and Fig.~\ref{fig:gpu_mem}, our framework requires approximately the same training time (8\% additional time) and only 30\% additional GPU memory. Despite this modest increase in computational overhead, our framework demonstrates significantly superior robustness across various noise scenarios. These results provide a practical perspective on the scalability and efficiency of our framework.  

\noindent \textbf{Remark.} The decreasing trend in overhead with increased noise can be attributed to the classification loss mechanism of SGCN, which relies on negative sampling. The introduction of random noise alters the positive-to-negative sample ratio, affecting the overhead. Since \framework employs SGCN as its encoder, it exhibits a similar trend.

\noindent \textbf{Comparison among robust SGNNs.} In addition to the runtime complexity analysis presented in Sec.\ref{sec:runtime}, we further evaluate the actual training overhead of \framework in comparison among robust SGNNs, including RSGNN and LRWSB.
As shown in Fig.\ref{fig:training_time_comparison} and Fig.~\ref{fig:training_mem_comparison}, \framework achieves significantly higher efficiency in both training time and GPU memory consumption.
While LRWSB requires less GPU memory due to its pre-processing step that filters transitive triangles containing exactly one unknown edge, it incurs substantially longer training time than \framework on Amazon-Book.

\begin{table}[h]
\centering
\caption{Robust SGNN comparison on large graphs. OOM is out of memory (GPU, 3090 24GiB). OOT is out of time (more than 48 hours).}
\vspace{-10pt}
\label{tab:large_graph}
\resizebox{\columnwidth}{!}{%
\begin{tabular}{l|ll|ccc}
\hline \hline
Dataset      & Noise(\%) & Metric & RIDGE & RSGNN & LRWSB \\ \hline \hline
MovieLens-1M & 10    & AUC    & 66.12 & 64.38 & OOT   \\
             &       & F1     & 68.35 & 66.75 & OOT   \\ \cline{2-6} 
             & 20    & AUC    & 64.87 & 62.19 & OOT   \\
             &       & F1     & 67.55 & 65.59 & OOT   \\ \hline
Amazon-Book  & 10    & AUC    & 61.42 & OOM   & 60.88 \\
             &       & F1     & 70.78 & OOM   & 67.25 \\ \cline{2-6} 
             & 20    & AUC    & 59.68 & OOM   & 56.87 \\
             &       & F1     & 68.84 & OOM   & 65.69 \\ \hline \hline
\end{tabular}%
}
\vspace{-10pt}
\end{table}

\begin{table*}[!tp]
\centering
\caption{Results of \framework along with baselines under adversarial noise. Bold indicates the best performance over all methods.}
\label{tab:ad_results}
\resizebox{\textwidth}{!}{%
\begin{tabular}{l|l|l|cccc|cccc|cccc|cccc}
\hline \hline
\multicolumn{3}{l|}{Dataset} & \multicolumn{4}{c|}{Bitcoin\_OTC} & \multicolumn{4}{c}{Bitcoin\_Alpha|} & \multicolumn{4}{c|}{Epinions} & \multicolumn{4}{c}{Slashdot} \\ \hline
\multicolumn{3}{l|}{Noise(\%)} & 5 & 10 & 15 & 20 & 5 & 10 & 15 & 20 & 5 & 10 & 15 & 20 & 5 & 10 & 15 & 20 \\ \hline \hline
\multirow{24}{*}{basic SGNN} & \multirow{4}{*}{SiNE} & AUC & 76.17 & 74.98 & 71.55 & 68.17 & 74.18 & 68.73 & 67.12 & 65.17 & 78.12 & 74.25 & 72.77 & 69.17 & 73.27 & 69.17 & 64.17 & 62.15 \\
 &  & Macro-F1 & 66.17 & 62.37 & 57.94 & 55.84 & 58.15 & 52.17 & 49.15 & 47.58 & 65.05 & 61.18 & 58.84 & 57.17 & 64.24 & 58.15 & 54.17 & 51.68 \\
 &  & Micro-F1 & 77.35 & 73.15 & 69.43 & 67.27 & 71.35 & 65.95 & 61.57 & 59.85 & 76.27 & 69.17 & 66.13 & 63.77 & 70.12 & 64.15 & 60.15 & 58.55 \\
 &  & Binary-F1 & 85.23 & 83.04 & 80.94 & 78.43 & 82.43 & 75.77 & 74.82 & 72.93 & 84.45 & 78.74 & 76.85 & 74.46 & 79.85 & 74.64 & 69.17 & 69.14 \\ \cline{2-19} 
 & \multirow{4}{*}{SGCN} & AUC & 77.58 & 75.08 & 72.75 & 69.82 & 73.58 & 69.17 & 68.52 & 65.32 & 77.15 & 73.87 & 71.47 & 68.85 & 74.37 & 68.97 & 64.92 & 63.44 \\
 &  & Macro-F1 & 64.75 & 61.47 & 58.18 & 56.25 & 56.43 & 51.28 & 48.32 & 47.08 & 66.07 & 61.22 & 58.47 & 56.02 & 63.37 & 57.14 & 54.08 & 52.02 \\
 &  & Micro-F1 & 78.25 & 74.77 & 70.63 & 68.36 & 72.65 & 66.04 & 62.02 & 60.25 & 77.37 & 70.35 & 67.14 & 64.56 & 71.95 & 65.15 & 60.22 & 59.35 \\
 &  & Binary-F1 & 86.64 & 84.05 & 81.07 & 79.22 & 81.37 & 76.82 & 75.65 & 73.43 & 85.25 & 79.73 & 77.04 & 74.94 & 80.94 & 75.25 & 70.96 & 70.25 \\ \cline{2-19} 
 & \multirow{4}{*}{SNEA} & AUC & 78.18 & 75.15 & 73.55 & 70.42 & 74.25 & 69.98 & 69.17 & 66.97 & 78.37 & 74.25 & 71.52 & 68.95 & 75.28 & 69.94 & 65.95 & 64.12 \\
 &  & Macro-F1 & 65.18 & 62.82 & 59.38 & 56.94 & 57.98 & 52.35 & 49.82 & 48.85 & 67.28 & 62.35 & 58.92 & 56.74 & 64.77 & 58.28 & 55.67 & 52.54 \\
 &  & Micro-F1 & 79.55 & 75.35 & 71.55 & 69.85 & 73.22 & 67.34 & 63.05 & 61.34 & 78.45 & 71.15 & 68.35 & 65.73 & 72.03 & 66.45 & 61.93 & 60.65 \\
 &  & Binary-F1 & 87.72 & 85.34 & 82.55 & 80.97 & 82.73 & 77.55 & 76.36 & 74.84 & 86.34 & 80.97 & 78.46 & 75.83 & 81.34 & 76.45 & 71.86 & 71.35 \\ \cline{2-19} 
 & \multirow{4}{*}{BESIDE} & AUC & 83.77 & 80.34 & 77.17 & 74.15 & 79.78 & 75.64 & 73.24 & 69.72 & 80.78 & 74.74 & 72.07 & 69.92 & 78.37 & 76.27 & 73.97 & 71.57 \\
 &  & Macro-F1 & 72.32 & 64.52 & 59.18 & 55.62 & 63.38 & 60.42 & 54.92 & 51.34 & 69.54 & 66.02 & 63.34 & 60.72 & 68.77 & 67.47 & 64.34 & 60.52 \\
 &  & Micro-F1 & 88.42 & 82.65 & 78.75 & 74.05 & 89.57 & 85.82 & 80.85 & 76.32 & 83.63 & 79.35 & 76.92 & 73.35 & 83.67 & 80.97 & 78.35 & 74.23 \\
 &  & Binary-F1 & 93.54 & 90.43 & 87.12 & 84.24 & 92.13 & 91.46 & 89.83 & 87.94 & 90.02 & 88.17 & 84.15 & 81.44 & 87.85 & 85.15 & 82.12 & 80.17 \\ \cline{2-19} 
 & \multirow{4}{*}{SDGNN} & AUC & 85.62 & 82.62 & 79.64 & 75.65 & 80.34 & 77.95 & 74.62 & 71.82 & 82.65 & 79.85 & 78.17 & 75.87 & 82.82 & 76.95 & 73.95 & 69.77 \\
 &  & Macro-F1 & 69.24 & 63.22 & 58.28 & 53.97 & 64.85 & 57.48 & 54.74 & 51.24 & 70.37 & 68.94 & 65.94 & 63.62 & 68.62 & 63.38 & 59.84 & 57.02 \\
 &  & Micro-F1 & 86.74 & 81.45 & 76.54 & 70.85 & 87.56 & 83.05 & 79.64 & 73.72 & 86.15 & 83.77 & 80.94 & 76.93 & 85.75 & 80.53 & 77.07 & 73.23 \\
 &  & Binary-F1 & 92.27 & 88.94 & 85.65 & 81.74 & 92.44 & 89.84 & 87.15 & 83.95 & 91.93 & 90.35 & 88.34 & 85.54 & 91.64 & 88.15 & 85.65 & 82.83 \\ \cline{2-19} 
 & \multirow{4}{*}{SDGCN} & AUC & 84.17 & 81.08 & 78.54 & 73.02 & 80.07 & 77.87 & 74.35 & 69.28 & 80.57 & 78.27 & 76.57 & 74.28 & 80.12 & 74.02 & 70.94 & 67.02 \\
 &  & Macro-F1 & 68.42 & 63.82 & 57.42 & 52.34 & 64.22 & 56.84 & 53.93 & 50.74 & 65.87 & 63.62 & 61.02 & 60.72 & 67.78 & 62.52 & 58.22 & 56.64 \\
 &  & Micro-F1 & 86.45 & 80.45 & 76.65 & 71.72 & 87.37 & 82.17 & 78.85 & 73.25 & 85.36 & 82.25 & 79.77 & 75.24 & 84.23 & 79.87 & 76.34 & 72.83 \\
 &  & Binary-F1 & 92.73 & 88.46 & 85.76 & 82.72 & 92.78 & 88.92 & 86.95 & 82.72 & 90.77 & 90.57 & 88.57 & 84.93 & 90.83 & 87.72 & 84.25 & 81.15 \\ \hline
\multirow{16}{*}{Attack-tolerant SGNN} & \multirow{4}{*}{RSGNN} & AUC & 79.58 & 74.57 & 69.27 & 66.07 & 74.59 & 70.68 & 61.94 & 60.22 & 75.32 & 74.14 & 72.52 & 69.85 & 71.88 & 65.68 & 63.72 & 59.72 \\
 &  & Macro-F1 & 65.78 & 58.44 & 54.82 & 51.98 & 58.42 & 53.27 & 50.18 & 48.32 & 67.34 & 64.37 & 62.05 & 59.67 & 64.88 & 58.17 & 55.52 & 52.12 \\
 &  & Micro-F1 & 79.74 & 71.37 & 67.84 & 64.27 & 75.65 & 68.05 & 66.02 & 63.05 & 77.33 & 73.45 & 70.65 & 68.35 & 74.47 & 67.15 & 63.75 & 60.05 \\
 &  & Binary-F1 & 87.64 & 81.55 & 79.05 & 76.22 & 85.36 & 79.87 & 78.65 & 76.34 & 85.45 & 82.37 & 80.13 & 78.33 & 82.22 & 77.64 & 74.63 & 71.62 \\ \cline{2-19} 
 & \multirow{4}{*}{SGCL} & AUC & 84.54 & 80.95 & 78.22 & 74.02 & 80.78 & 76.19 & 72.52 & 69.14 & 80.38 & 78.85 & 74.47 & 71.32 & 81.52 & 74.62 & 69.14 & 65.84 \\
 &  & Macro-F1 & 72.14 & 66.15 & 60.95 & 57.12 & 65.42 & 60.52 & 56.85 & 53.17 & 66.65 & 65.32 & 61.62 & 58.77 & 65.02 & 56.15 & 50.04 & 50.05 \\
 &  & Micro-F1 & 90.77 & 85.45 & 81.32 & 77.34 & 91.62 & 88.45 & 85.62 & 80.54 & 84.75 & 82.15 & 77.55 & 73.82 & 82.84 & 75.75 & 72.24 & 70.05 \\
 &  & Binary-F1 & 94.44 & 91.97 & 89.43 & 86.62 & 93.55 & 91.72 & 90.05 & 87.34 & 91.22 & 89.54 & 86.33 & 83.74 & 89.92 & 85.56 & 83.34 & 81.66 \\ \cline{2-19} 
 & \multirow{4}{*}{UGCL} & AUC & 86.07 & 83.24 & 79.55 & 76.58 & 82.62 & 78.47 & 74.82 & 72.48 & 83.57 & 81.22 & 78.75 & 77.15 & 84.74 & 77.75 & 73.62 & 69.15 \\
 &  & Macro-F1 & 75.54 & 68.67 & 64.07 & 59.48 & 67.23 & 62.22 & 57.72 & 54.22 & 68.42 & 67.84 & 66.37 & 64.62 & 69.44 & 66.54 & 63.92 & 58.75 \\
 &  & Micro-F1 & 91.53 & 86.65 & 82.37 & 77.55 & 92.12 & 88.35 & 84.55 & 79.95 & 87.23 & 86.75 & 85.35 & 83.25 & 87.35 & 85.36 & 83.04 & 78.23 \\
 &  & Binary-F1 & 95.34 & 92.35 & 89.67 & 86.45 & 95.76 & 93.67 & 91.32 & 87.54 & 92.84 & 92.45 & 91.55 & 90.15 & 92.75 & 91.53 & 90.14 & 87.07 \\ \cline{2-19} 
 & \multirow{4}{*}{BA-SGCL} & AUC & 87.78 & 84.72 & 81.38 & 79.22 & 84.65 & 79.92 & 77.12 & 74.74 & 85.27 & 84.48 & 80.32 & 78.82 & 85.65 & 80.12 & 77.05 & 76.47 \\
 &  & Macro-F1 & 78.18 & 69.88 & 64.52 & 61.15 & 69.04 & 65.23 & 59.98 & 57.47 & 72.07 & \textbf{73.25} & 70.37 & 67.77 & 74.82 & 73.24 & 68.47 & 65.85 \\
 &  & Micro-F1 & 92.27 & 85.93 & 81.37 & 77.46 & 92.94 & 90.34 & 84.32 & 80.85 & 87.15 & 86.05 & 84.54 & 82.15 & 87.44 & 87.05 & 83.94 & 78.14 \\
 &  & Binary-F1 & 95.74 & 91.85 & 88.93 & 86.33 & 96.23 & 94.58 & 91.63 & 88.05 & 92.75 & 91.84 & 90.87 & 89.35 & 92.65 & 91.06 & 90.57 & 86.34 \\ \hline \hline
\multirow{4}{*}{Ours} & \multirow{4}{*}{\framework} & AUC & \textbf{89.12} & \textbf{86.25} & \textbf{83.15} & \textbf{81.05} & \textbf{86.34} & \textbf{81.52} & \textbf{78.95} & \textbf{76.82} & \textbf{86.85} & \textbf{85.66} & \textbf{82.17} & \textbf{80.56} & \textbf{87.21} & \textbf{81.85} & \textbf{78.94} & \textbf{77.65} \\
 &  & Macro-F1 & \textbf{79.35} & \textbf{71.25} & \textbf{66.53} & \textbf{63.18} & \textbf{70.85} & \textbf{67.15} & \textbf{61.88} & \textbf{59.54} & \textbf{72.85} & 72.44 & \textbf{70.92} & \textbf{69.27} & \textbf{75.27} & \textbf{74.95} & \textbf{70.25} & \textbf{67.47} \\
 &  & Micro-F1 & \textbf{93.13} & \textbf{87.45} & \textbf{82.87} & \textbf{79.15} & \textbf{93.59} & \textbf{91.23} & \textbf{85.95} & \textbf{82.13} & \textbf{87.36} & \textbf{86.75} & \textbf{85.71} & \textbf{83.65} & \textbf{87.75} & \textbf{87.15} & \textbf{85.26} & \textbf{79.85} \\
 &  & Binary-F1 & \textbf{96.15} & \textbf{92.54} & \textbf{90.12} & \textbf{87.85} & \textbf{96.75} & \textbf{95.24} & \textbf{92.45} & \textbf{89.26} & \textbf{93.01} & \textbf{92.56} & \textbf{91.62} & \textbf{90.18} & \textbf{93.05} & \textbf{91.84} & \textbf{91.22} & \textbf{88.63} \\ \hline \hline
\end{tabular}%
}
\end{table*}

\subsection{Large Graphs}
\label{sec:large_graph}
In addition to the four commonly used datasets, we further evaluate the robustness of our method on large-scale graphs with random noise levels of 10\% and 20\%. Specifically, we consider MovieLens-1M (6040 nodes, 1,000,209 edges) and Amazon-Book (38,121 nodes, 1,960,674 edges), both containing over one million edges.
As shown in Table~\ref{tab:large_graph}, our approach presents strong practical effectiveness, even on these large graphs.
Notably, on the MovieLens-1M dataset, LRWSB requires over 54 hours for a single training run to filter transitive triangles containing exactly one unknown edge, highlighting the computational efficiency of our method.

\begin{figure}[!ht]
  \centering
  \begin{subfigure}{0.48\linewidth}
    \centering
    \includegraphics[width=\linewidth]{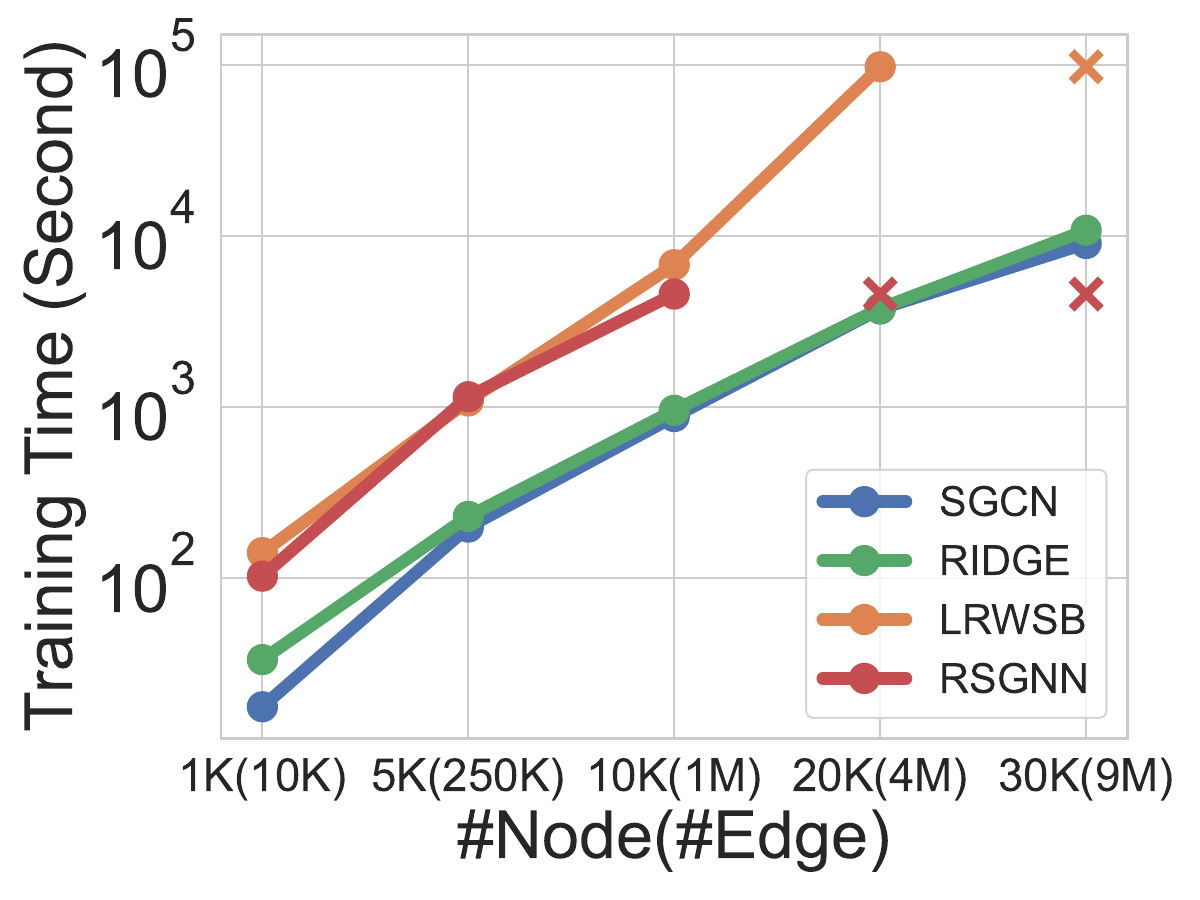}
    \vskip -5pt
    \caption{Time (s).}
    % \label{fig:node_cls}
  \end{subfigure}
  \begin{subfigure}{0.48\linewidth}
    \centering
    \includegraphics[width=\linewidth]{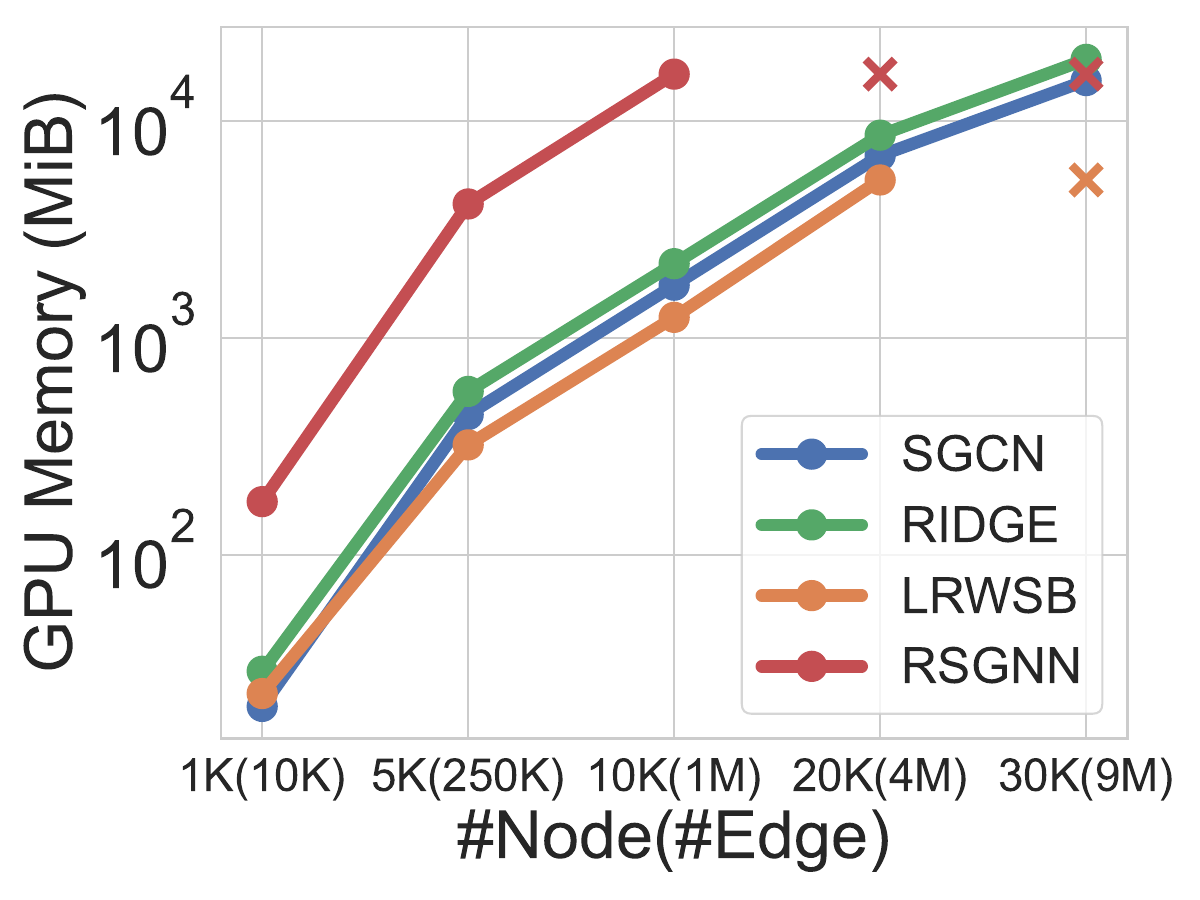}
    \vskip -5pt
    \caption{GPU Memory (MiB).}
    % \label{fig:link_pre}
  \end{subfigure} 
  \vskip -5pt
  \caption{\textbf{Scalability.} X for LRWSB is OOT. X for RSGNN indicates OOM.}
  \label{fig:scalability}
\end{figure}

\subsection{Scalability}
\label{sec:scalability}
Building on the high efficiency demonstrated in Fig.\ref{fig:training_time} to Fig.\ref{fig:training_mem_comparison}, we further assess the scalability of our method. Specifically, we adopt the synthetic SSBM dataset~\cite{cucuringu2019sponge}, following the same configuration used in MSGNN~\cite{he2022msgnn}, with a variable number of nodes and fixed parameters: $K = 5$, $p = 0.01$, and $\rho = 1.5$, where $K$ denotes the number of clusters, $p$ is the edge probability (for both signs), and $\rho$ indicates the approximate ratio between the largest and smallest cluster sizes.
As shown in Fig.~\ref{fig:scalability}, our method exhibits strong scalability, with both training time and GPU memory usage increasing approximately linearly with that of SGCN. In particular, even for a graph containing 30,000 nodes and 9 million edges, our method requires only 18.9 GiB of GPU memory.
While LRWSB consumes less GPU memory due to off-loaded preprocessing, it incurs significantly higher preprocessing time, limiting its overall efficiency.

\section{Adversarial Noise}
To further assess the robustness of our method, we introduce adversarial attack for model validation. Conventional adversarial attack strategies such as Nettack~\cite{zugner2018adversarial} and Metattack~\cite{zugner2019adversarial} assume access to node labels and features, which are typically unavailable in widely used signed graph datasets. To address this, we adopt Balance Attack~\cite{zhou2024black}, which perturbs the graph structure by reducing its balance degree, aligning with the characteristics of signed graphs.

\noindent \textbf{Experiment setup.}
For a fair comparison with the state-of-the-art adversarial defense method BA-SGCL~\cite{zhou2024adversarially}, the experimental setup under adversarial noise differs slightly from that under random noise. Specifically, SDGCN~\cite{ko2023spectral} is employed as the signed graph encoder. Results are averaged over five runs on a fixed dataset split, with noise varying from 5\% to 20\% of the total edges. Evaluation metrics include AUC, Micro-F1, Macro-F1, and Binary-F1.

\noindent \textbf{Baselines.}
We compare against SiNE~\cite{wang2017signed}, SGCN~\cite{derr2018signed}, SNEA~\cite{li2020learning}, BESIDE~\cite{chen2018bridge}, SDGNN~\cite{huang2021sdgnn}, SDGCN~\cite{ko2023spectral}, RSGNN~\cite{zhang2023rsgnn}, SGCL~\cite{shu2021sgcl}, UGCL~\cite{ko2023universal}, and BA-SGCL~\cite{zhou2024adversarially}. The results of baselines are derived from BA-SGCL~\cite{zhou2024adversarially}.

\noindent \textbf{Results.}
Table~\ref{tab:ad_results} presents the performance under adversarial noise. \framework consistently surpasses all baselines across varying noise levels, underscoring the robustness of our information-theoretic design against adversarial attacks. Notably, Balance Attack operates under the assumption that clean signed graphs exhibit a high balance degree, a metric rooted in classical balance theory. Prior work has demonstrated that balance theory is equivalent to a simplified structural assumption: nodes can be partitioned into two disjoint subsets, with positive links within subsets and negative links across subsets. However, this assumption is overly idealized and rarely holds in real-world signed graphs. This discrepancy may explain why our method maintains superior robustness under adversarial perturbations: rather than explicitly enforcing a higher balance degree, our approach is guided by the GIB theory, enabling more generalizable and resilient representations.

\section{Link Deletion and Addition}
\label{sec:link_del_add}
Besides the random link sign flipping, here, we further validate our method under a variety of noise types, including random link deletion, random positive link deletion, random negative link deletion, random link addition, random positive link addition, and random negative link addition.
This comprehensive validation ensures the robustness of our method under different types of perturbations in the network structure. In the experiment, the perturbation rates are among $\{20\%, 40\%, 60\%, 80\%\}$. Generally, \framework exhibits superior performance compared to the corresponding backbones under a variety of noise conditions.

\subsection{Random Link Deletion}
\begin{figure*}[!ht]
  \centering
  \begin{subfigure}{\linewidth}
    \centering
    \includegraphics[width=0.8\linewidth]{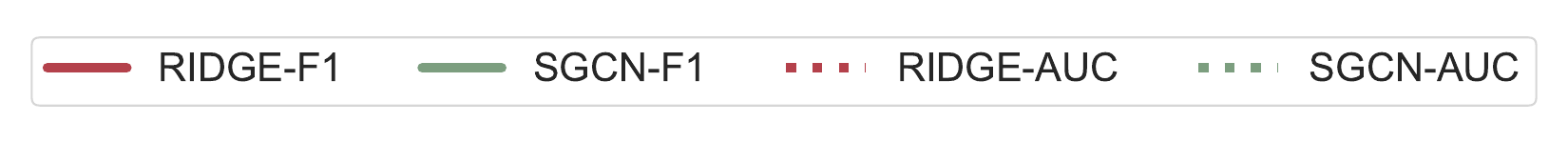}
  \end{subfigure} \\
  \vskip -0.1in
  \begin{subfigure}{0.24\linewidth}
    \centering
    \includegraphics[width=\linewidth]{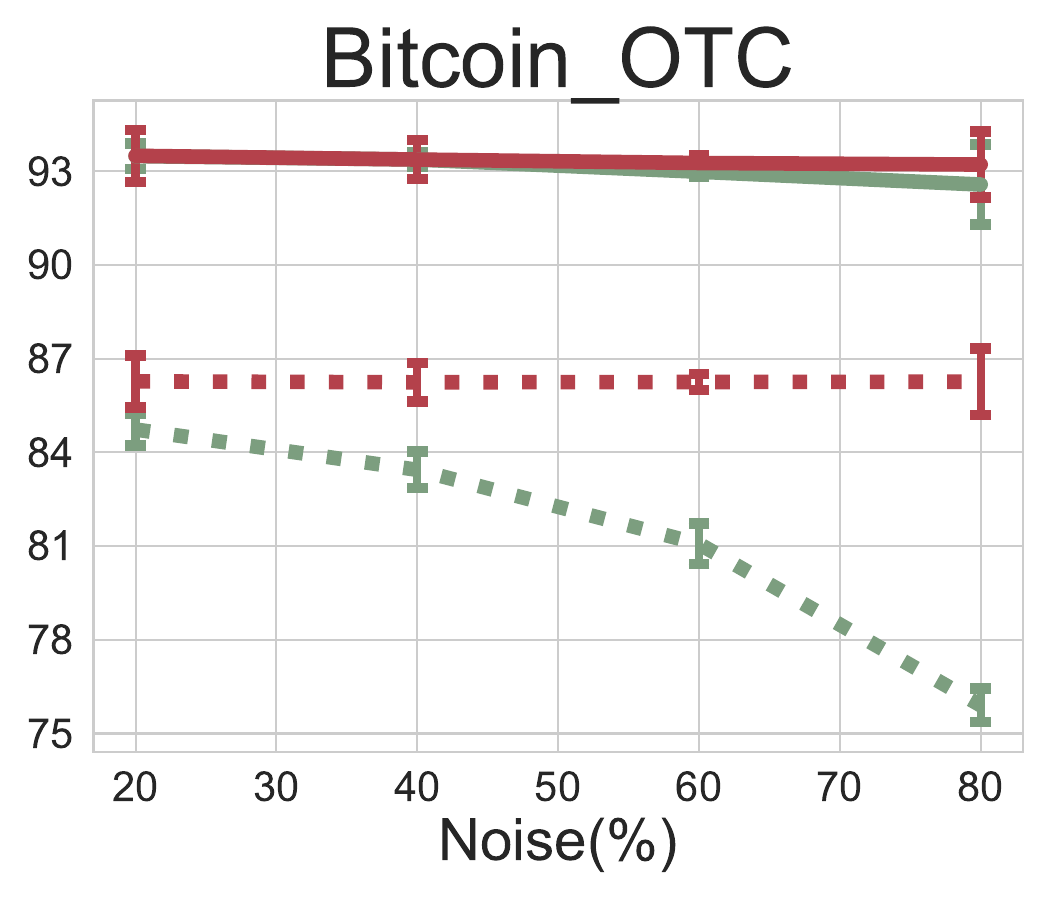}
    % \caption{Bitcoin\_OTC.}
    % \label{fig:node_cls}
  \end{subfigure}
  \begin{subfigure}{0.24\linewidth}
    \centering
    \includegraphics[width=\linewidth]{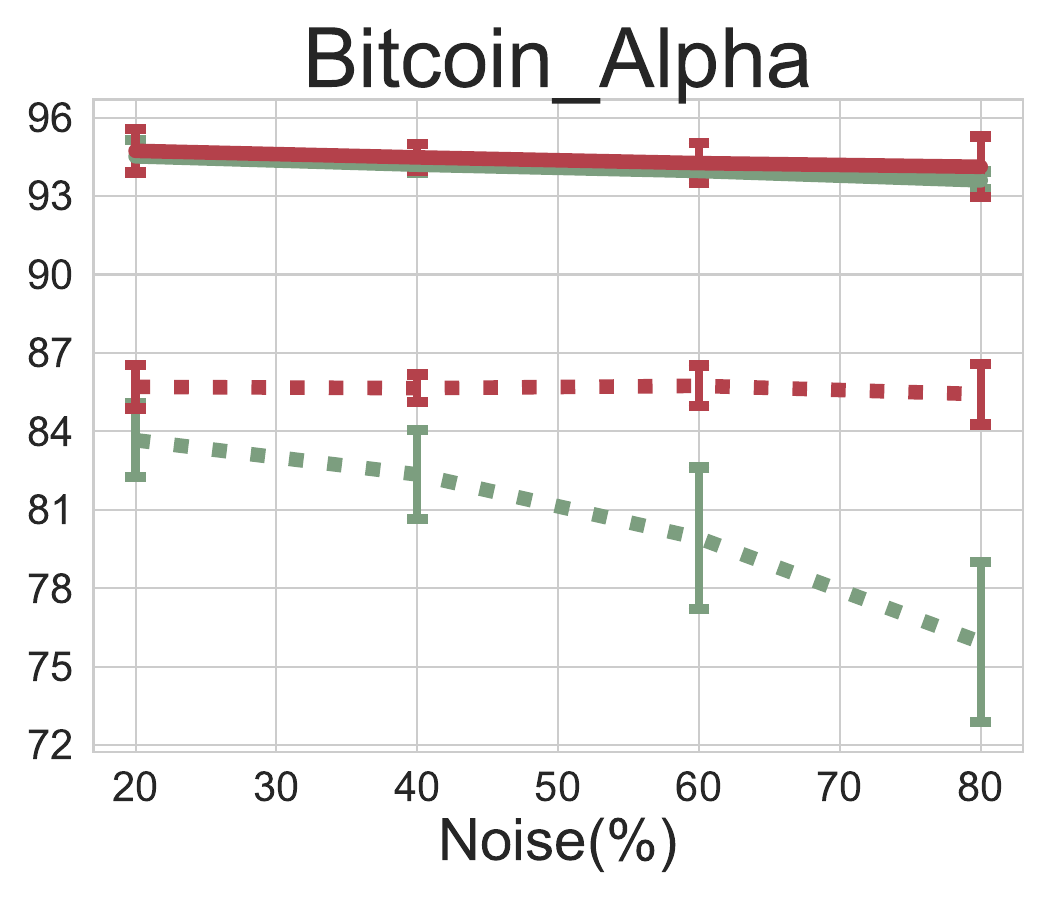}
    % \caption{Bitcoin\_Alpha}
    % \label{fig:link_pre}
  \end{subfigure} 
  \begin{subfigure}{0.24\linewidth}
    \centering
    \includegraphics[width=\linewidth]{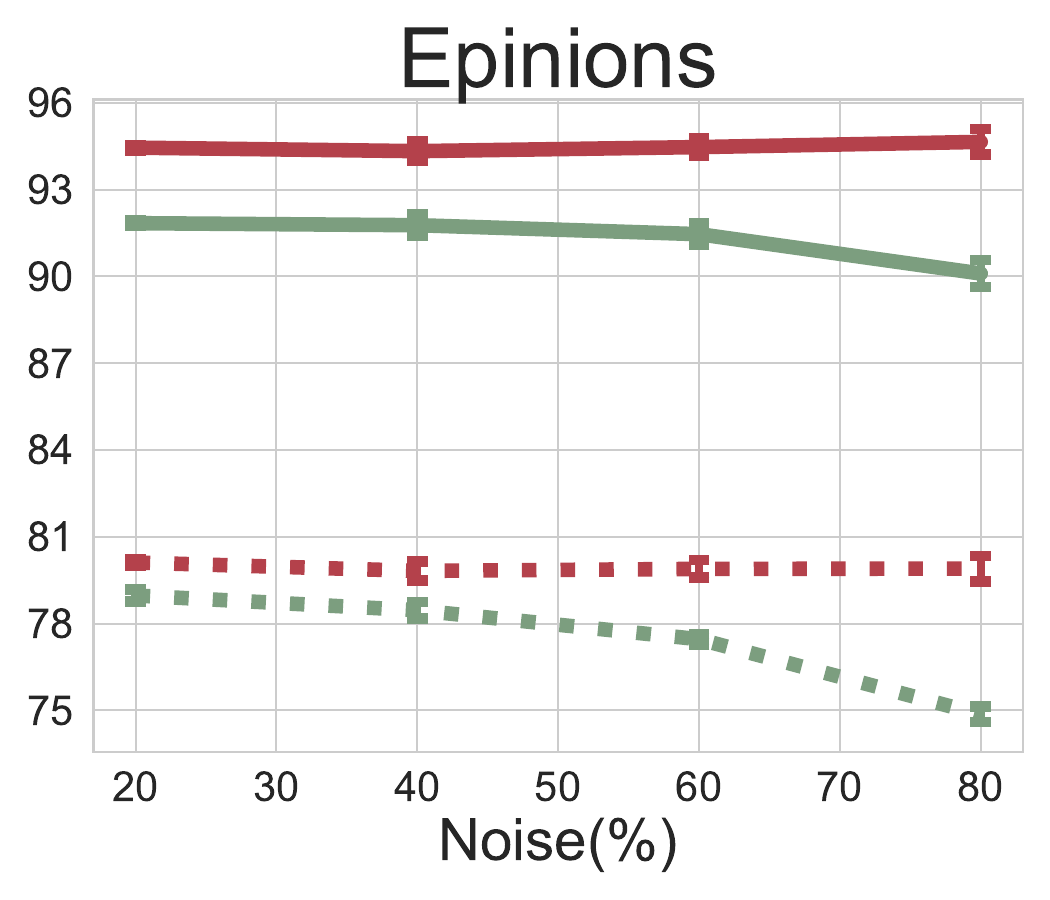}
    % \caption{Epinions.}
    % \label{fig:link_sign_pre}
  \end{subfigure}
  \begin{subfigure}{0.24\linewidth}
    \centering
    \includegraphics[width=\linewidth]{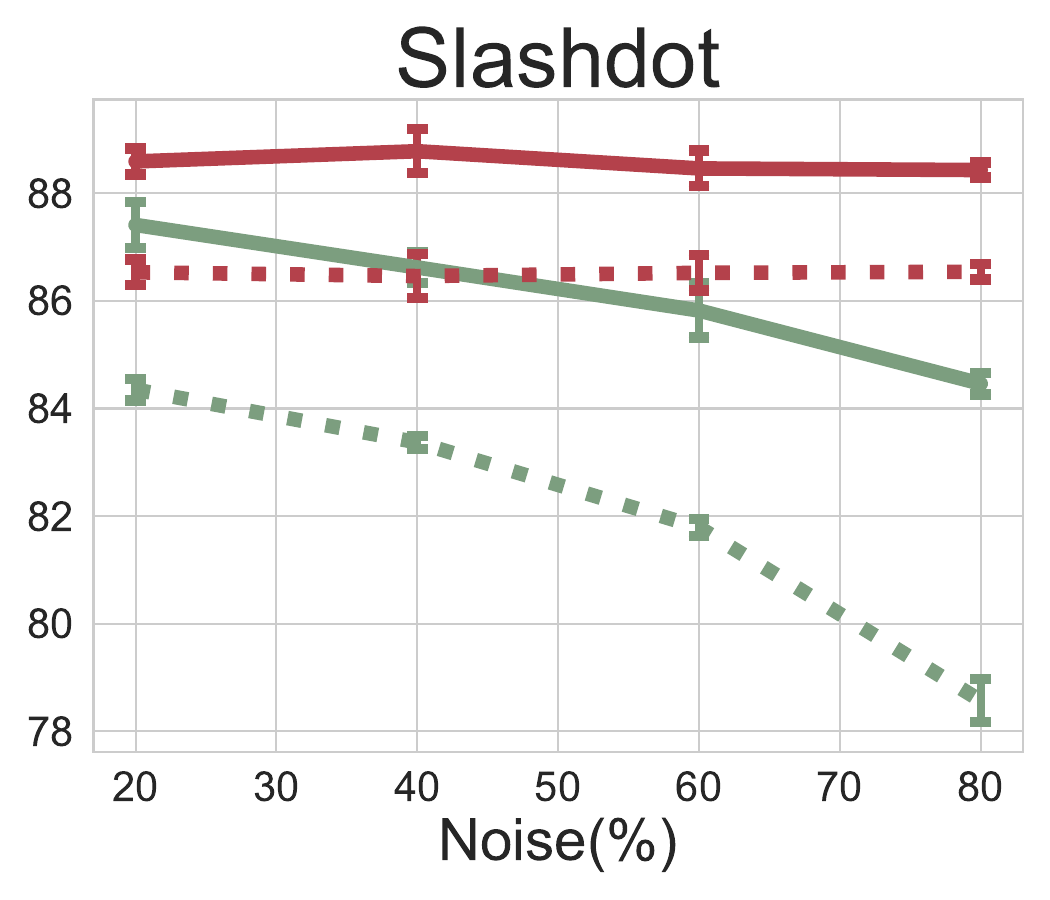}
    % \caption{Slashdot.}
  \end{subfigure} \\ \vskip -0.1in
  \begin{subfigure}{\linewidth}
    \centering
    \caption{\framework with SGCN as backbone under random link deletion.}
    \label{fig:del_sgcn}
  \end{subfigure} \\

  \begin{subfigure}{\linewidth}
    \centering
    \includegraphics[width=0.8\linewidth]{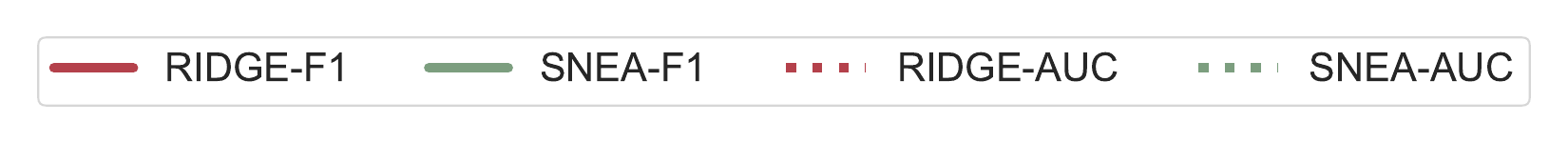}
  \end{subfigure} \\
  \vskip -0.1in
  \begin{subfigure}{0.24\linewidth}
    \centering
    \includegraphics[width=\linewidth]{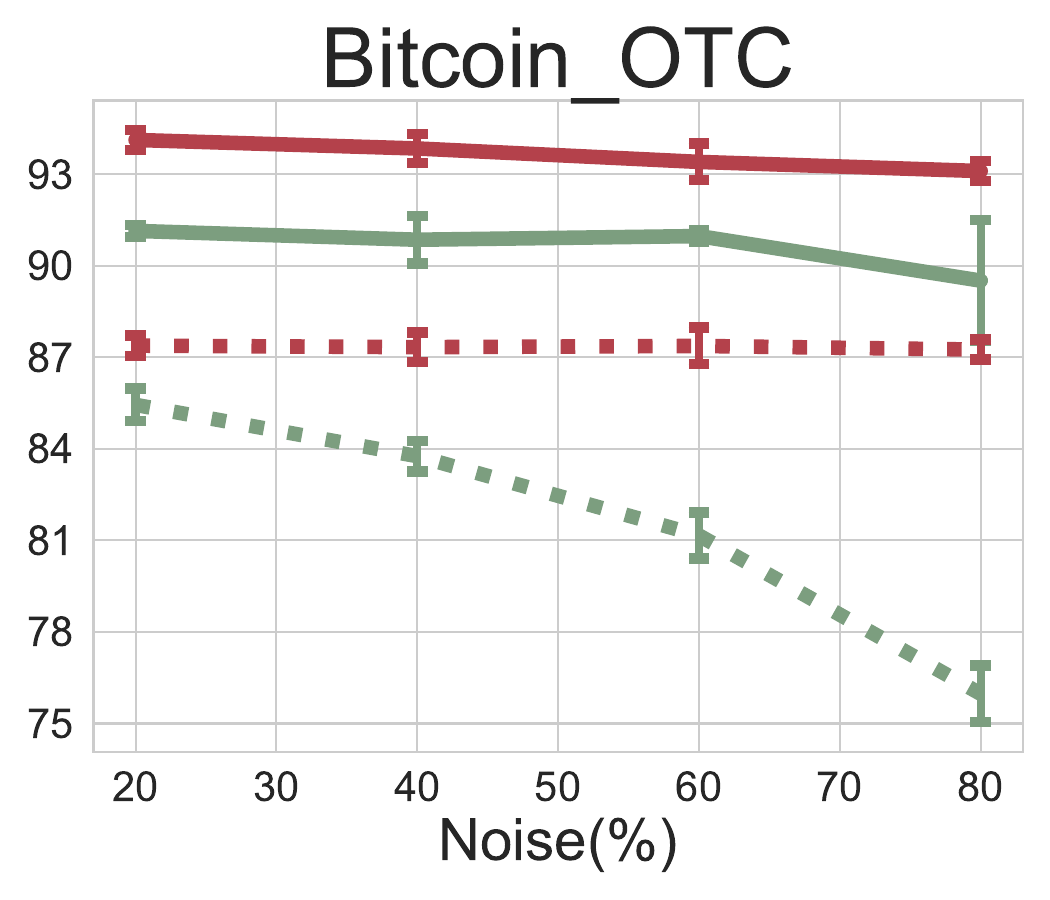}
    % \caption{Bitcoin\_OTC.}
    % \label{fig:node_cls}
  \end{subfigure}
  \begin{subfigure}{0.24\linewidth}
    \centering
    \includegraphics[width=\linewidth]{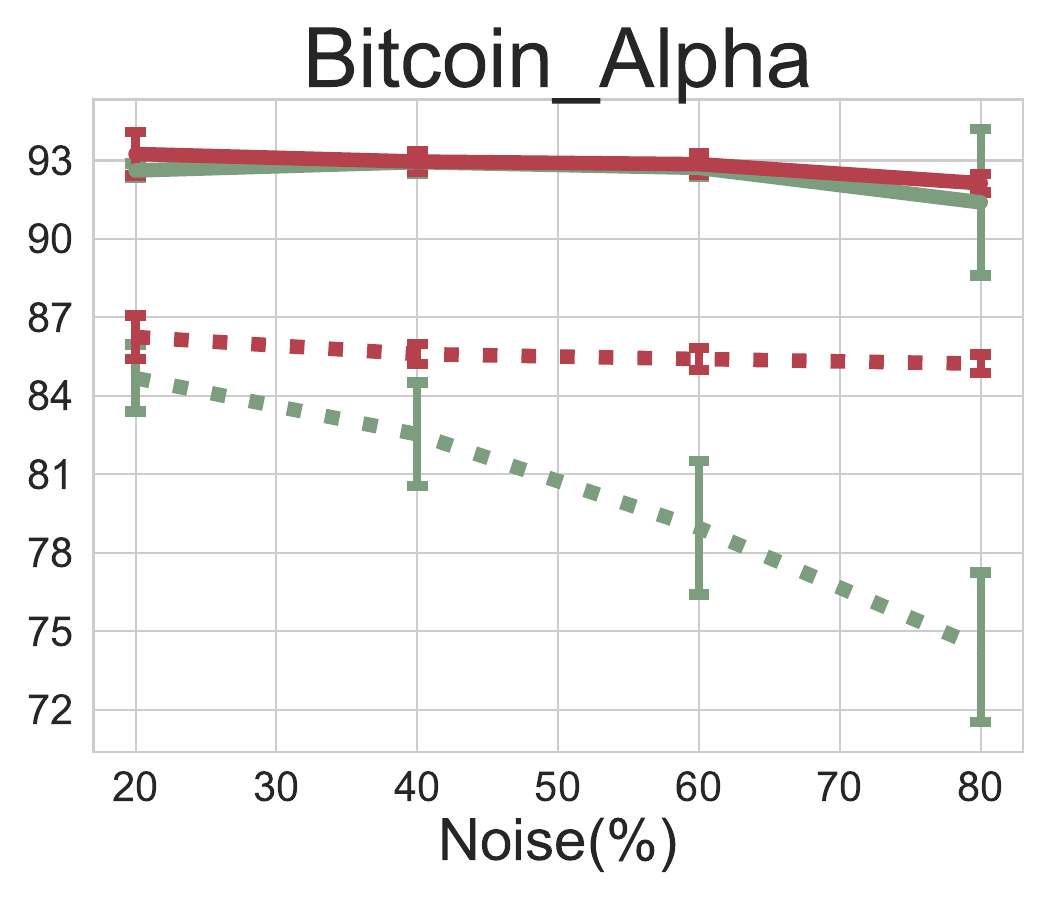}
    % \caption{Bitcoin\_Alpha}
    % \label{fig:link_pre}
  \end{subfigure} 
  \begin{subfigure}{0.24\linewidth}
    \centering
    \includegraphics[width=\linewidth]{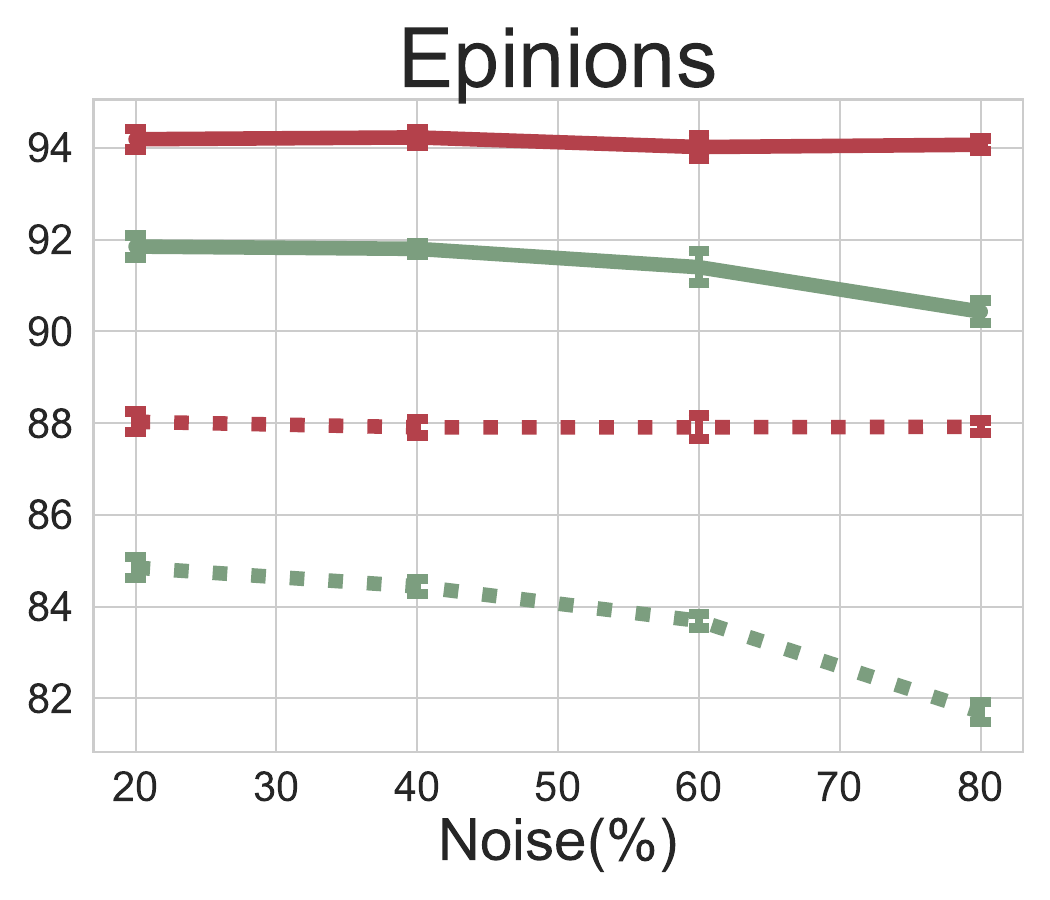}
    % \caption{Epinions.}
    % \label{fig:link_sign_pre}
  \end{subfigure}
  \begin{subfigure}{0.24\linewidth}
    \centering
    \includegraphics[width=\linewidth]{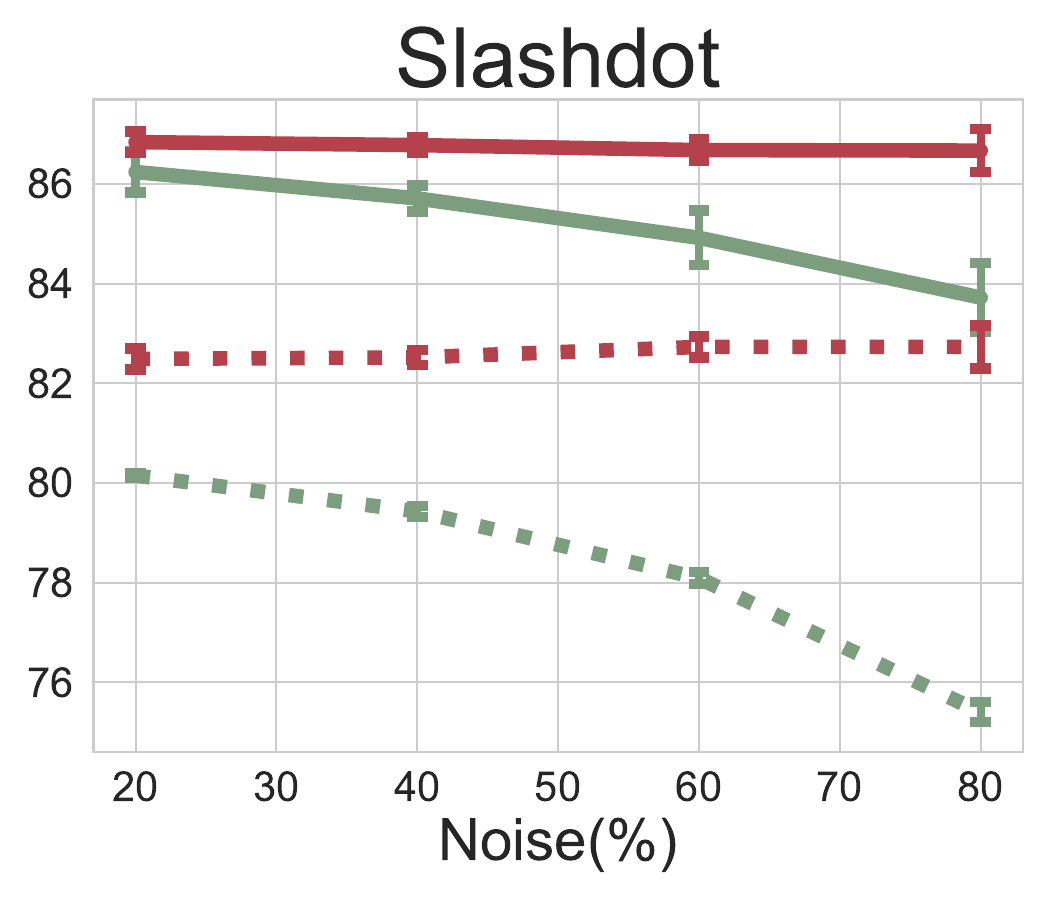}
    % \caption{Slashdot.}
  \end{subfigure} \\ \vskip -0.1in
  \begin{subfigure}{\linewidth}
    \centering  
    \caption{\framework with SNEA as backbone under random link deletion.}
    \label{fig:del_snea}
  \end{subfigure}
  \vskip -0.1in
  \caption{\textbf{Results of \framework with different backbones under random link deletion.}}
  \label{fig:link_del}
\end{figure*}

Given a signed graph $\mathcal{G}=\{\mathcal{U}, \mathcal{E}^+, \mathcal{E}^-\}$, a subset of links is randomly selected from the input signed graph and their signs are deleted. 
The proportion of links selected corresponds to the level of noise introduced to the graph. 
Specifically, for a given noise ratio $\gamma$, the noisy adjacency matrix $\tilde{A}$ is generated by multiplying each element in $A$ by 0 with probability $\gamma$, such that $\tilde{A} = \text{NoiseMask}\,\odot\, A$. This ensures that $\gamma = \sfrac{|\mathbf{nonzero}(A)| - |\mathbf{nonzero}(\tilde{A})|}{|\mathbf{nonzero}(A)|}$.
Results are shown in Fig.~\ref{fig:link_del}.

\subsection{Random Positive Link Deletion}
\begin{figure*}[!ht]
  \centering
  \begin{subfigure}{\linewidth}
    \centering
    \includegraphics[width=0.8\linewidth]{figs/legend-SGCN.pdf}
  \end{subfigure} \\
  \vskip -0.1in
  \begin{subfigure}{0.24\linewidth}
    \centering
    \includegraphics[width=\linewidth]{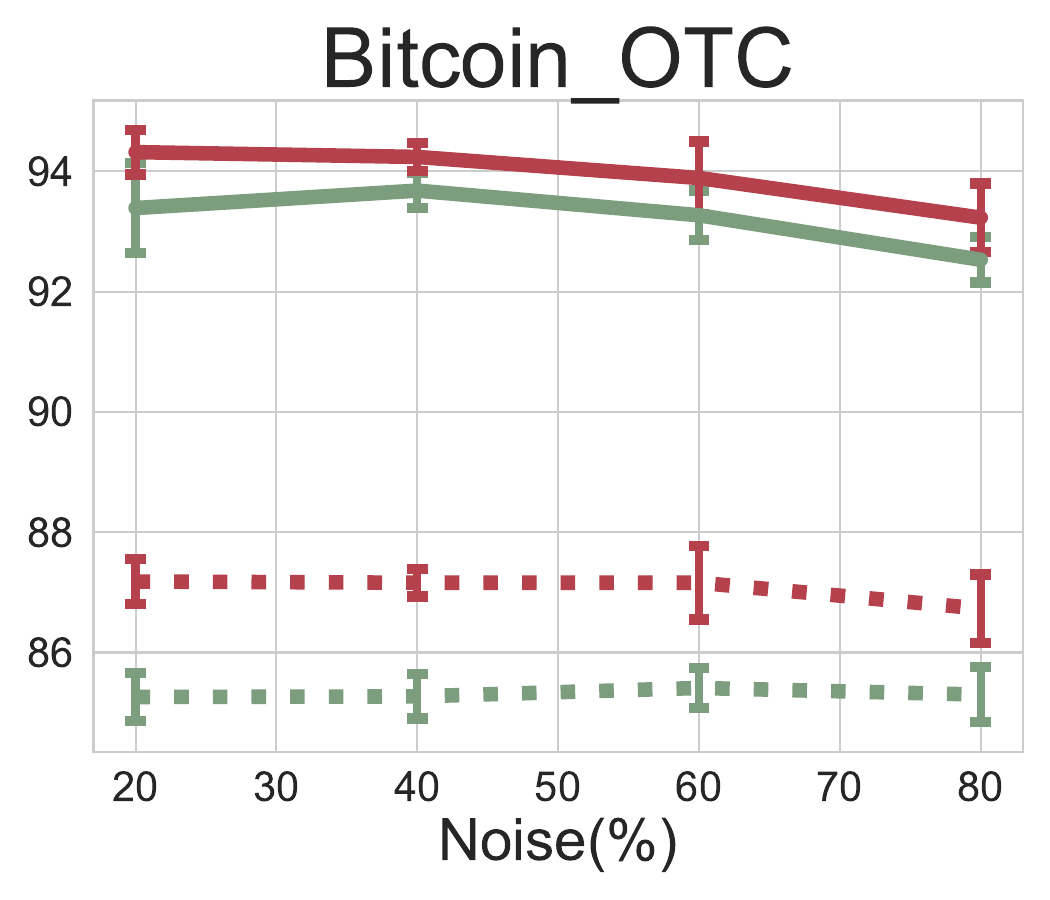}
    % \caption{Bitcoin\_OTC.}
    % \label{fig:node_cls}
  \end{subfigure}
  \begin{subfigure}{0.24\linewidth}
    \centering
    \includegraphics[width=\linewidth]{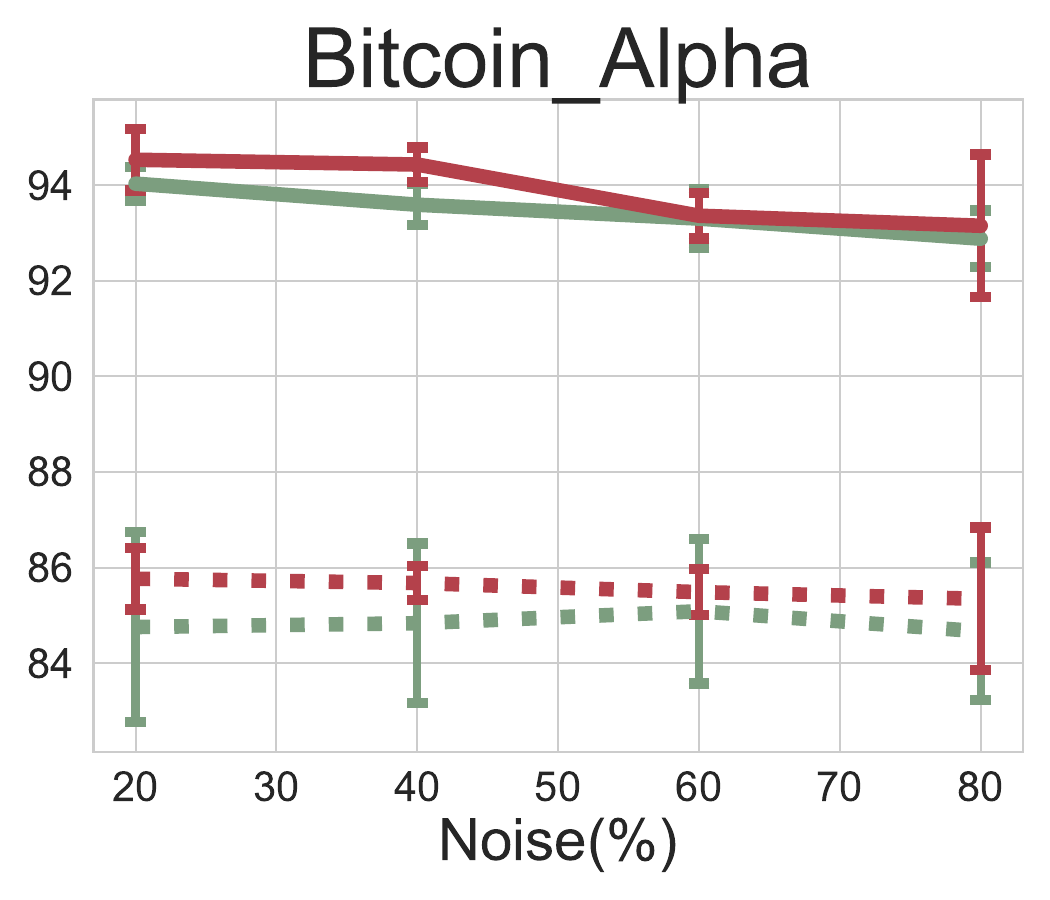}
    % \caption{Bitcoin\_Alpha}
    % \label{fig:link_pre}
  \end{subfigure} 
  \begin{subfigure}{0.24\linewidth}
    \centering
    \includegraphics[width=\linewidth]{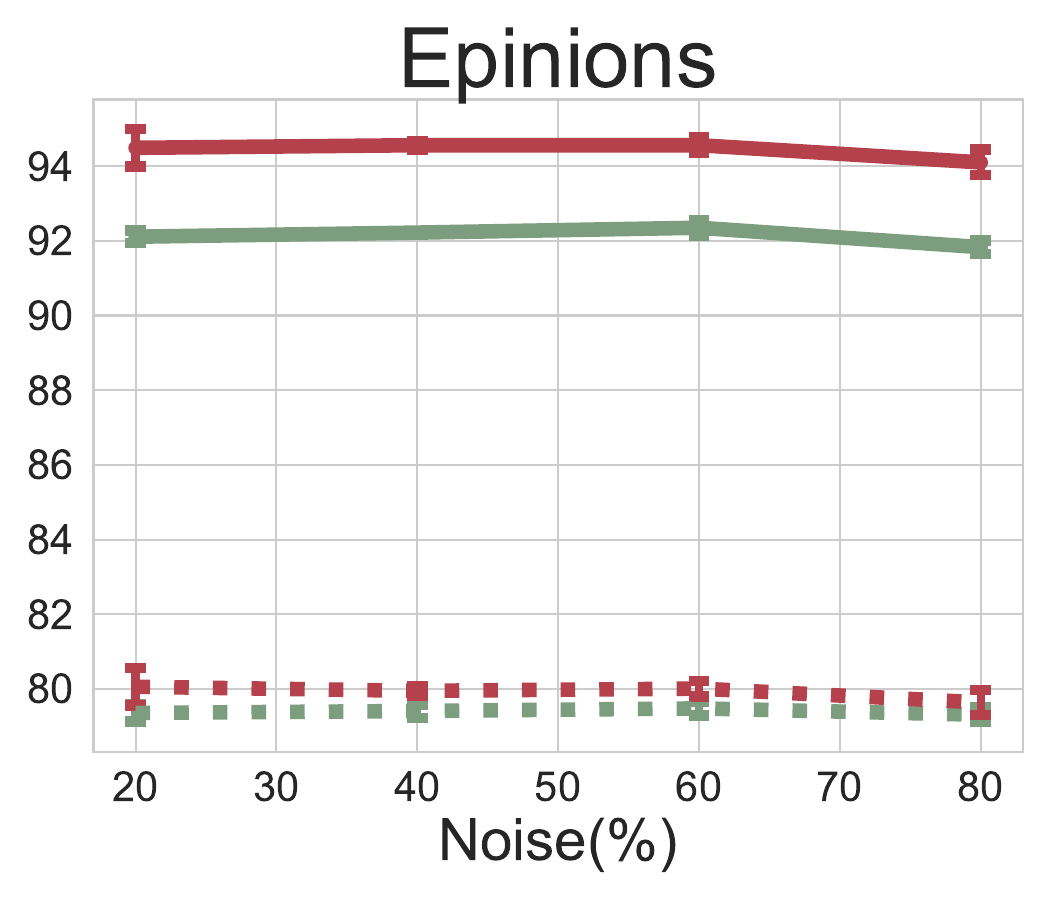}
    % \caption{Epinions.}
    % \label{fig:link_sign_pre}
  \end{subfigure}
  \begin{subfigure}{0.24\linewidth}
    \centering
    \includegraphics[width=\linewidth]{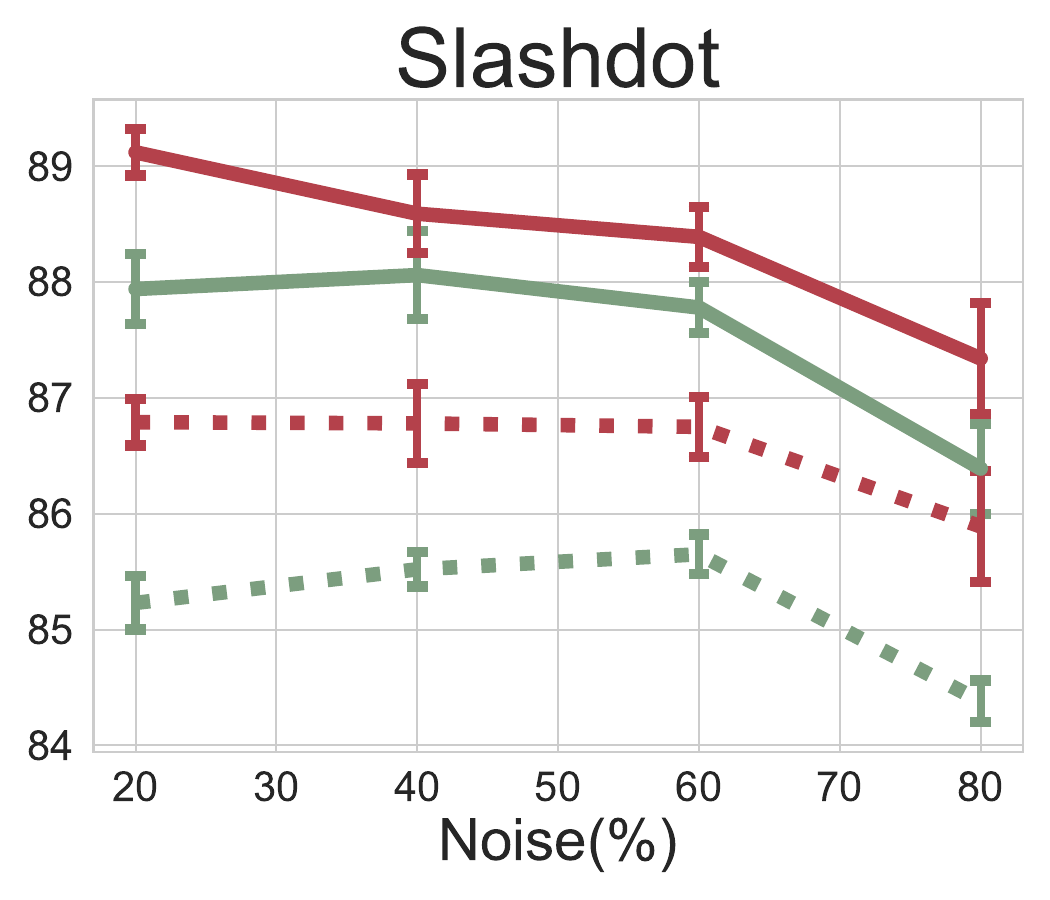}
    % \caption{Slashdot.}
  \end{subfigure} \\ \vskip -0.1in
  \begin{subfigure}{\linewidth}
    \centering
    \caption{\framework with SGCN as backbone under random positive link deletion.}
    \label{fig:pos_del_sgcn}
  \end{subfigure} \\

  \begin{subfigure}{\linewidth}
    \centering
    \includegraphics[width=0.8\linewidth]{figs/legend-SNEA.pdf}
  \end{subfigure} \\
  \vskip -0.1in
  \begin{subfigure}{0.24\linewidth}
    \centering
    \includegraphics[width=\linewidth]{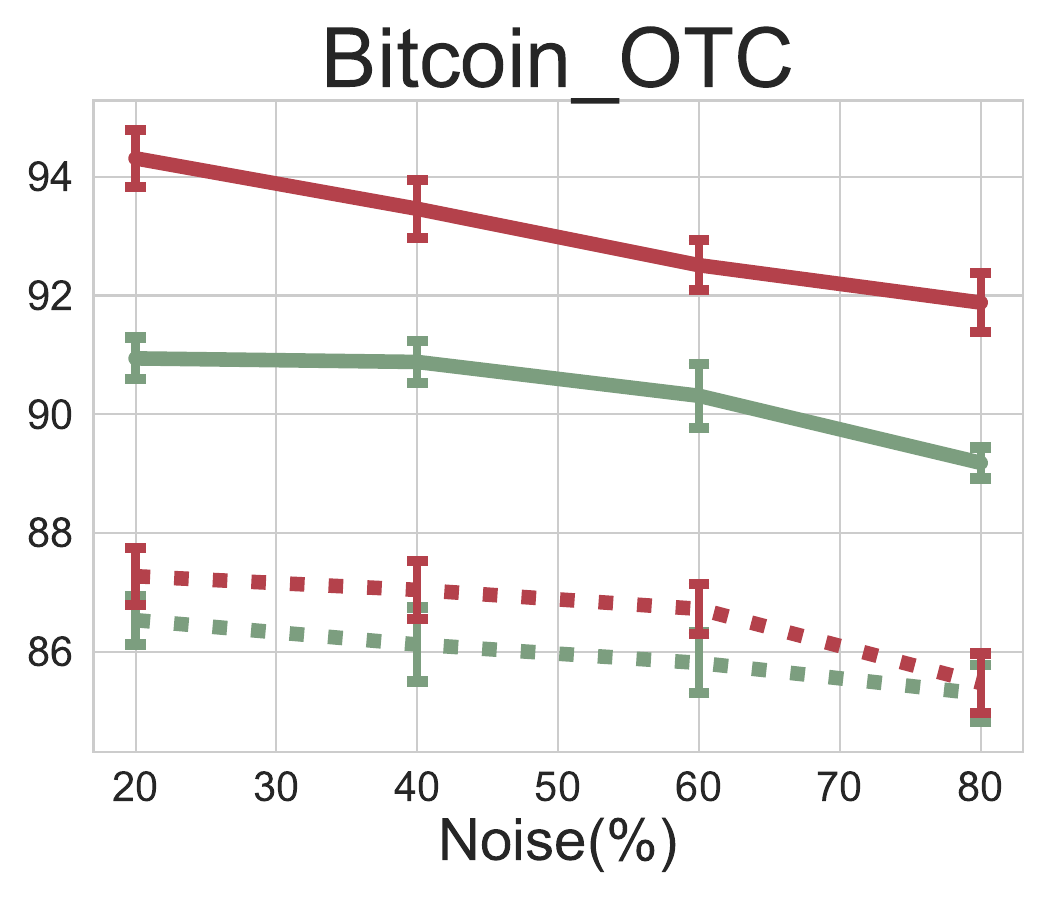}
    % \caption{Bitcoin\_OTC.}
    % \label{fig:node_cls}
  \end{subfigure}
  \begin{subfigure}{0.24\linewidth}
    \centering
    \includegraphics[width=\linewidth]{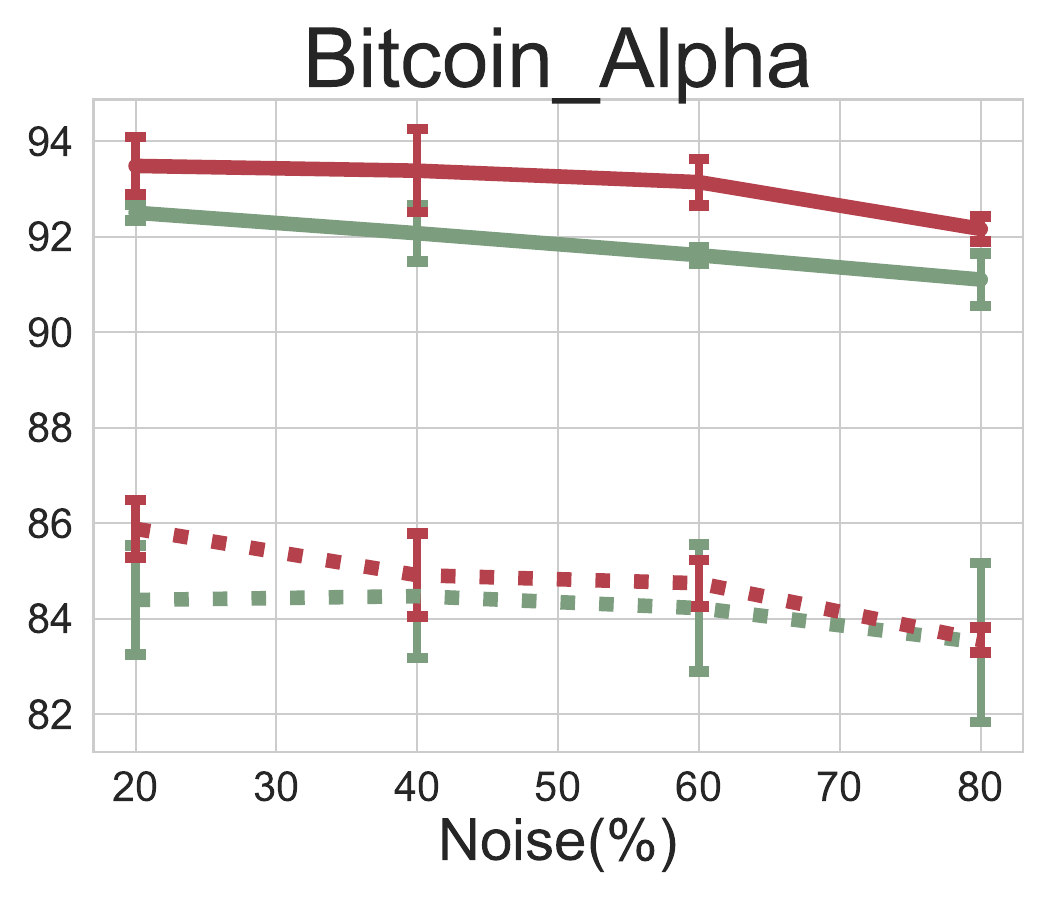}
    % \caption{Bitcoin\_Alpha}
    % \label{fig:link_pre}
  \end{subfigure} 
  \begin{subfigure}{0.24\linewidth}
    \centering
    \includegraphics[width=\linewidth]{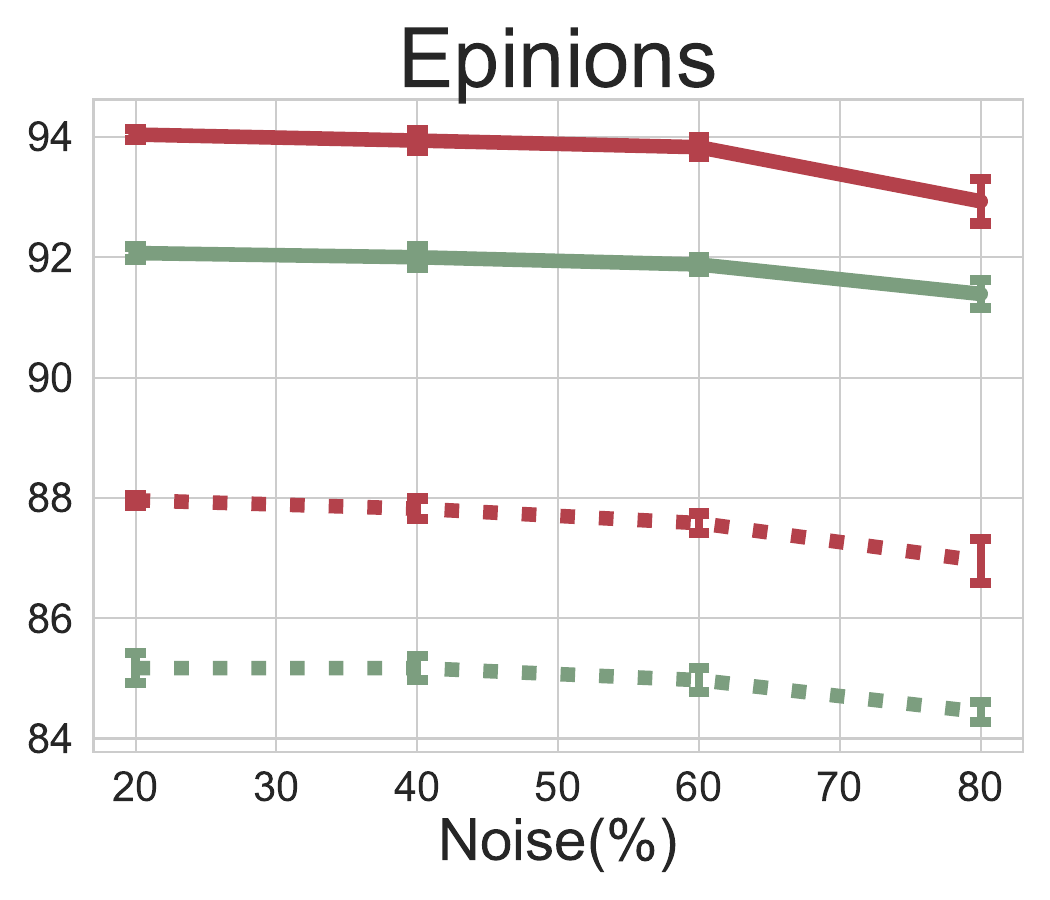}
    % \caption{Epinions.}
    % \label{fig:link_sign_pre}
  \end{subfigure}
  \begin{subfigure}{0.24\linewidth}
    \centering
    \includegraphics[width=\linewidth]{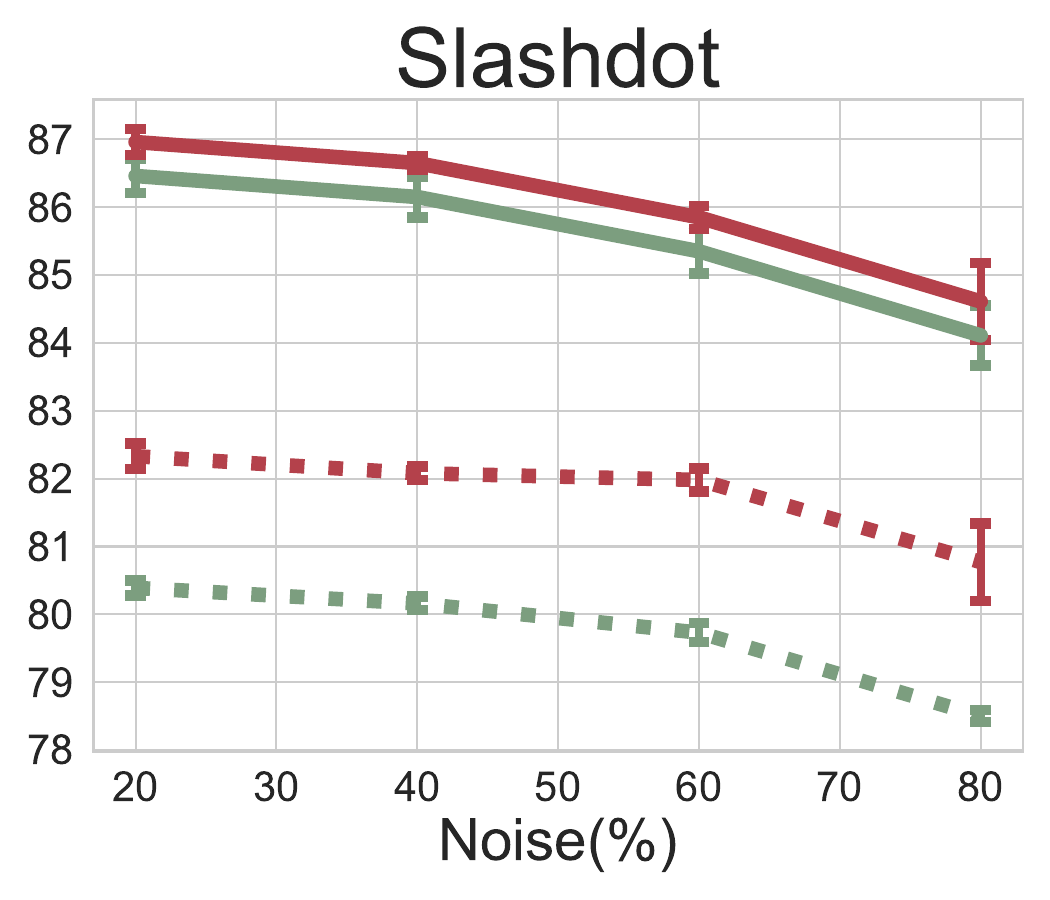}
    % \caption{Slashdot.}
  \end{subfigure} \\ \vskip -0.1in
  \begin{subfigure}{\linewidth}
    \centering  
    \caption{\framework with SNEA as backbone under random positive link deletion.}
    \label{fig:pos_del_snea}
  \end{subfigure}
   \vskip -0.1in
  \caption{\textbf{Results of \framework with different backbones under random positive link deletion.}}
  \label{fig:pos_link_del}
\end{figure*}

Given a signed graph $\mathcal{G}=\{\mathcal{U}, \mathcal{E}^+, \mathcal{E}^-\}$, a subset of positive links is randomly selected from the input signed graph and their signs are deleted. 
For simplicity, $A^+\in\{0, 1\}$ derived from $\mathcal{E}^+$ denotes the positive matrix, $A^-\in\{0, -1\}$ derived from $\mathcal{E}^-$ denotes the negative matrix and $A = A^+ + A^-$.
The proportion of links selected corresponds to the level of noise introduced to the graph. Specifically, for a given noise ratio $\gamma$, the noisy adjacency matrix $\tilde{A}$ is generated by multiplying each element in $A^+$ by 0 with probability $\gamma$, such that $\tilde{A} = \text{NoiseMask}\,\odot\, A^+ + A^-$. This ensures that $\gamma = \sfrac{|\mathbf{nonzero}(A)| - |\mathbf{nonzero}(\tilde{A})|}{|\mathbf{nonzero}(A^+)|}$.
Results are shown in Fig.~\ref{fig:pos_link_del}.

\begin{figure*}[!ht]
  \centering
  \begin{subfigure}{\linewidth}
    \centering
    \includegraphics[width=0.8\linewidth]{figs/legend-SGCN.pdf}
  \end{subfigure} \\
  \vskip -0.1in
  \begin{subfigure}{0.24\linewidth}
    \centering
    \includegraphics[width=\linewidth]{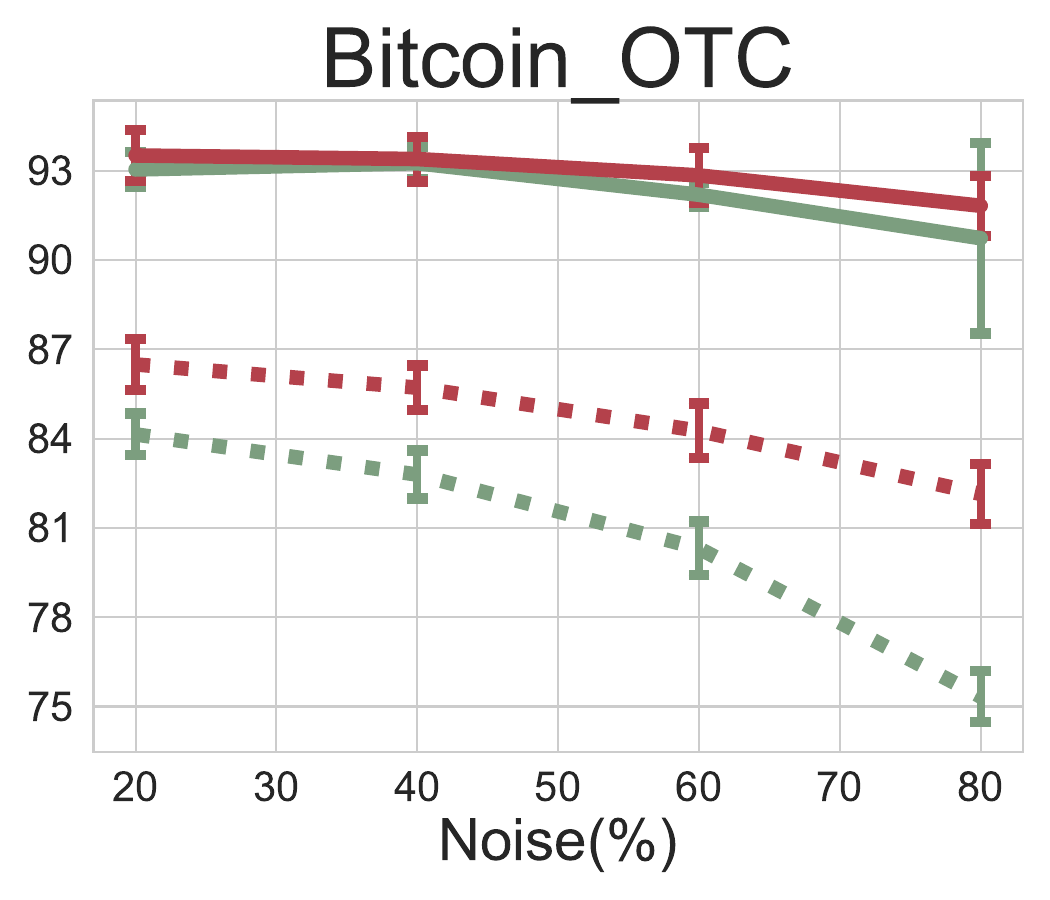}
    % \caption{Bitcoin\_OTC.}
    % \label{fig:node_cls}
  \end{subfigure}
  \begin{subfigure}{0.24\linewidth}
    \centering
    \includegraphics[width=\linewidth]{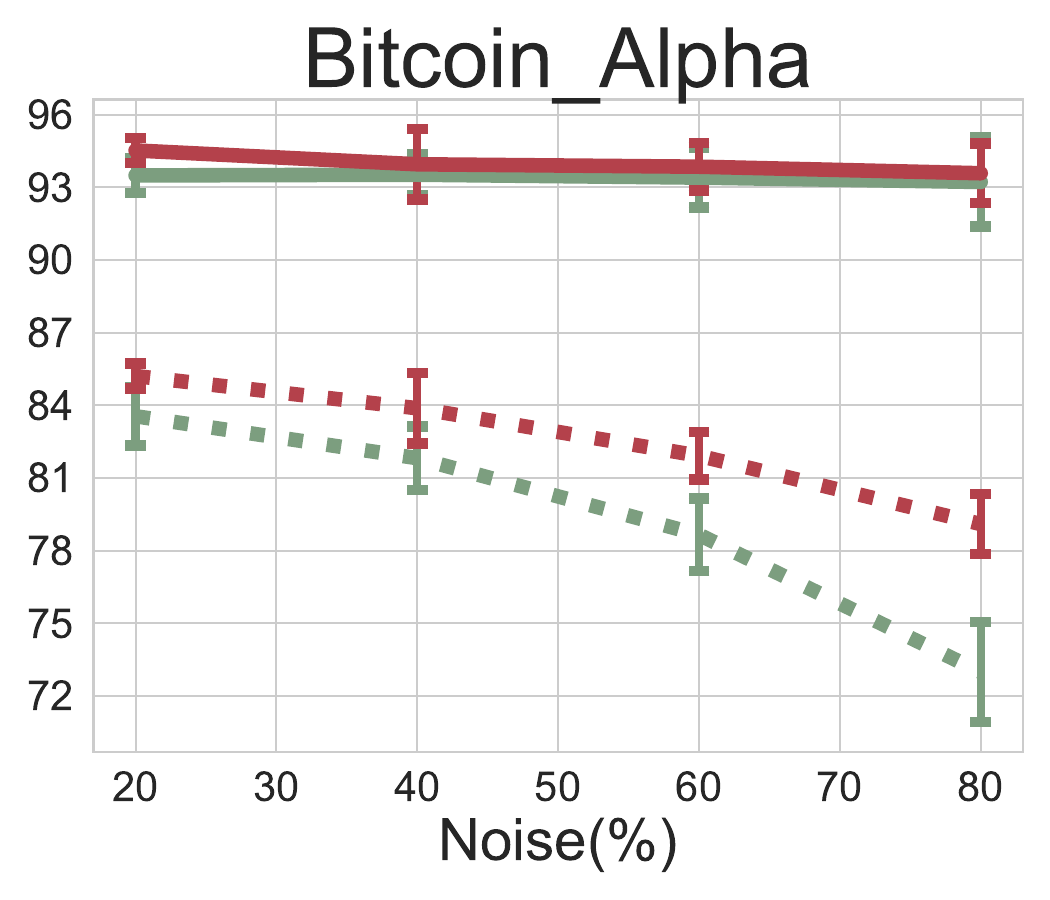}
    % \caption{Bitcoin\_Alpha}
    % \label{fig:link_pre}
  \end{subfigure} 
  \begin{subfigure}{0.24\linewidth}
    \centering
    \includegraphics[width=\linewidth]{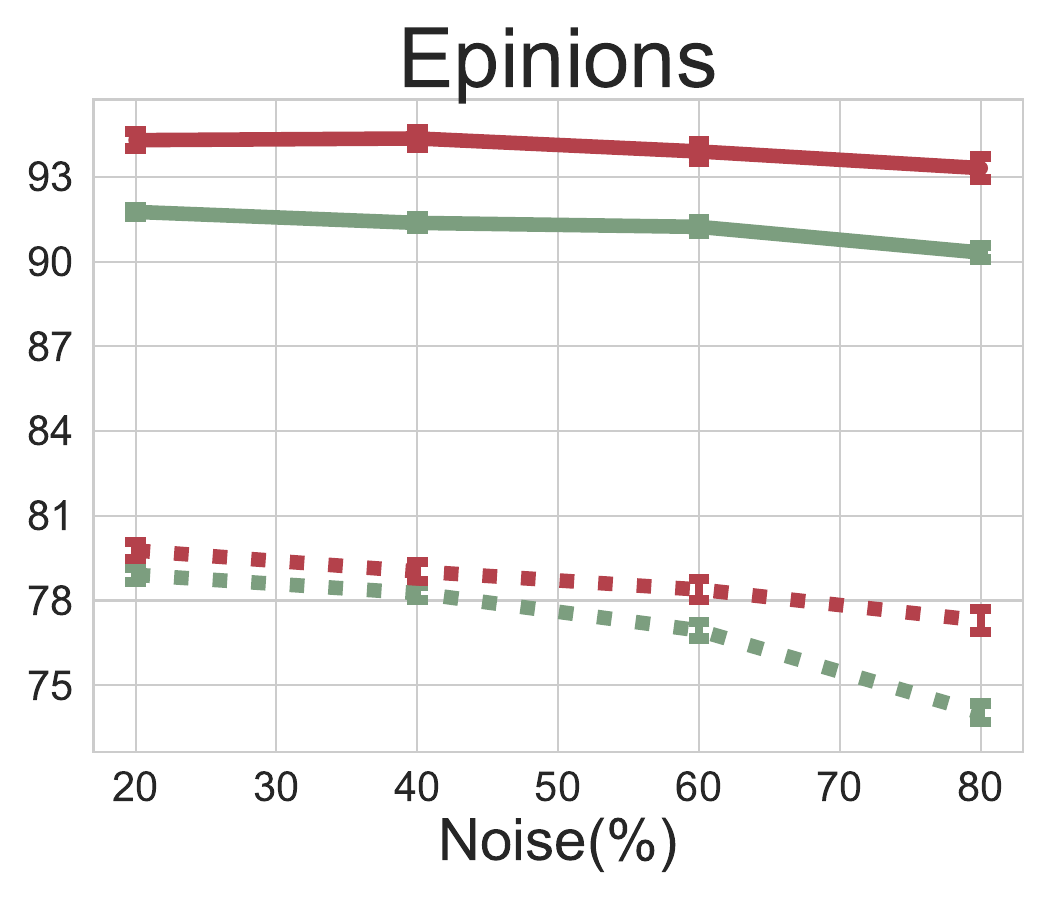}
    % \caption{Epinions.}
    % \label{fig:link_sign_pre}
  \end{subfigure}
  \begin{subfigure}{0.24\linewidth}
    \centering
    \includegraphics[width=\linewidth]{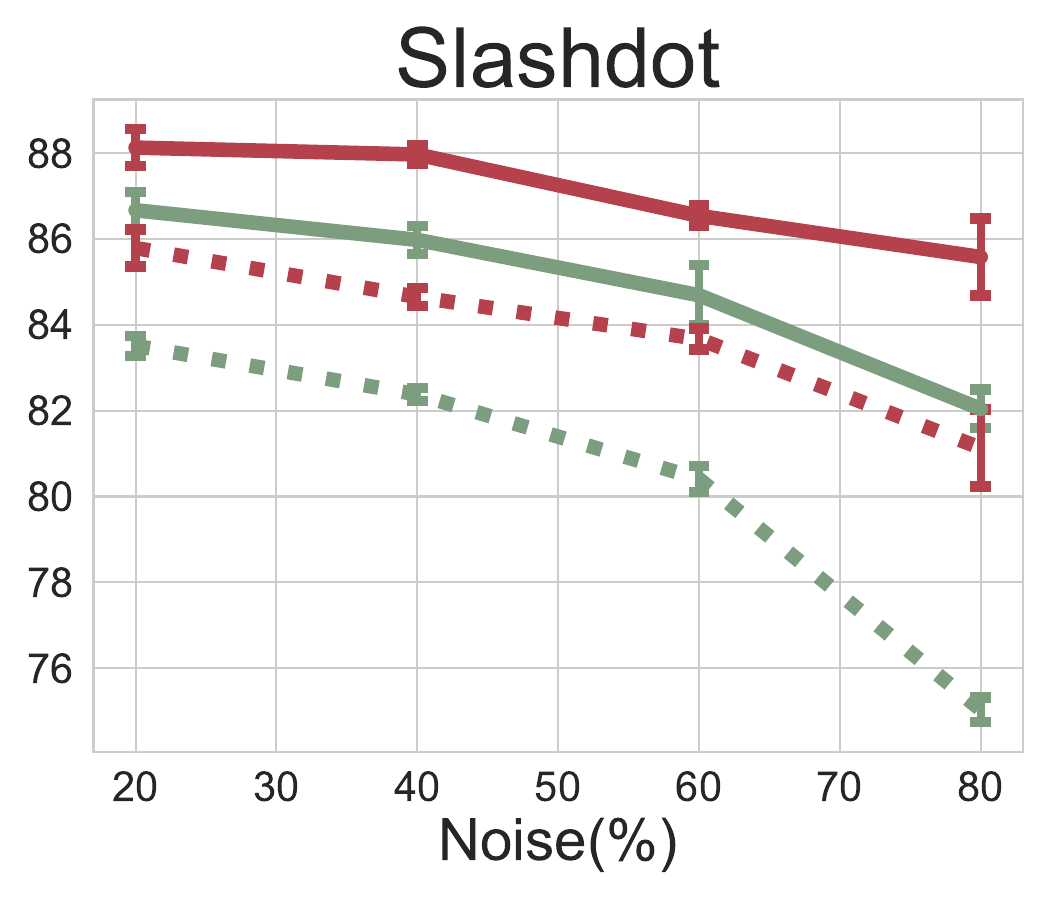}
    % \caption{Slashdot.}
  \end{subfigure} \\ \vskip -0.1in
  \begin{subfigure}{\linewidth}
    \centering
    \caption{\framework with SGCN as backbone under random negative link deletion.}
    \label{fig:neg_del_sgcn}
  \end{subfigure} \\

  \begin{subfigure}{\linewidth}
    \centering
    \includegraphics[width=0.8\linewidth]{figs/legend-SNEA.pdf}
  \end{subfigure} \\
  \vskip -0.1in
  \begin{subfigure}{0.24\linewidth}
    \centering
    \includegraphics[width=\linewidth]{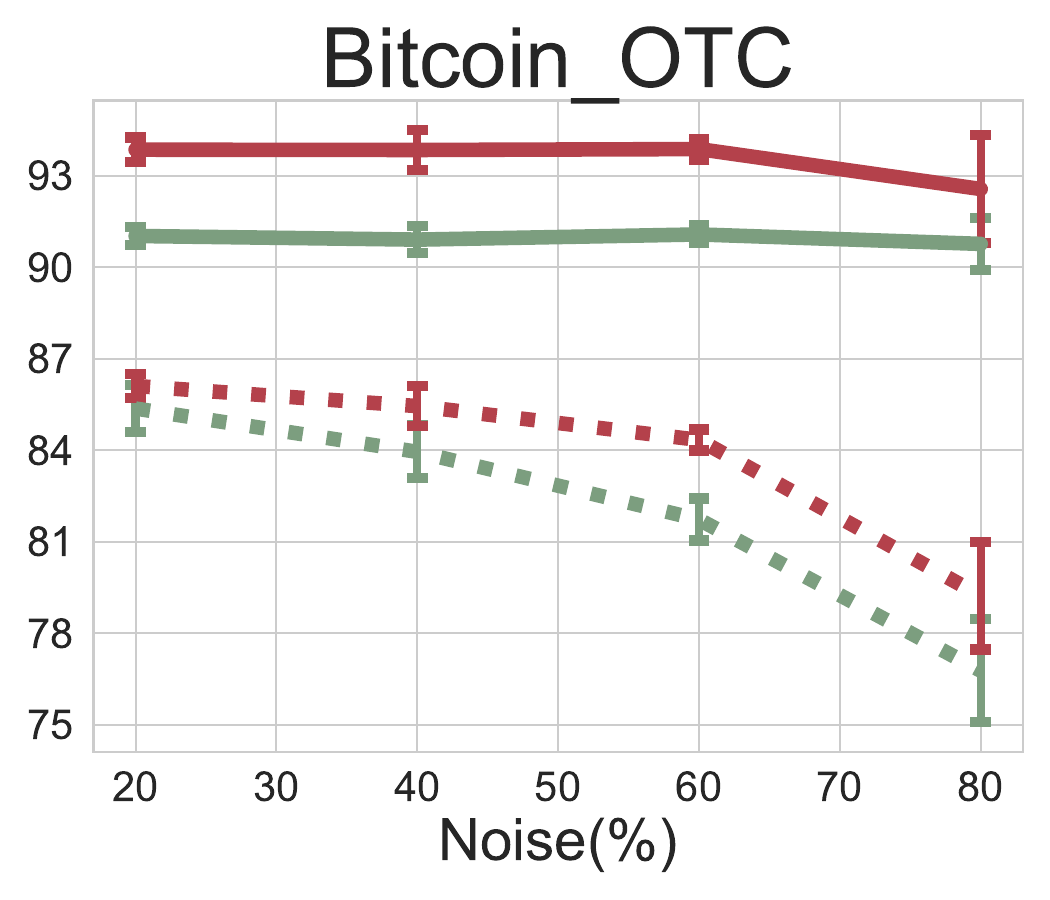}
    % \caption{Bitcoin\_OTC.}
    % \label{fig:node_cls}
  \end{subfigure}
  \begin{subfigure}{0.24\linewidth}
    \centering
    \includegraphics[width=\linewidth]{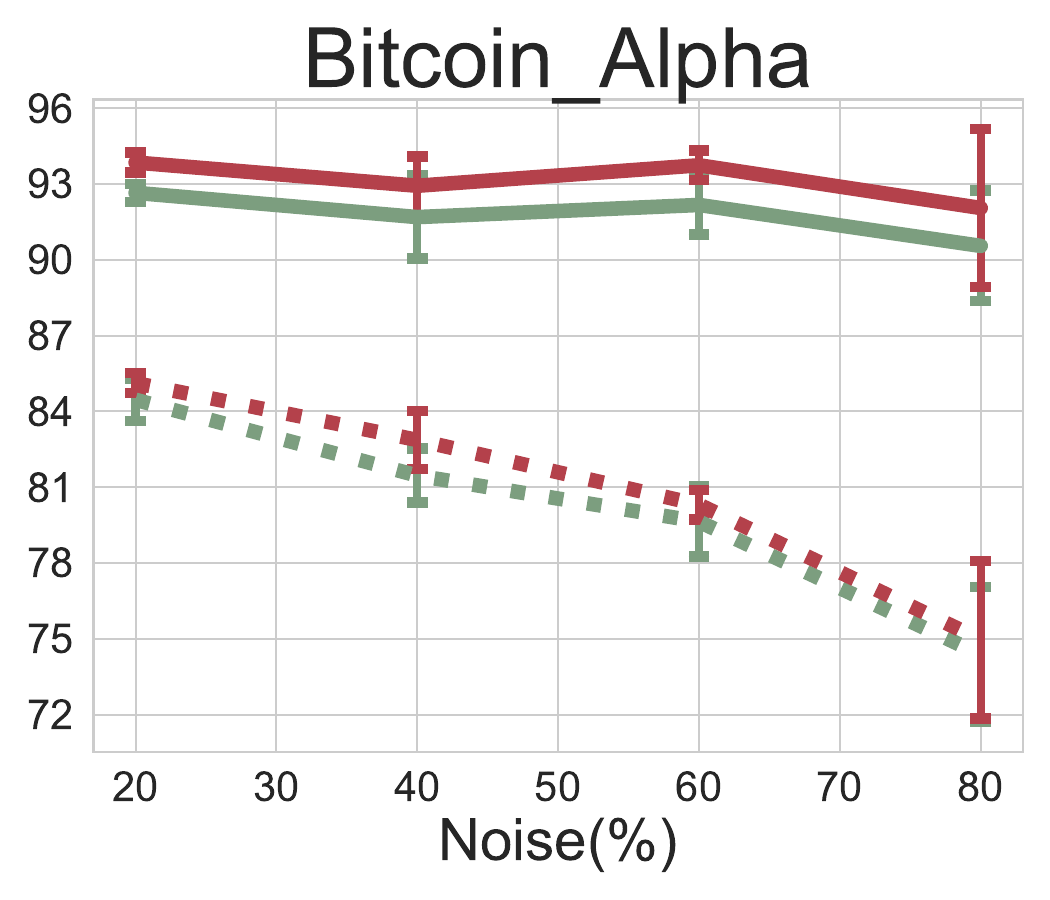}
    % \caption{Bitcoin\_Alpha}
    % \label{fig:link_pre}
  \end{subfigure} 
  \begin{subfigure}{0.24\linewidth}
    \centering
    \includegraphics[width=\linewidth]{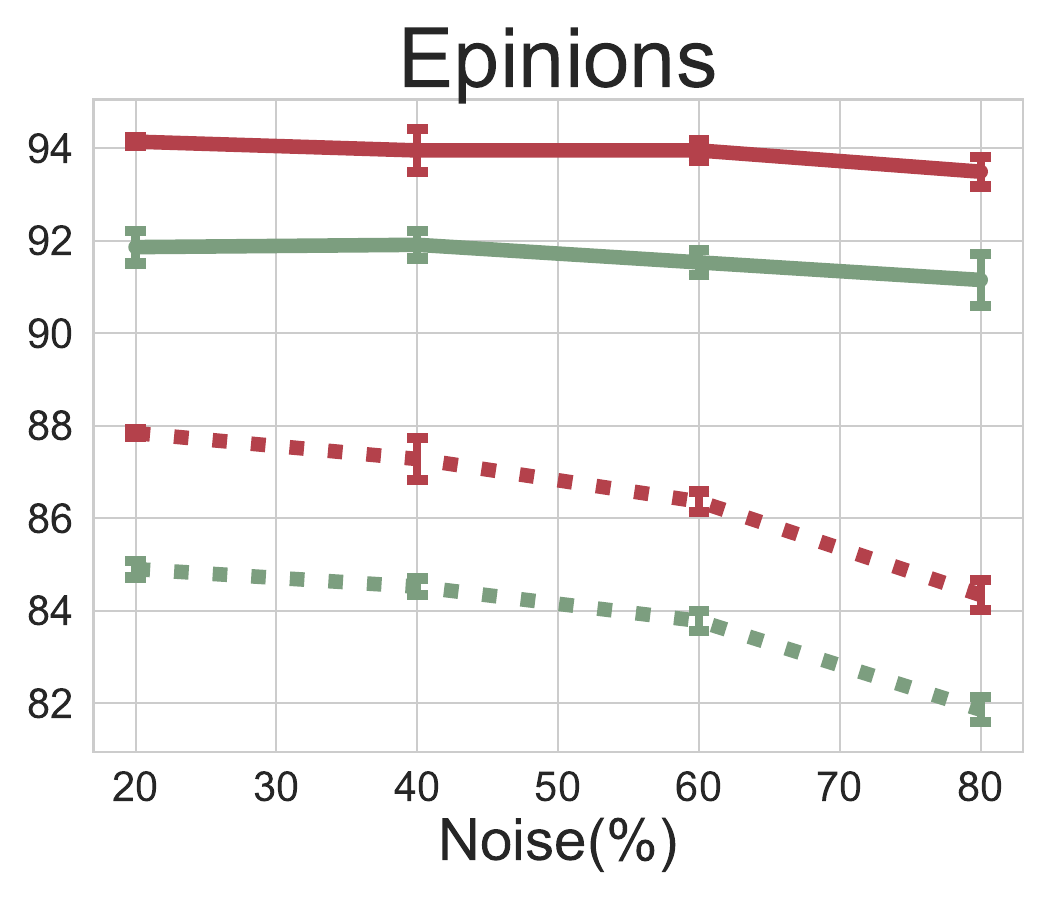}
    % \caption{Epinions.}
    % \label{fig:link_sign_pre}
  \end{subfigure}
  \begin{subfigure}{0.24\linewidth}
    \centering
    \includegraphics[width=\linewidth]{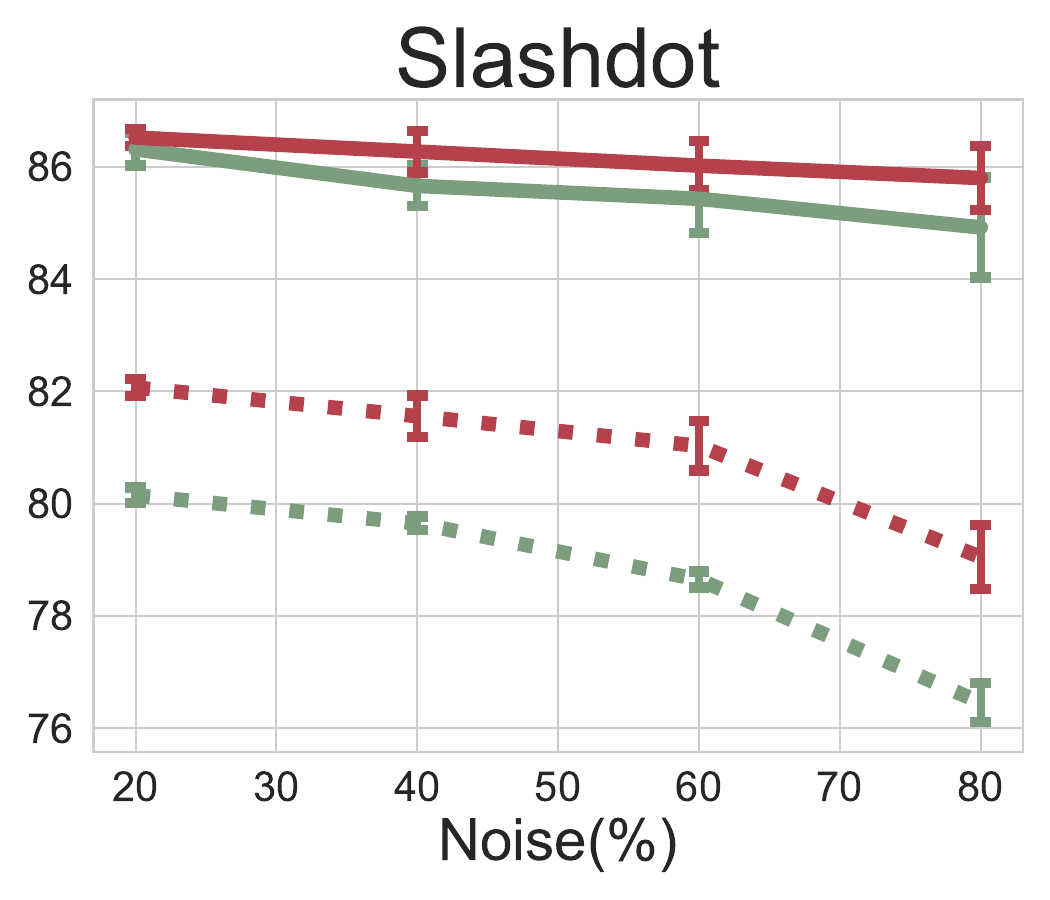}
    % \caption{Slashdot.}
  \end{subfigure} \\ \vskip -0.1in
  \begin{subfigure}{\linewidth}
    \centering  
    \caption{\framework with SNEA as backbone under random negative link deletion.}
    \label{fig:neg_del_snea}
  \end{subfigure}
   \vskip -0.1in
  \caption{\textbf{Results of \framework with different backbones under random negative link deletion.}}
  \label{fig:neg_link_del}
\end{figure*}

\subsection{Random Negative Link Deletion}
Given a signed graph $\mathcal{G}=\{\mathcal{U}, \mathcal{E}^+, \mathcal{E}^-\}$, a subset of negative links is randomly selected from the input signed graph and their signs are deleted. 
For simplicity, $A^+\in\{0, 1\}$ derived from $\mathcal{E}^+$ denotes the positive matrix, $A^-\in\{0, -1\}$ derived from $\mathcal{E}^-$ denotes the negative matrix and $A = A^+ + A^-$.
The proportion of links selected corresponds to the level of noise introduced to the graph. Specifically, for a given noise ratio $\gamma$, the noisy adjacency matrix $\tilde{A}$ is generated by multiplying each element in $A^-$ by 0 with probability $\gamma$, such that $\tilde{A} = \text{NoiseMask}\,\odot\, A^- + A^+$. This ensures that $\gamma = \sfrac{|\mathbf{nonzero}(A)| - |\mathbf{nonzero}(\tilde{A})|}{|\mathbf{nonzero}(A^-)|}$.
Results are shown in Fig.~\ref{fig:neg_link_del}.

\subsection{Random Link Addition}
\begin{figure*}[!ht]
  \centering
  \begin{subfigure}{\linewidth}
    \centering
    \includegraphics[width=0.8\linewidth]{figs/legend-SGCN.pdf}
  \end{subfigure} \\
  \vskip -0.1in
  \begin{subfigure}{0.24\linewidth}
    \centering
    \includegraphics[width=\linewidth]{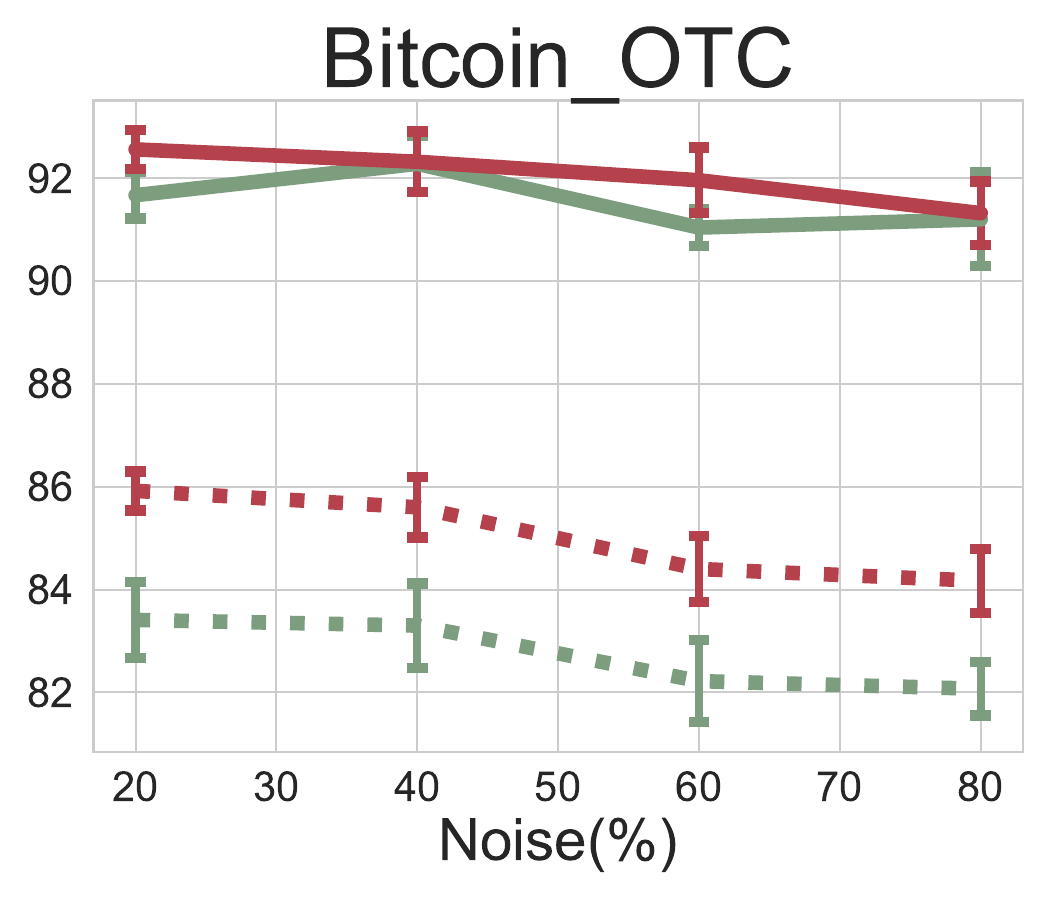}
    % \caption{Bitcoin\_OTC.}
    % \label{fig:node_cls}
  \end{subfigure}
  \begin{subfigure}{0.24\linewidth}
    \centering
    \includegraphics[width=\linewidth]{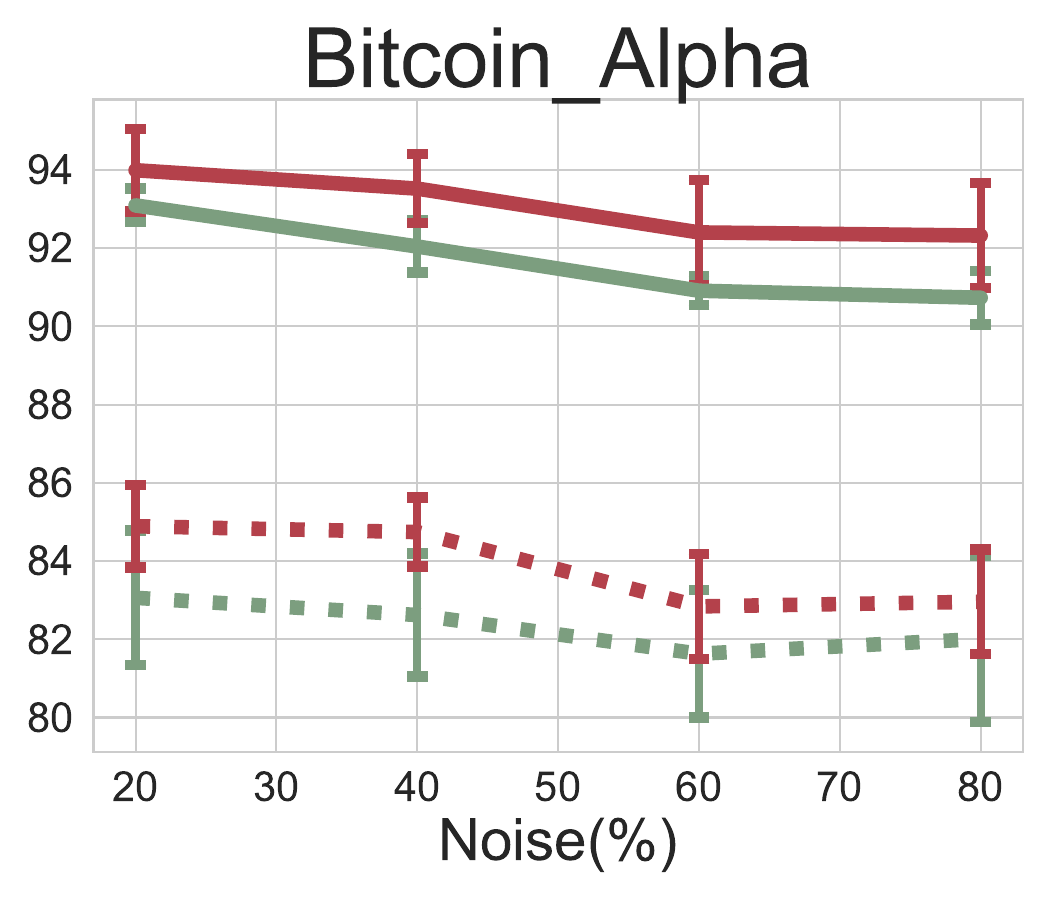}
    % \caption{Bitcoin\_Alpha}
    % \label{fig:link_pre}
  \end{subfigure} 
  \begin{subfigure}{0.24\linewidth}
    \centering
    \includegraphics[width=\linewidth]{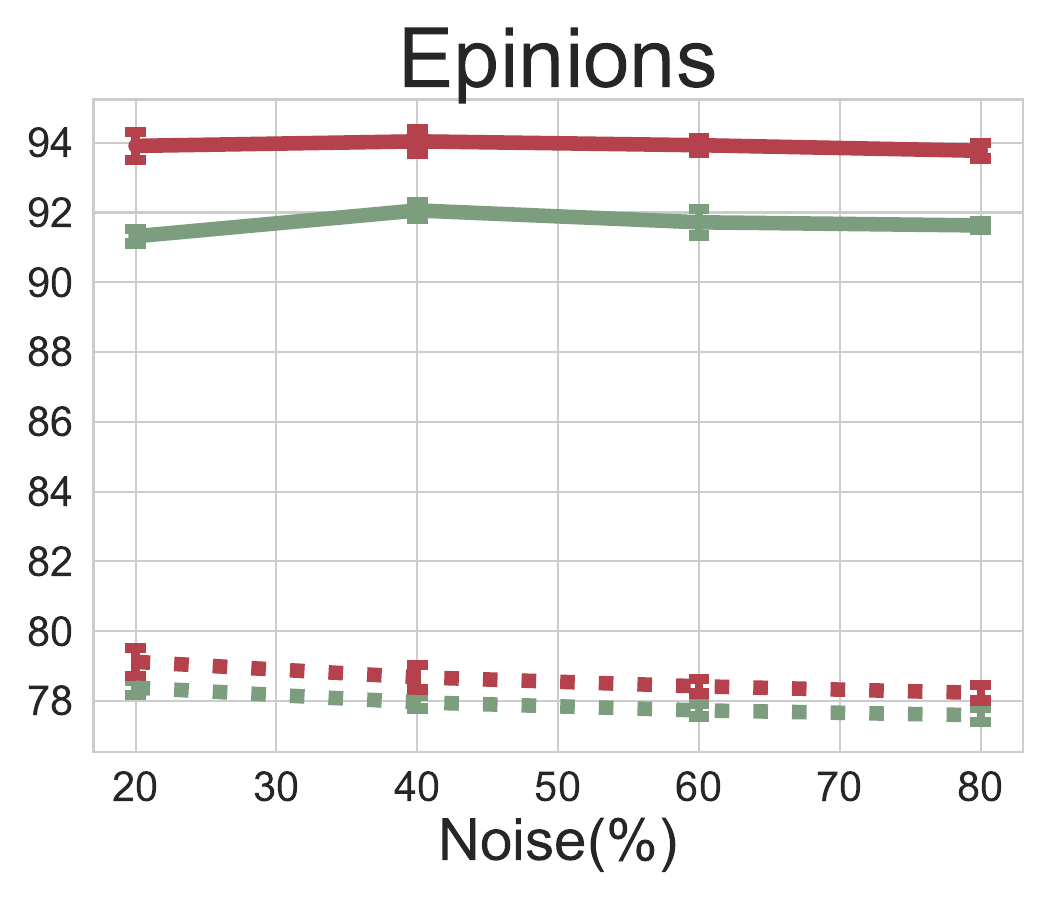}
    % \caption{Epinions.}
    % \label{fig:link_sign_pre}
  \end{subfigure}
  \begin{subfigure}{0.24\linewidth}
    \centering
    \includegraphics[width=\linewidth]{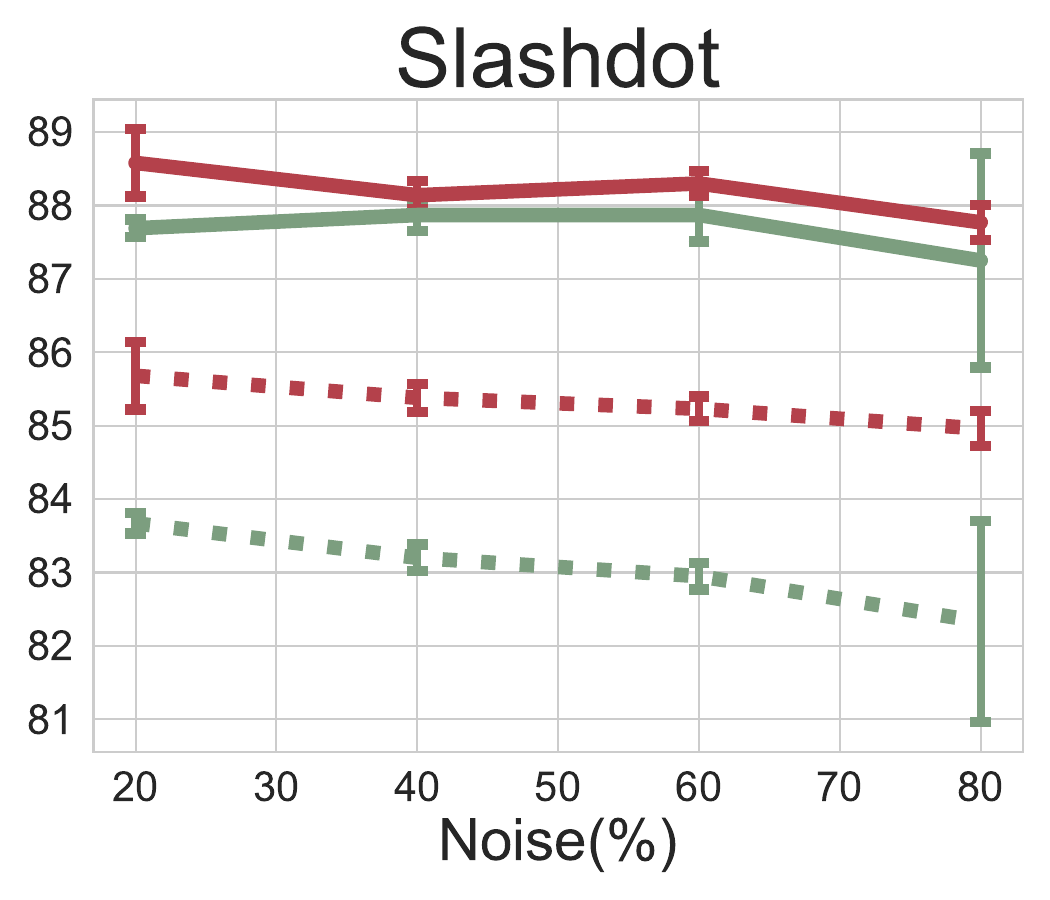}
    % \caption{Slashdot.}
  \end{subfigure} \\ \vskip -0.1in
  \begin{subfigure}{\linewidth}
    \centering
    \caption{\framework with SGCN as backbone under random link addition.}
    \label{fig:add_sgcn}
  \end{subfigure} \\

  \begin{subfigure}{\linewidth}
    \centering
    \includegraphics[width=0.8\linewidth]{figs/legend-SNEA.pdf}
  \end{subfigure} \\
  \vskip -0.1in
  \begin{subfigure}{0.24\linewidth}
    \centering
    \includegraphics[width=\linewidth]{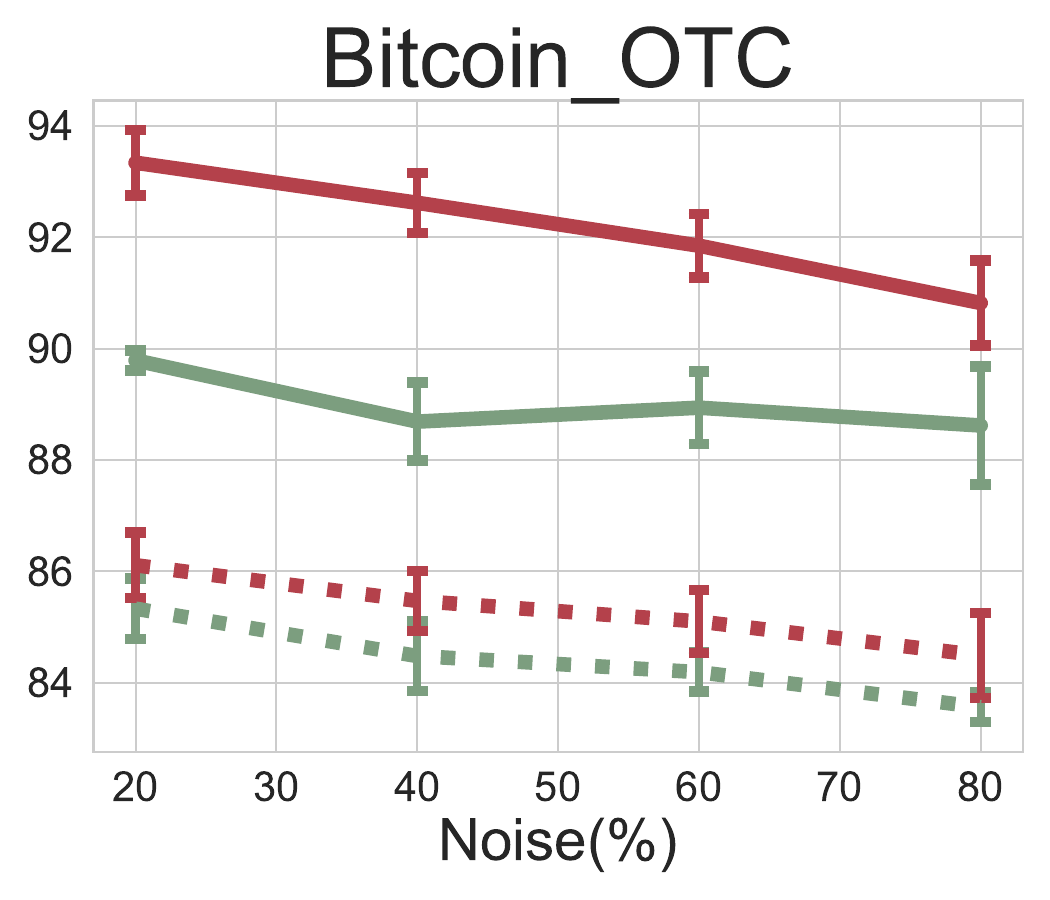}
    % \caption{Bitcoin\_OTC.}
    % \label{fig:node_cls}
  \end{subfigure}
  \begin{subfigure}{0.24\linewidth}
    \centering
    \includegraphics[width=\linewidth]{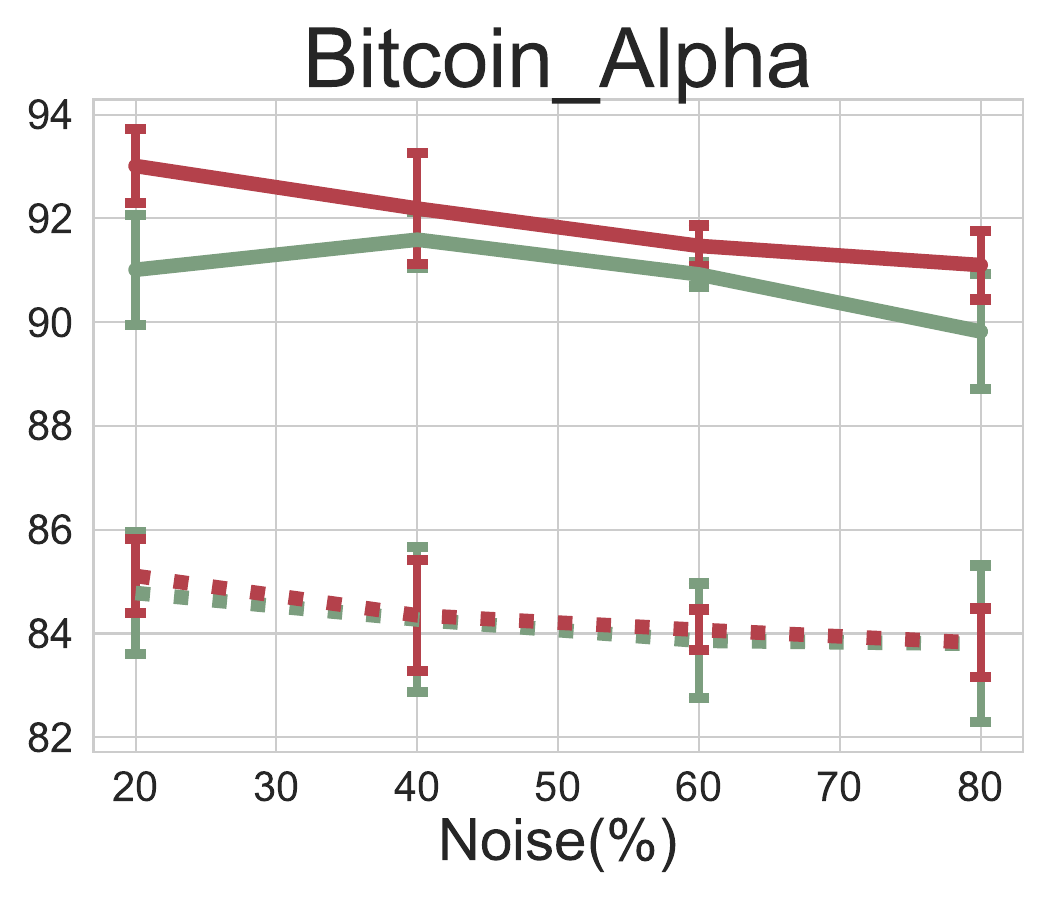}
    % \caption{Bitcoin\_Alpha}
    % \label{fig:link_pre}
  \end{subfigure} 
  \begin{subfigure}{0.24\linewidth}
    \centering
    \includegraphics[width=\linewidth]{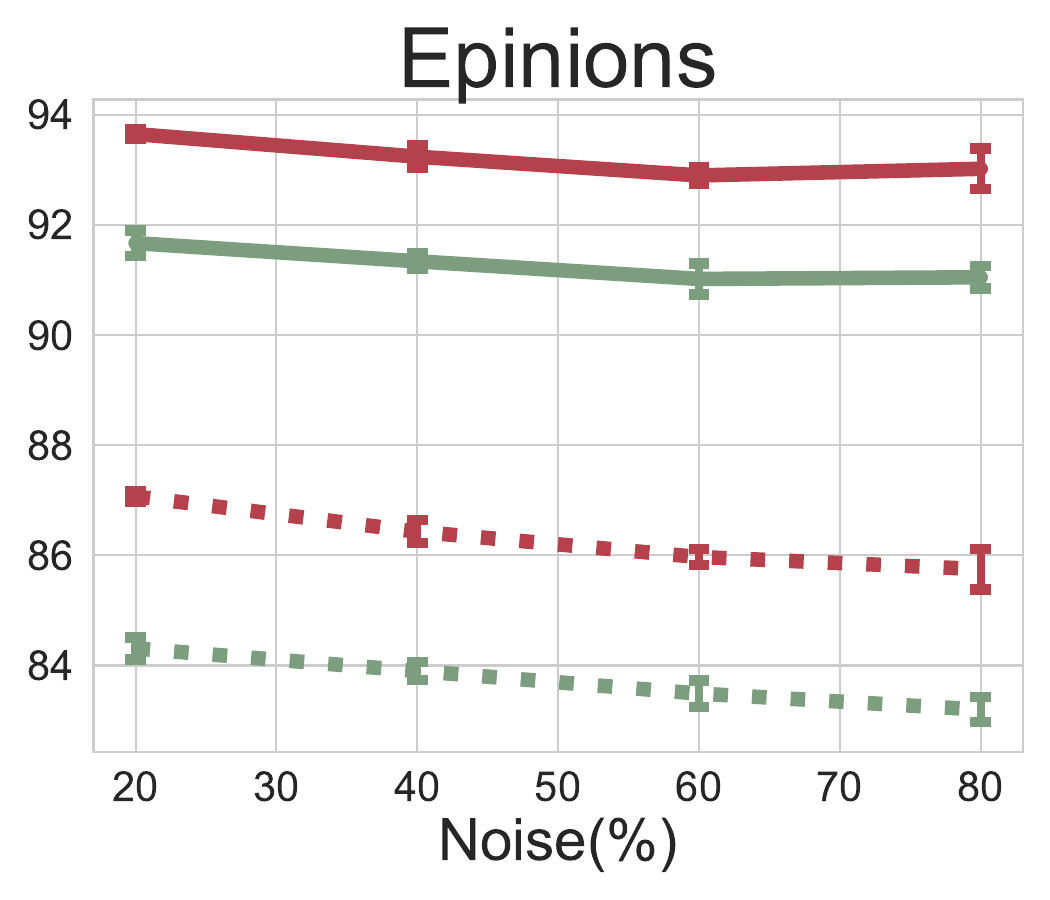}
    % \caption{Epinions.}
    % \label{fig:link_sign_pre}
  \end{subfigure}
  \begin{subfigure}{0.24\linewidth}
    \centering
    \includegraphics[width=\linewidth]{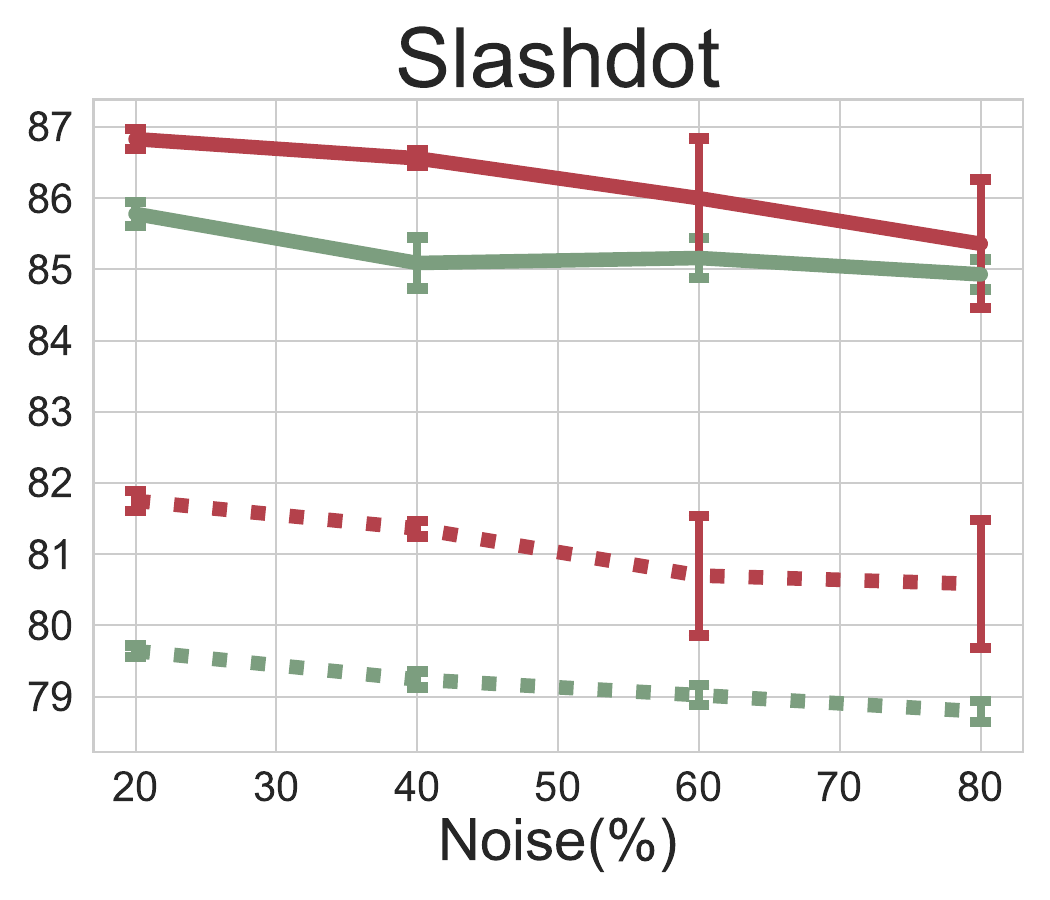}
    % \caption{Slashdot.}
  \end{subfigure} \\ \vskip -0.1in
  \begin{subfigure}{\linewidth}
    \centering  
    \caption{\framework with SNEA as backbone under random link addition.}
    \label{fig:add_snea}
  \end{subfigure}
   \vskip -0.1in
  \caption{\textbf{Results of \framework with different backbones under random link addition.}}
  \label{fig:link_add}
\end{figure*}

Given a signed graph $\mathcal{G}=\{\mathcal{U}, \mathcal{E}^+, \mathcal{E}^-\}$, the noisy adjacency matrix $\tilde{A}$ is generated by directly adding link noise to the clean $A$. 
The proportion of links added corresponds to the level of noise introduced to the graph.
Notably, the ratio of positive links to negative links in addition remains the same as in the original adjacency matrix.
Specifically, for a given noise ratio $\gamma$, the added noisy adjacency matrix $A'\,(\tilde{A} = A + A')$ is generated by flipping the zero elements in $A$ to 1 with probability $\gamma * PosRatio$ and to -1 with probability $\gamma * (1-PosRatio)$, where $PosRatio = |\mathcal{E}^+| / |\mathcal{E}^+ \cup \mathcal{E}^-|$. It satisfies that $A'\,\odot\, A=O$ and $\gamma = \sfrac{|\mathbf{nonzero}(\tilde{A})| - |\mathbf{nonzero}(A)|}{|\mathbf{nonzero}(A)|}$.
Results are shown in Fig.~\ref{fig:link_add}.

\subsection{Random Positive Link Addition}
\begin{figure*}[!ht]
  \centering
  \begin{subfigure}{\linewidth}
    \centering
    \includegraphics[width=0.8\linewidth]{figs/legend-SGCN.pdf}
  \end{subfigure} \\
  \vskip -0.1in
  \begin{subfigure}{0.24\linewidth}
    \centering
    \includegraphics[width=\linewidth]{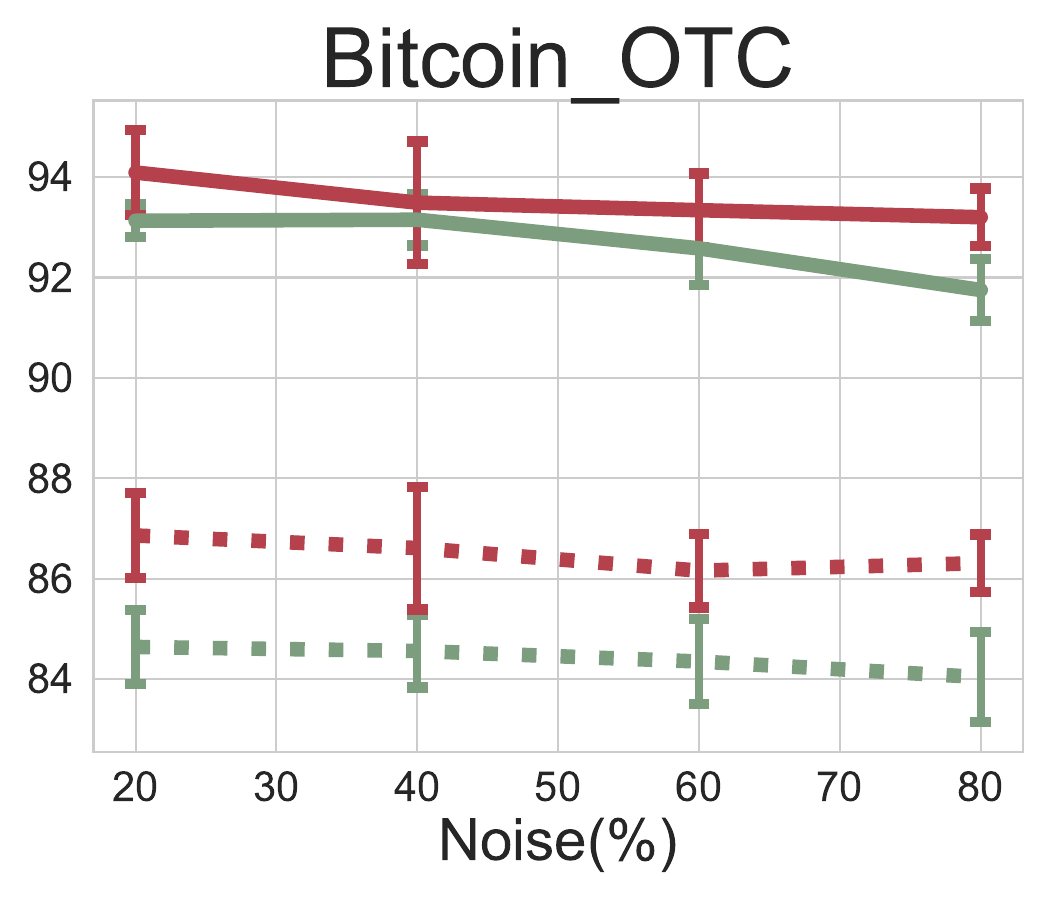}
    % \caption{Bitcoin\_OTC.}
    % \label{fig:node_cls}
  \end{subfigure}
  \begin{subfigure}{0.24\linewidth}
    \centering
    \includegraphics[width=\linewidth]{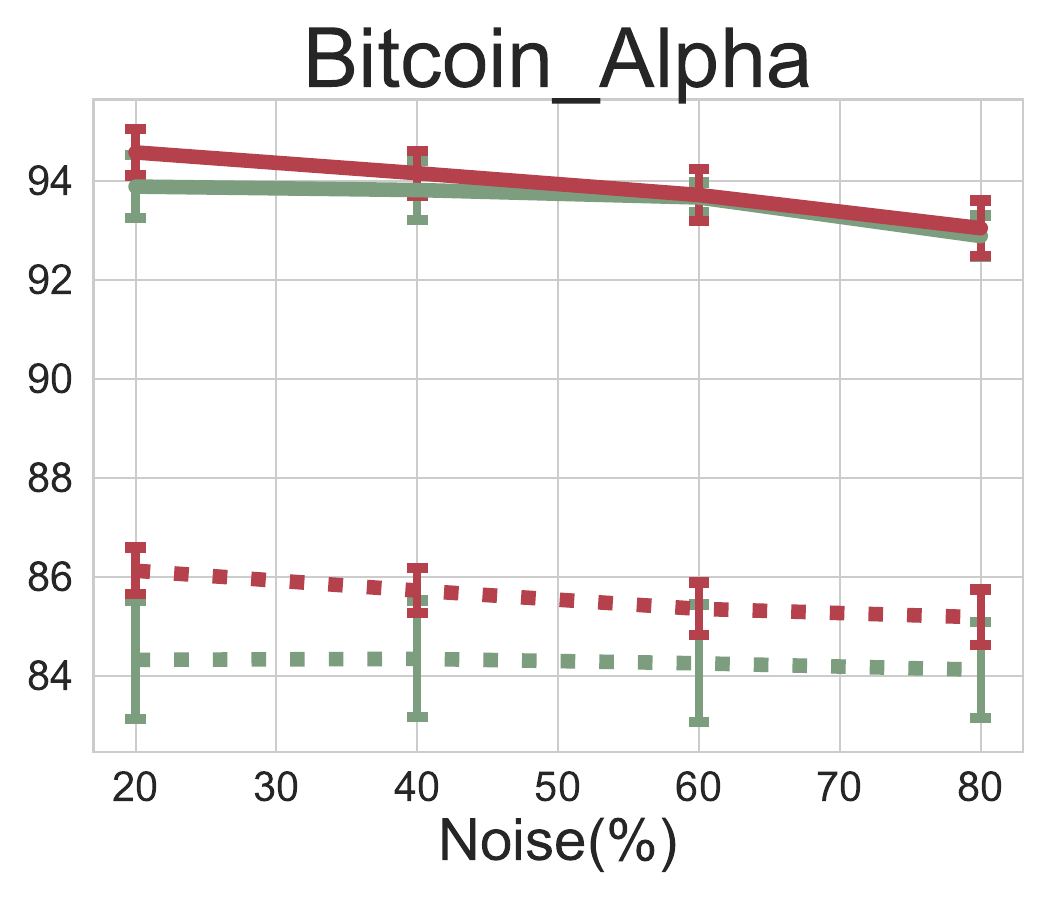}
    % \caption{Bitcoin\_Alpha}
    % \label{fig:link_pre}
  \end{subfigure} 
  \begin{subfigure}{0.24\linewidth}
    \centering
    \includegraphics[width=\linewidth]{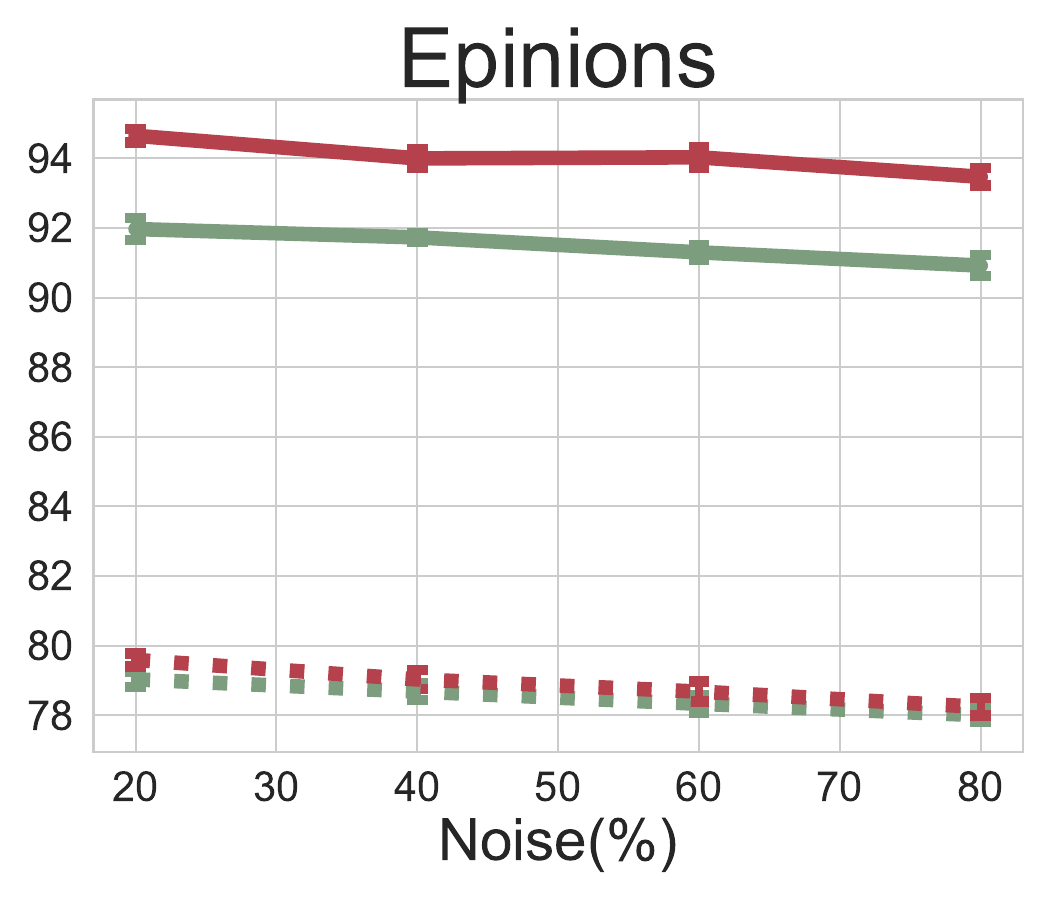}
    % \caption{Epinions.}
    % \label{fig:link_sign_pre}
  \end{subfigure}
  \begin{subfigure}{0.24\linewidth}
    \centering
    \includegraphics[width=\linewidth]{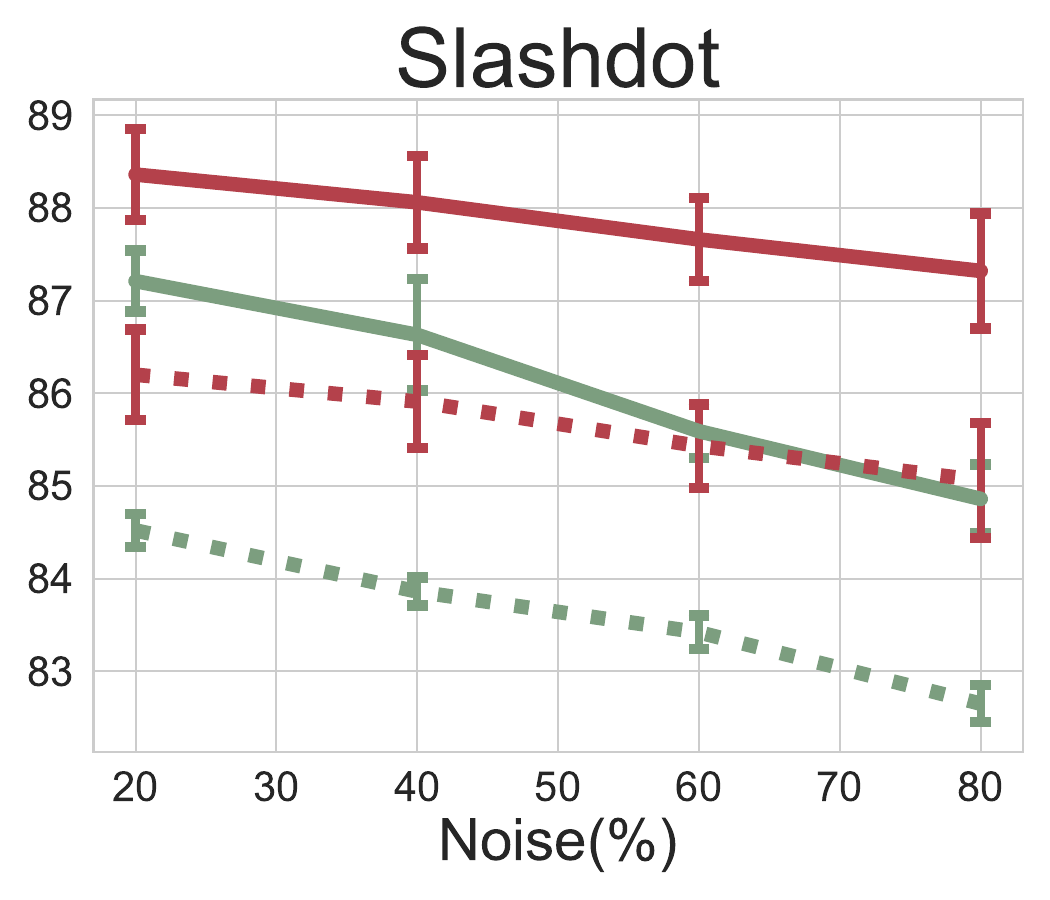}
    % \caption{Slashdot.}
  \end{subfigure} \\ \vskip -0.1in
  \begin{subfigure}{\linewidth}
    \centering
    \caption{\framework with SGCN as backbone under random positive link addition.}
    \label{fig:pos_add_sgcn}
  \end{subfigure} \\

  \begin{subfigure}{\linewidth}
    \centering
    \includegraphics[width=0.8\linewidth]{figs/legend-SNEA.pdf}
  \end{subfigure} \\
  \vskip -0.1in
  \begin{subfigure}{0.24\linewidth}
    \centering
    \includegraphics[width=\linewidth]{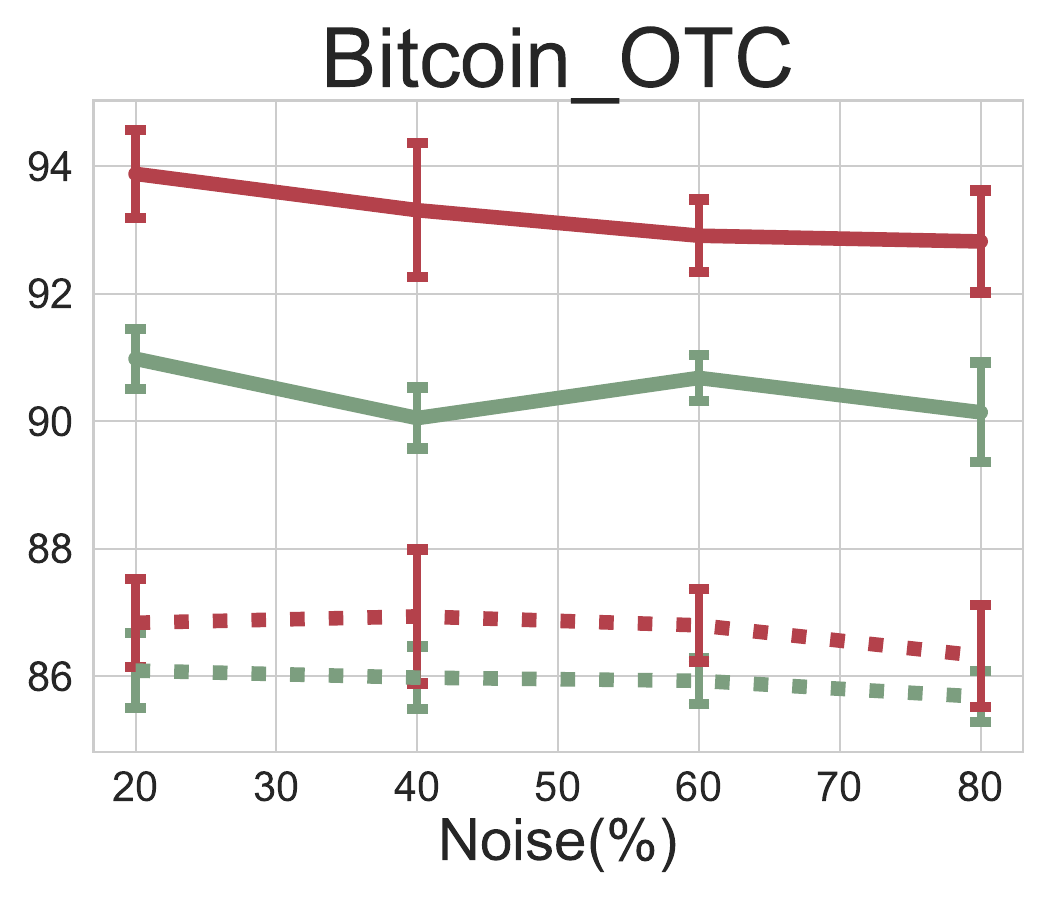}
    % \caption{Bitcoin\_OTC.}
    % \label{fig:node_cls}
  \end{subfigure}
  \begin{subfigure}{0.24\linewidth}
    \centering
    \includegraphics[width=\linewidth]{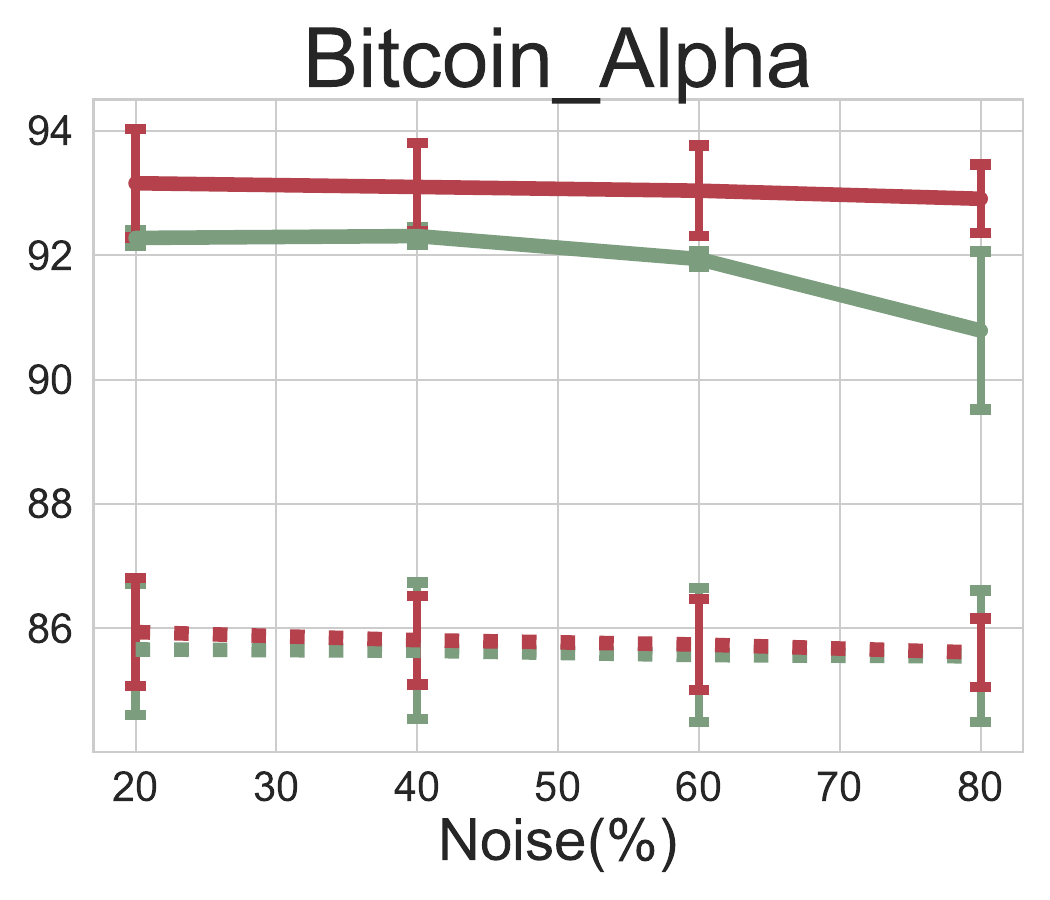}
    % \caption{Bitcoin\_Alpha}
    % \label{fig:link_pre}
  \end{subfigure} 
  \begin{subfigure}{0.24\linewidth}
    \centering
    \includegraphics[width=\linewidth]{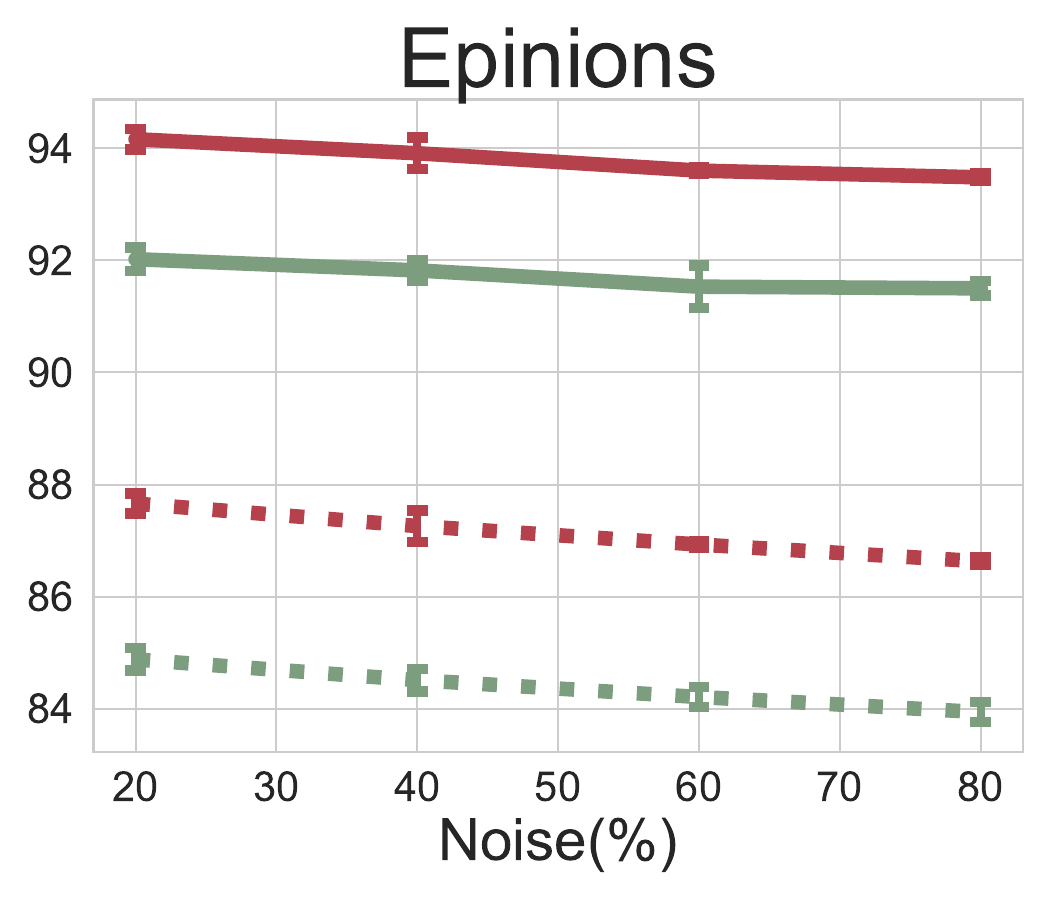}
    % \caption{Epinions.}
    % \label{fig:link_sign_pre}
  \end{subfigure}
  \begin{subfigure}{0.24\linewidth}
    \centering
    \includegraphics[width=\linewidth]{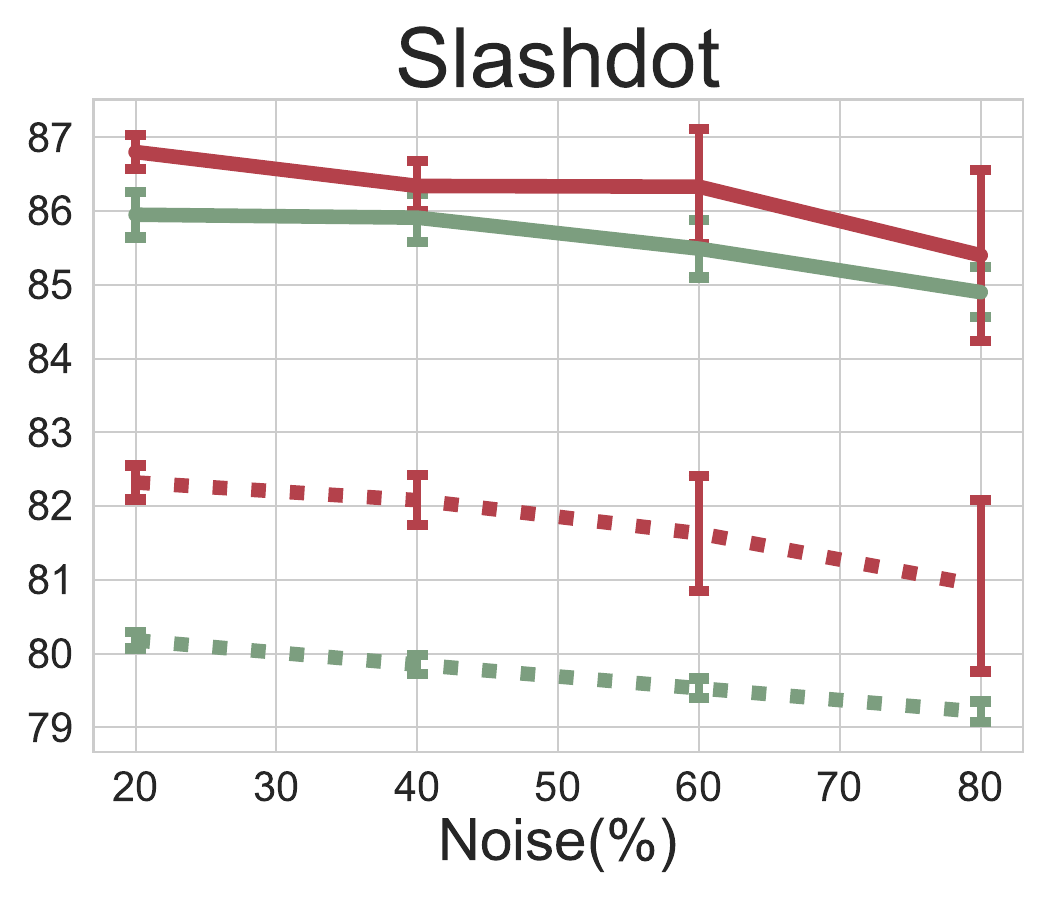}
    % \caption{Slashdot.}
  \end{subfigure} \\ \vskip -0.1in
  \begin{subfigure}{\linewidth}
    \centering  
    \caption{\framework with SNEA as backbone under random positive link addition.}
    \label{fig:pos_add_snea}
  \end{subfigure}
   \vskip -0.1in
  \caption{\textbf{Results of \framework with different backbones under random positive link addition.}}
  \label{fig:pos_link_add}
\end{figure*}

Given a signed graph $\mathcal{G}=\{\mathcal{U}, \mathcal{E}^+, \mathcal{E}^-\}$, the noisy adjacency matrix $\tilde{A}$ is generated by directly adding positive link noise to the clean $A$. 
The proportion of links added corresponds to the level of noise introduced to the graph.
Specifically, for a given noise ratio $\gamma$, the added noisy adjacency matrix $A'\,(\tilde{A} = A + A')$ is generated by flipping the zero elements in $A$ to 1 with probability $\gamma$. It satisfies that $A'\,\odot\, A=O$ and $\gamma = \sfrac{|\mathbf{nonzero}(\tilde{A})| - |\mathbf{nonzero}(A)|}{|\mathbf{nonzero}(A)|}$.
Results are shown in Fig.~\ref{fig:pos_link_add}.

\subsection{Random Negative Link Addition}
\begin{figure*}[!ht]
  \centering
  \begin{subfigure}{\linewidth}
    \centering
    \includegraphics[width=0.8\linewidth]{figs/legend-SGCN.pdf}
  \end{subfigure} \\
  \vskip -0.1in
  \begin{subfigure}{0.24\linewidth}
    \centering
    \includegraphics[width=\linewidth]{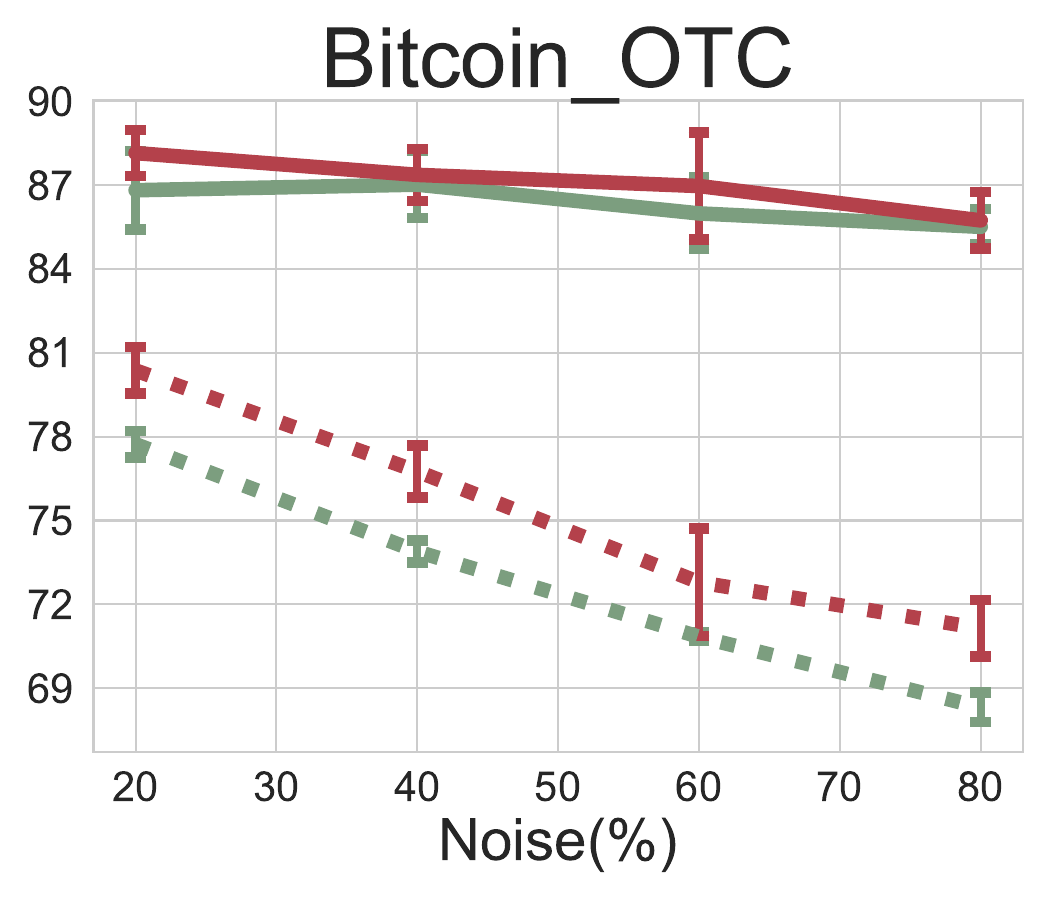}
    % \caption{Bitcoin\_OTC.}
    % \label{fig:node_cls}
  \end{subfigure}
  \begin{subfigure}{0.24\linewidth}
    \centering
    \includegraphics[width=\linewidth]{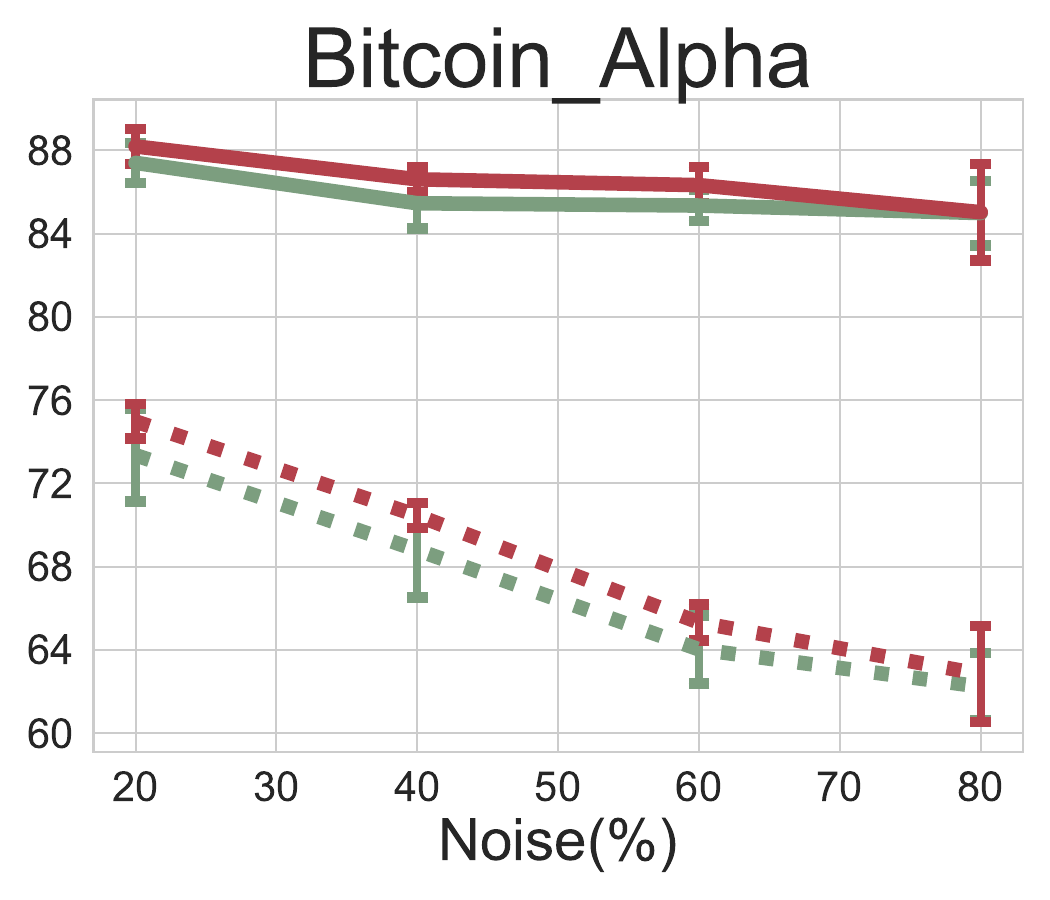}
    % \caption{Bitcoin\_Alpha}
    % \label{fig:link_pre}
  \end{subfigure} 
  \begin{subfigure}{0.24\linewidth}
    \centering
    \includegraphics[width=\linewidth]{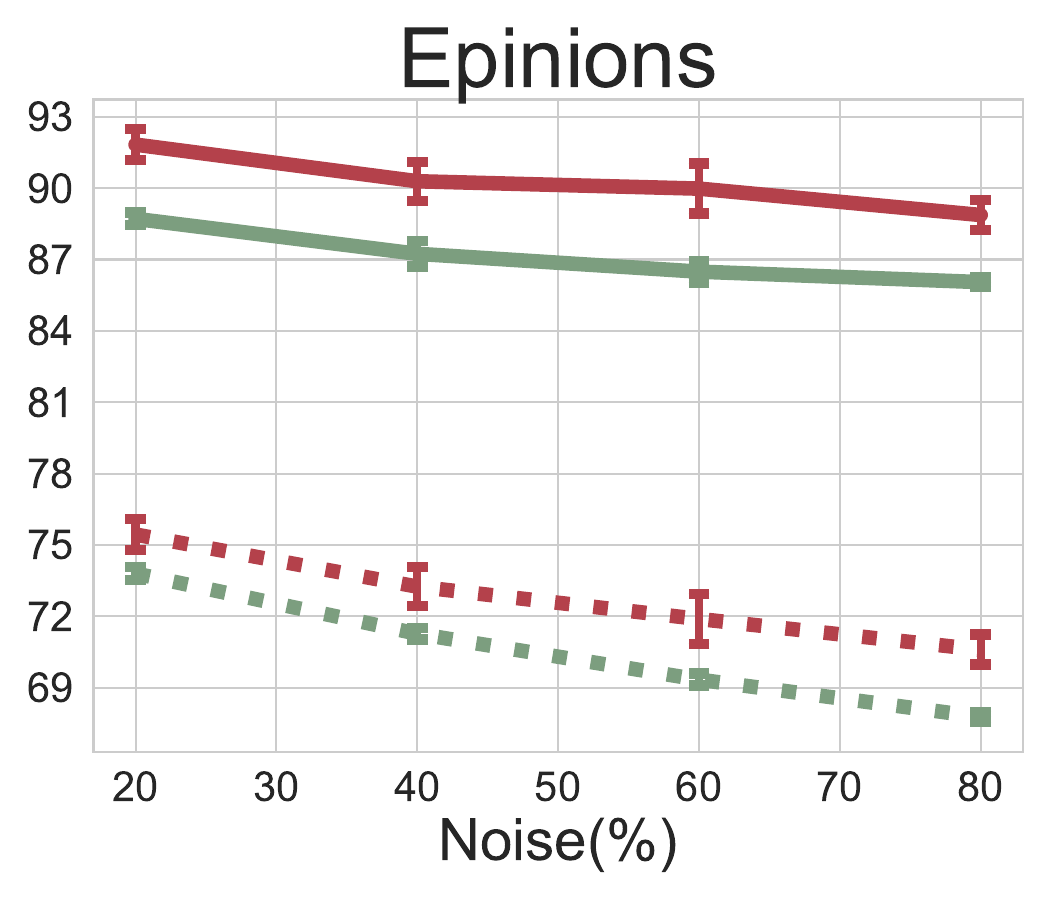}
    % \caption{Epinions.}
    % \label{fig:link_sign_pre}
  \end{subfigure}
  \begin{subfigure}{0.24\linewidth}
    \centering
    \includegraphics[width=\linewidth]{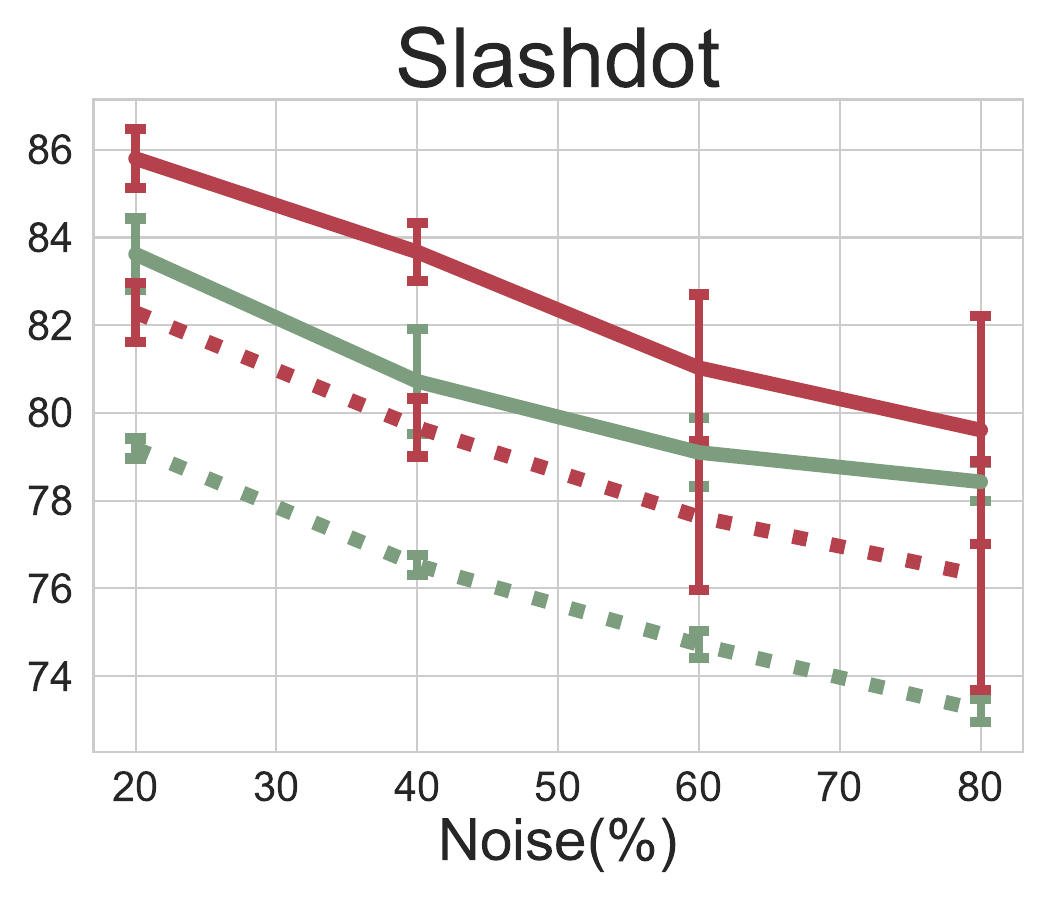}
    % \caption{Slashdot.}
  \end{subfigure} \\ \vskip -0.1in
  \begin{subfigure}{\linewidth}
    \centering
    \caption{\framework with SGCN as backbone under random negative link addition.}
    \label{fig:neg_add_sgcn}
  \end{subfigure} \\

  \begin{subfigure}{\linewidth}
    \centering
    \includegraphics[width=0.8\linewidth]{figs/legend-SNEA.pdf}
  \end{subfigure} \\
  \vskip -0.1in
  \begin{subfigure}{0.24\linewidth}
    \centering
    \includegraphics[width=\linewidth]{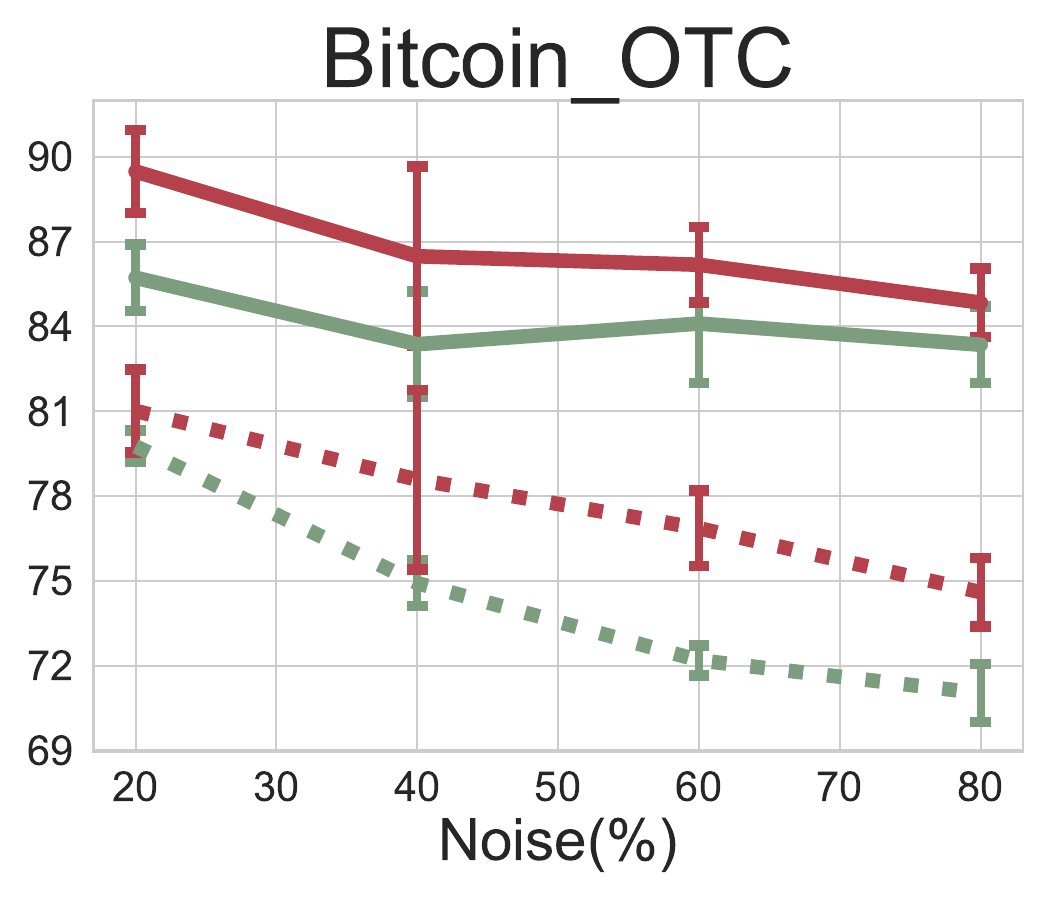}
    % \caption{Bitcoin\_OTC.}
    % \label{fig:node_cls}
  \end{subfigure}
  \begin{subfigure}{0.24\linewidth}
    \centering
    \includegraphics[width=\linewidth]{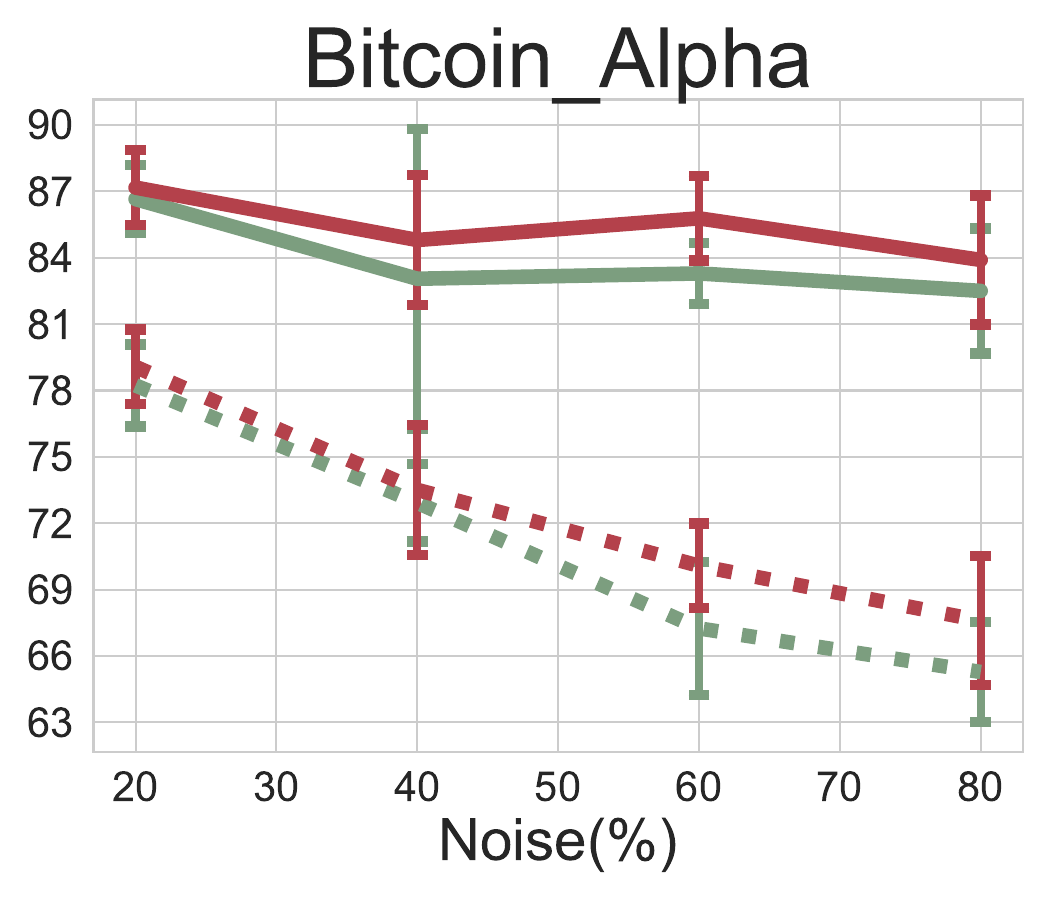}
    % \caption{Bitcoin\_Alpha}
    % \label{fig:link_pre}
  \end{subfigure} 
  \begin{subfigure}{0.24\linewidth}
    \centering
    \includegraphics[width=\linewidth]{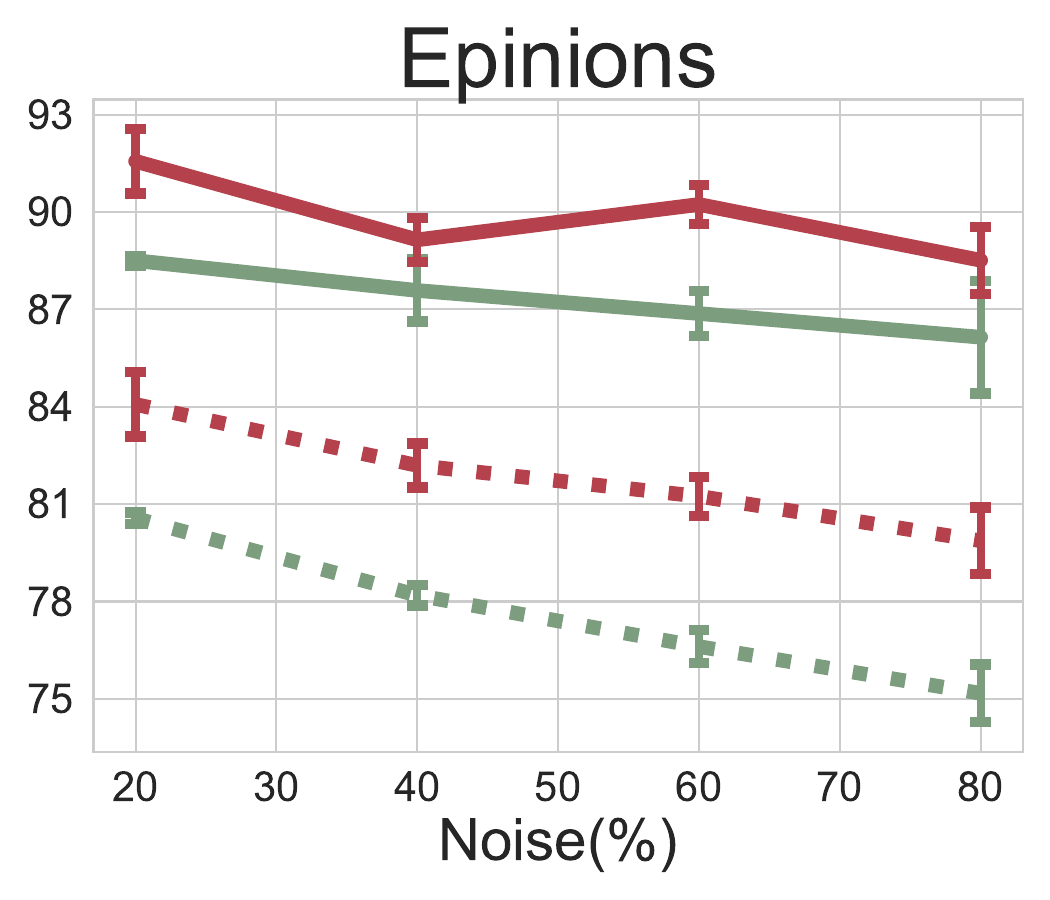}
    % \caption{Epinions.}
    % \label{fig:link_sign_pre}
  \end{subfigure}
  \begin{subfigure}{0.24\linewidth}
    \centering
    \includegraphics[width=\linewidth]{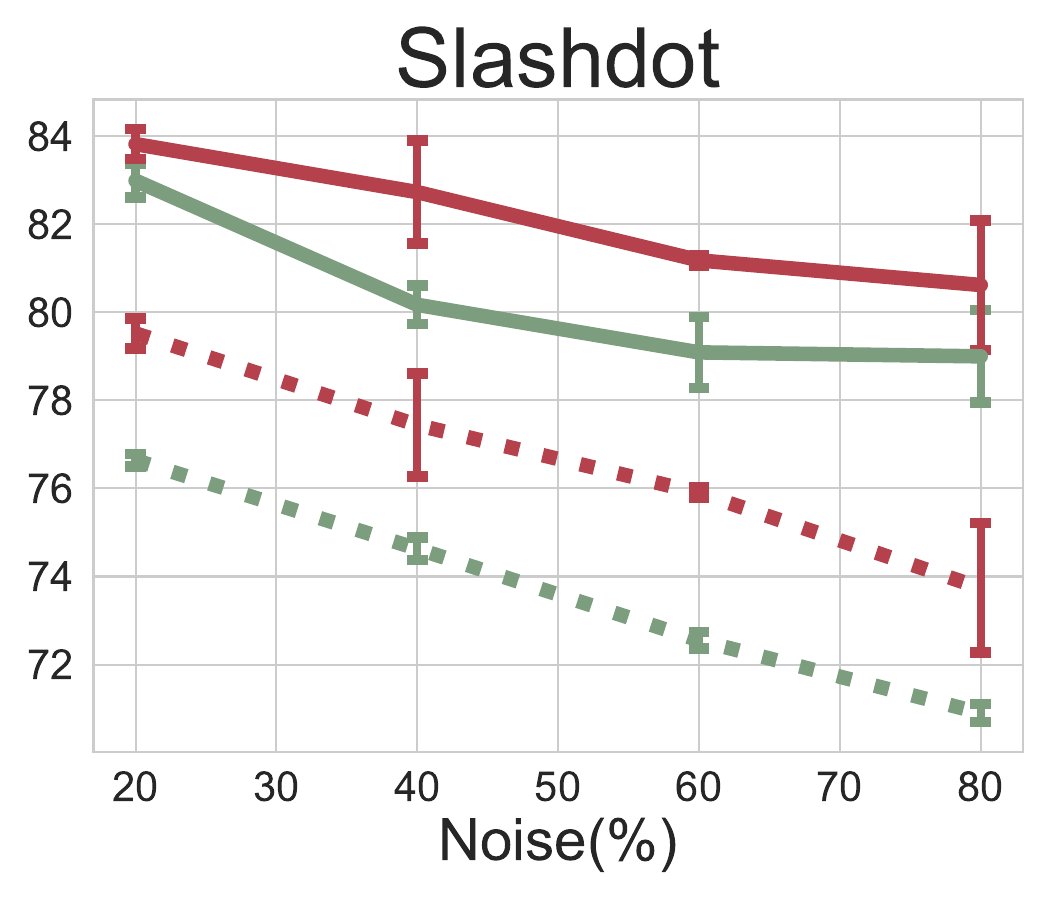}
    % \caption{Slashdot.}
  \end{subfigure} \\ \vskip -0.1in
  \begin{subfigure}{\linewidth}
    \centering  
    \caption{\framework with SNEA as backbone under random negative link addition.}
    \label{fig:neg_add_snea}
  \end{subfigure}
   \vskip -0.1in
  \caption{\textbf{Results of \framework with different backbones under random negative link addition.}}
  \label{fig:neg_link_add}
\end{figure*}

Given a signed graph $\mathcal{G}=\{\mathcal{U}, \mathcal{E}^+, \mathcal{E}^-\}$, the noisy adjacency matrix $\tilde{A}$ is generated by directly adding negative link noise to the clean $A$. 
The proportion of links added corresponds to the level of noise introduced to the graph.
Specifically, for a given noise ratio $\gamma$, the added noisy adjacency matrix $A'\,(\tilde{A} = A + A')$ is generated by flipping the zero elements in $A$ to -1 with probability $\gamma$. It satisfies that $A'\,\odot\, A=O$ and $\gamma = \sfrac{|\mathbf{nonzero}(\tilde{A})| - |\mathbf{nonzero}(A)|}{|\mathbf{nonzero}(A)|}$.
Results are shown in Fig.~\ref{fig:neg_link_add}.

\end{document}